%% file: main_arxiv.tex
\documentclass{article}

\usepackage{iclr2027_conference,times}
\usepackage{amsmath,amssymb,amsthm,mathtools}
\usepackage{booktabs,multirow,array}
\usepackage{graphicx}
\usepackage{enumitem}
\usepackage{placeins}
\usepackage{xcolor}
\usepackage{microtype}
\usepackage{tikz}
\usepackage{hyperref}
\hypersetup{hidelinks,
  pdftitle={Bridging the Data Gap: Source-Informed Task-Space Transport for Few-Shot Diffusion},
  pdfauthor={Jinjian Jiang, Jing Peng, Rehan Deen, Yu Cao, Ziquan Liu, Muhammad Salman Haleem, David Mguni}
}
\usetikzlibrary{arrows.meta,positioning,fit,calc}

\setlist{nosep,leftmargin=*}

\newtheorem{proposition}{Proposition}

\newcommand{\E}{\mathbb{E}}
\newcommand{\KT}{K_T}
\newcommand{\MS}{M_S}
\newcommand{\Ttrain}{\mathcal{Y}_{\mathrm{train}}}
\newcommand{\Tval}{\mathcal{Y}_{\mathrm{val}}}
\newcommand{\Ttest}{\mathcal{Y}_{\mathrm{test}}}
\newcommand{\epshat}{\widehat{\epsilon}}
\newcommand{\sigstar}{^{\ast}}

\title{Bridging the Data Gap: Source-Informed Task-Space Transport for Few-Shot Diffusion}
\author{
\begin{tabular}{ccccc}
\textbf{Jinjian Jiang}\textsuperscript{1} &
\textbf{Jing Peng}\textsuperscript{2} &
\textbf{Rehan Deen}\textsuperscript{3} &
\textbf{Yu Cao}\textsuperscript{4} &
\\[0.45em]
&\textbf{Ziquan Liu}\textsuperscript{4}&
\textbf{Muhammad Salman Haleem}\textsuperscript{4} 
&\textbf{David Mguni}\textsuperscript{4}\thanks{Corresponding author: \textlangle d.mguni@qmul.ac.uk\textrangle}
\end{tabular}
\\[1.3em]
\begin{tabular}{c}
\textsuperscript{1}University of Warwick
\quad
\textsuperscript{2}Columbia University
\quad
\textsuperscript{3}Independent Researcher
\\[.5em]
\quad\qquad\qquad\textsuperscript{4}Queen Mary University of London
\end{tabular}
}

\iclrfinalcopy
\date{}

\begin{document}
\maketitle

\begin{abstract}
When an object category is well represented in one domain but scarcely observed in another, adapting a diffusion model to the target can be unreliable. Yet the same source--target relationship may have been observed for other categories. We ask whether this recurring relation can be learned and reused for a new semantic task. We introduce \emph{Meta-Diff}, which embeds source and target distributions in a shared low-dimensional task space, learns an affine transport map across paired training tasks, and generates from the transported source coordinate using a shared diffusion backbone. At deployment, it requires source examples and the identity of a previously learned relation, but neither target images nor target-specific optimisation. Gaussian-mixture experiments examine how transfer depends on task information retained by the encoder, relation predictability and the size of the source--target shift. On held-out CIFAR-100 validation tasks, correct-source transport achieves 13.9\% lower mean sliced Wasserstein distance than wrong-source transport. In the one-target final-test comparison, it achieves 14.7\% lower mean distance than target-only adaptation and 32.7\% lower than full fine-tuning under validation-selected settings. These gains have clear limits: directly applying the known corruption to clean-source generations performs better on CIFAR, and requested-class identity remains weak. Office-Home shows domain alignment without reliable object identity; clinical stress tests show no resolved source-specific advantage. The results characterise when reusable task-space relations help and where their generative limitations remain.
\end{abstract}

\section{Introduction}
Consider an object category for which product photographs are plentiful but clipart images are scarce. A conventional approach to generating clipart would adapt a pretrained diffusion model using the few available examples. Yet the difference between photographs and clipart is not unique to that object: we may already have seen it for many other categories. Those earlier source--target pairs offer information that is missed when every new target is treated as a separate adaptation problem. Can a generator learn this recurring relationship and apply it to a category it has never encountered in the target domain?

The same question arises more broadly when data are collected under different acquisition conditions, at different institutions or from unevenly represented populations. In medical imaging, for example, differences in sampling and acquisition can contribute to performance disparities \citep{larrazabal2020gender,seyyedkalantari2021underdiagnosis,daneshjou2022ddi}. Collecting representative target data remains essential, and synthetic data cannot replace clinical validation. Nevertheless, such imbalances motivate a general problem in generative modelling: how much can be learned about a sparsely observed target from a related source and from source--target shifts observed elsewhere?

Most approaches to few-shot diffusion generation seek to make adaptation to the new target more efficient, whether by fine-tuning selected weights, learning a compact embedding or predicting an update to a pretrained generator \citep{ruiz2023dreambooth,kumari2023customdiffusion,hu2022lora,gal2023textualinversion,ruiz2024hyperdreambooth}. Their main source of information about a new target remains its target examples. We investigate a complementary source of information: a relationship between domains that recurs across different semantic tasks. Instead of fitting the target independently each time, we aim to learn the relationship from earlier tasks and reuse it when only the new task's source observations are available.

During meta-training, we observe paired source and target distributions for semantic tasks $y_1,\ldots,y_n$,
\[
P_S^{y_1}\!\rightarrow P_T^{y_1},\ldots,P_S^{y_n}\!\rightarrow P_T^{y_n}.
\]
At deployment, a previously unseen task $y^\star$ has source observations $D_S^{y^\star}$, but target observations need not be available. We assume that the applicable source--target relation was represented during meta-training and that its identity is known. Our problem is thus generalisation to a new semantic task under a familiar relation, not transfer to an arbitrary, unseen domain shift.

\begin{figure*}[!htbp]
\centering
\definecolor{sourceblue}{RGB}{54,105,165}
\definecolor{targetorange}{RGB}{215,122,46}
\definecolor{softgray}{RGB}{246,246,246}
\begin{minipage}[t]{0.31\textwidth}
\centering
\begin{tikzpicture}[x=1cm,y=1cm,every node/.style={font=\scriptsize}]
\node[font=\small\bfseries] at (2.0,2.55) {(a) Observe recurring pairs};
\node[draw=sourceblue,fill=sourceblue!8,rounded corners,minimum width=.9cm,minimum height=.43cm] (s1) at (.62,1.95) {$D_S^{y_1}$};
\node[draw=targetorange,fill=targetorange!10,rounded corners,minimum width=.9cm,minimum height=.43cm] (t1) at (3.35,1.95) {$D_T^{y_1}$};
\draw[-{Latex[length=1.7mm]},thick] (s1)--node[above,font=\tiny]{same relation}(t1);
\node[draw=sourceblue,fill=sourceblue!8,rounded corners,minimum width=.9cm,minimum height=.43cm] (s2) at (.62,1.20) {$D_S^{y_2}$};
\node[draw=targetorange,fill=targetorange!10,rounded corners,minimum width=.9cm,minimum height=.43cm] (t2) at (3.35,1.20) {$D_T^{y_2}$};
\draw[-{Latex[length=1.7mm]},thick] (s2)--node[above,font=\tiny]{same relation}(t2);
\node[draw=sourceblue,fill=sourceblue!8,rounded corners,minimum width=.9cm,minimum height=.43cm] (sn) at (.62,.45) {$D_S^{y_n}$};
\node[draw=targetorange,fill=targetorange!10,rounded corners,minimum width=.9cm,minimum height=.43cm] (tn) at (3.35,.45) {$D_T^{y_n}$};
\draw[-{Latex[length=1.7mm]},thick] (sn)--node[above,font=\tiny]{same relation}(tn);
\node[draw,rounded corners,fill=softgray,align=center,minimum width=1.65cm] at (1.98,-.15) {learn across\\semantic tasks};
\end{tikzpicture}
\end{minipage}\hfill
\begin{minipage}[t]{0.33\textwidth}
\centering
\begin{tikzpicture}[x=1cm,y=1cm,every node/.style={font=\scriptsize}]
\node[font=\small\bfseries] at (2.0,2.55) {(b) Learn task-space transport};
\draw[->,gray] (.25,.18)--(3.85,.18); \draw[->,gray] (.25,.18)--(.25,2.12);
\foreach \sx/\sy/\tx/\ty in {.8/.55/1.45/1.03,1.15/1.60/1.82/1.88,2.0/.62/2.72/1.08}{
 \fill[sourceblue] (\sx,\sy) circle (2.2pt); \fill[targetorange] (\tx,\ty) circle (2.2pt);
 \draw[-{Latex[length=1.5mm]},gray!75,thick] (\sx,\sy)--(\tx,\ty);
}
\fill[sourceblue] (2.65,1.62) circle (2.6pt); \node[above left] at (2.65,1.62) {$z_S^{y^\star}$};
\fill[targetorange] (3.40,2.00) circle (2.6pt); \node[above right] at (3.40,2.00) {$\widetilde z_T^{y^\star}$};
\draw[-{Latex[length=1.9mm]},very thick] (2.65,1.62)--(3.40,2.00);
\node[font=\tiny] at (1.05,-.15) {source}; \node[sourceblue] at (.62,-.15) {$\bullet$};
\node[font=\tiny] at (3.12,-.15) {target}; \node[targetorange] at (2.67,-.15) {$\bullet$};
\end{tikzpicture}
\end{minipage}\hfill
\begin{minipage}[t]{0.33\textwidth}
\centering
\resizebox{\linewidth}{!}{%
\begin{tikzpicture}[scale=.80,transform shape,node distance=2.1mm and 3.2mm,every node/.style={font=\scriptsize},box/.style={draw,rounded corners,align=center,minimum height=5.3mm,inner sep=2.1pt},arr/.style={-{Latex[length=1.6mm]},thick}]
\node[font=\small\bfseries] (title) {(c) Deploy on unseen $y^\star$};
\node[box,draw=sourceblue,fill=sourceblue!8,below=4mm of title] (ds) {$D_S^{y^\star}$};
\node[box,right=of ds] (enc) {set\\encoder};
\node[box,right=of enc] (zs) {$z_S$};
\node[box,right=of zs] (map) {$A z_S+b_c$};
\node[font=\tiny,above=2mm of map] (rel) {known relation $c$};
\draw[arr] (rel)--(map);
\node[box,draw=targetorange,fill=targetorange!10,below=5mm of map] (zt) {$\widetilde z_T$};
\node[box,left=of zt] (diff) {shared\\diffusion};
\node[box,left=of diff,draw=targetorange,fill=targetorange!10] (gen) {target\\samples};
\draw[arr] (ds)--(enc); \draw[arr] (enc)--(zs); \draw[arr] (zs)--(map); \draw[arr] (map)--(zt); \draw[arr] (zt)--(diff); \draw[arr] (diff)--(gen);
\node[font=\tiny,below=4mm of diff,align=center] {all headline $K_T$: no target images\\no coordinate or backbone update};
\end{tikzpicture}%
}
\end{minipage}
\caption{\textbf{Meta-Diff learns a relation between task distributions, rather than a separate adaptation for each target.} Meta-training observes the same source$\rightarrow$target relation across different semantic tasks; a shared encoder embeds these distributions in task space; and the learned relation is applied to the source coordinate of a held-out task. At every headline $\KT$, the main route generates directly from the transported coordinate, with no target images, coordinate refinement or backbone update. Target-support budgets apply to independent baselines.}
\label{fig:method}
\end{figure*}
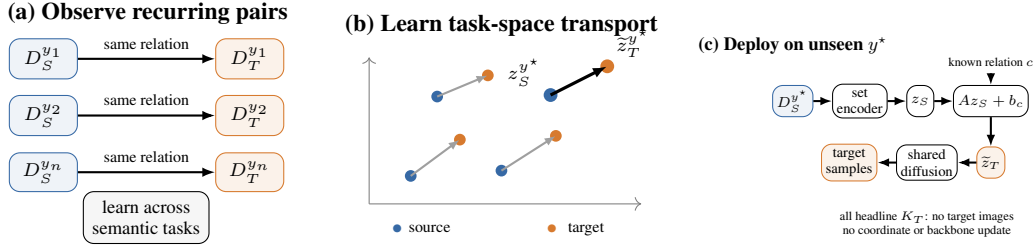

We introduce \emph{Meta-Diff} to exploit this structure (Figure~\ref{fig:method}). A permutation-invariant encoder represents each task distribution by a low-dimensional coordinate. An affine map, learned from earlier source--target pairs, predicts the target coordinate for a new source task. A shared diffusion model then generates from that prediction. Once trained, the route requires no target examples, coordinate refinement or backbone update. Its potential advantage is not simply that the transport map is small, but that it encodes information collected across earlier tasks rather than estimating each new target from scratch.

Whether that information is useful depends on several distinct factors. The source coordinate must retain enough information to distinguish semantic tasks; the learned relation must be predictable across those tasks; transport must preserve task identity; and the shift must be large enough for transport to improve on direct source reuse. Finally, the generator must actually express the transported information in its samples. We establish concentration of source coordinates and stability of an idealised score-based generator under coordinate perturbations, but these guarantees alone do not ensure semantic transfer. We therefore examine the remaining questions experimentally.

We begin with Gaussian mixtures, where source identifiability, relation predictability and transport utility can be studied separately. On held-out CIFAR-100 classes, correct-source transport improves measured distributional distances over wrong-source controls and can outperform target-only adaptation when little target data are available. Yet applying the known corruption directly to clean-source generations performs better on the primary distance metric, and generated class identity remains weak. Office-Home reveals a related limitation under a natural Product$\rightarrow$Clipart shift: target-domain appearance can improve without preserving the requested object. The clinical stress tests likewise provide no resolved source-specific advantage. These findings delimit a useful but narrower claim than general few-shot generative success: task-space relationships can transfer distributional information, while semantic fidelity and the choice of representation remain central obstacles.

\FloatBarrier
\section{Related Work}
\paragraph{Few-shot diffusion and compact adaptation.}
Few-Shot Diffusion Models condition on support sets from unseen classes \citep{giannone2022fsdm}; Conditional Distribution Modelling transfers conditioning statistics \citep{gupta2024cdm}; and CRDI uses diffusion inversion to obtain sample-wise guidance \citep{cao2024crdi}. Personalisation methods adapt pretrained text-to-image generators through full or restricted fine-tuning, learned embeddings and amortised hypernetwork updates \citep{ruiz2023dreambooth,gal2023textualinversion,kumari2023customdiffusion,ruiz2024hyperdreambooth}. LoRA restricts weight updates to a low-rank space \citep{hu2022lora}, while work on customised-model weight spaces suggests that useful adaptation can be represented compactly \citep{dravid2024weights2weights}. These methods reduce the cost of adaptation to a particular target. Our question is different: when several training categories exhibit the same source--target change, can that change be applied to a category for which no target examples are available?

\paragraph{Task representations and cross-domain transfer.}
Meta-learning seeks to reuse information across tasks rather than solve each independently \citep{finn2017maml}. We represent support sets with a permutation-invariant encoder based on Deep Sets \citep{zaheer2017deepsets}, and condition the denoiser through an additive basis or FiLM \citep{perez2018film}. Transfer Guided Diffusion Processes also exploit source information and target-domain guidance \citep{ouyang2024tgdp}. Here the unit of transfer is a relation between distributions: semantic classes are held out at evaluation, while the source--target relation has been observed on other tasks. Its success depends as much on what the task coordinate preserves as on the adaptation mechanism.

\paragraph{Data asymmetry and its limitations.}
Uneven population and institutional coverage can lead to subgroup degradation in medical imaging \citep{larrazabal2020gender,seyyedkalantari2021underdiagnosis,daneshjou2022ddi}; dataset audits also document gaps in demographic coverage and reporting \citep{daneshjou2021transparency,wen2022skindatasets}. These findings motivate better use of related abundant data, but do not imply that synthetic images can replace representative collection. We therefore report the clinical investigations in Appendix~\ref{app:clinical} as stress tests, not evidence for clinical deployment.

\section{Problem and Method}
\label{sec:method}
\textbf{Data-asymmetric task transfer}

Let $y\in\mathcal{Y}$ denote a semantic task and $d\in\{S,T\}$ a source or target domain, with distribution $P_d^y$. Semantic tasks are partitioned into disjoint $\Ttrain,\Tval,\Ttest$. At deployment, the comparison protocol provides a source support set $D_S^y$ of size $\MS$ and permits independent baselines to use a target support set $D_T^y$ of size $\KT$; evaluation uses a disjoint target query set. The correct source $+$ transport main route consumes only $D_S^y$ and the relation descriptor for every $\KT$. We focus on $\MS\gg\KT$.
A shared permutation-invariant encoder maps any support set to a task coordinate,
\begin{equation}
z=r_\psi(D)=\rho_\psi\!\left(\frac{1}{|D|}\sum_{x\in D}h_\psi(x)\right).
\label{eq:encoder}
\end{equation}
The same encoder places source and target distributions in a common task space. Transfer therefore becomes a prediction problem in that space rather than a new high-dimensional optimisation problem for every target.

\textbf{Conditioning and transport}

The coordinate must next control a shared generator. We use standard DDPM noise prediction \citep{ho2020ddpm}, with the continuous-time score interpretation of \citet{song2021score}. The additive-basis parameterisation writes
\begin{equation}
\epshat_z(x_t,t)=\epshat_0(x_t,t)+\sum_{\ell=1}^{k}z_\ell R_\ell(x_t,t),
\label{eq:basis}
\end{equation}
where $R_\ell$ are learned residual noise-prediction fields. To test whether the task-space effect depends on this construction, we also use FiLM. At U-Net block $j$, the feature tensor $H^{(j)}$ is modulated as
\[
H^{(j)}\leftarrow(1+\gamma_j(z))\odot H^{(j)}+\beta_j(z),
\]
with learned projections $\gamma_j,\beta_j$ from the task coordinate to feature-wise scales and shifts.

For a source set, $z_S=r_\psi(D_S^y)$. Let $c=c_{S\rightarrow T}$ index the applicable source--target relation, assumed known at deployment. It selects a relation-specific bias learned during meta-training. We evaluate new semantic tasks under relations observed during meta-training; an unseen relation is outside this protocol. We compare identity, relation-only offsets, residual MLPs and linear maps. In one-seed ablations (Section~\ref{sec:cifar}, Table~\ref{tab:cifar-ablation}), source dependence matters and a nonlinear map does not improve on the linear one, so the final transport is
\begin{equation}
\widetilde z_T=A z_S+b_{c_{S\rightarrow T}},
\label{eq:transport}
\end{equation}
where the relation-specific bias reduces to a single $b$ when only one relation is present. Across all headline $\KT$, the main route directly decodes $\widetilde z_T$ with $J=0$ target-time coordinate-refinement steps and no backbone update. Target-only and full-fine-tuning baselines use their stated target-support budgets. Strict source reuse decodes $z_S$ directly; source-initialised target adaptation is named separately. In the matched refinement ablation (Appendix~\ref{app:cifar-selection}, Table~\ref{tab:cifar-refinement}), 25 refinement steps worsened transformation-statistic distance at $\KT=1$ and $5$ and were unresolved at $\KT=2,10,20$; this separate intervention does not show that refinement is universally ineffective.

\subsection{Meta-training and stability under limited observations}
Each training episode contains source and target support/query sets for one semantic task. With $z_S=r_\psi(D_S^y)$, $z_T=r_\psi(D_T^y)$ and $\widetilde z_T$ from Equation~\eqref{eq:transport}, we minimise
\begin{equation}
\mathcal{L}_{\mathrm{meta}}=\mathcal{L}_{S}(z_S)+\lambda_T\mathcal{L}_{T}(z_T)+\lambda_{\mathrm{tr}}\mathcal{L}_{T}(\widetilde z_T)+\lambda_z(\|z_S\|_2^2+\|z_T\|_2^2),
\label{eq:meta}
\end{equation}
where each $\mathcal{L}$ is the standard diffusion denoising loss evaluated on a disjoint query set. Target supports and queries supervise meta-training through separate branches. The transport map takes only the source coordinate and relation; at deployment the main route reads no target support or query and updates neither coordinates nor backbone weights.

\paragraph{Why task-space transfer can help in the low-data regime.}
A compact coordinate is useful only if abundant source data estimate it reliably and transport errors do not destabilise generation. Two results formalise these requirements.

\begin{proposition}[Source-coordinate concentration]
\label{prop:source-concentration}
For i.i.d.\ $X_1,\ldots,X_M$, let $\mu=\E h_\psi(X)$ and
$z_M=\rho_\psi(\bar h_M)$ with $\bar h_M=M^{-1}\sum_i h_\psi(X_i)$. If
$\E\|h_\psi(X)\|_2^2<\infty$ and $\rho_\psi$ is $L_\rho$-Lipschitz, then
\[
\E\|z_M-\rho_\psi(\mu)\|_2^2
\leq
\frac{L_\rho^2\,\mathrm{tr}(\mathrm{Cov}[h_\psi(X)])}{M}.
\]
\end{proposition}

Thus source-coordinate estimation error decays at the usual $O(M^{-1})$ mean-squared rate. This motivates the CIFAR source-budget check, but does not establish a generated-distance plateau at any particular $\MS$.

\begin{proposition}[Stability to coordinate perturbations]
\label{prop:coordinate-stability}
Write the reverse-time drift in diffusion-time notation as
$v_z(x,t)=f(x,t)-g(t)^2[s_0(x,t)+B(x,t)z]$, where, with $\sigma_t$ the noise scale at time $t$, Equation~\eqref{eq:basis} gives $s_0=-\epshat_0/\sigma_t$ and $B=-[R_1,\ldots,R_k]/\sigma_t$, and $k$ is the coordinate dimension ($k=16$ in the main experiments). In the increasing sampling clock
$u=\tau-t$, the drift is $\bar v_z(x,u)=-v_z(x,\tau-u)$ and the diffusion coefficient is $g(\tau-u)$.
Assume well-posed solutions with finite second moments, a common initial law
$\mu_{\rm init}$ independent of $z$ with $\int\|x\|_2^2\,d\mu_{\rm init}(x)<\infty$,
uniform $L_x$-Lipschitz continuity of $v_z$ in $x$, and
$\sup_{x,t}\|g(t)^2B(x,t)\|_{\rm op}\leq L_z$.
Couple the processes using $X_0^z=X_0^{z'}\sim\mu_{\rm init}$ and the same Brownian path. Then
\[
\E\sup_{u\leq\tau}\|X_u^z-X_u^{z'}\|_2^2
\leq C_\tau\|z-z'\|_2^2,
\qquad
W_2(P_z,P_{z'})\leq\sqrt{C_\tau}\|z-z'\|_2,
\]
where $P_z$ denotes the law of $X_\tau^z$, $C_\tau=L_z^2\phi_{L_x}(\tau)^2$ and
$\phi_{L_x}(\tau)=(e^{L_x\tau}-1)/L_x$ (with value $\tau$ at $L_x=0$).
\end{proposition}

These results establish coordinate-estimation and continuous-time stability properties under the stated assumptions. The second proposition concerns the additive-score-conditioned process with uniform Lipschitz bounds; it does not certify a small practical error for the trained DDIM sampler or automatically cover a general FiLM-conditioned denoiser. Neither proposition establishes semantic task identity, relation predictability on unseen tasks, or semantic use by the generator. Section~\ref{sec:gmm} tests task identity and relation predictability, and Section~\ref{sec:office} probes semantic use by the generator; full proofs are in Appendix~\ref{app:theory}.

\section{Experimental Design}
\label{sec:experiments}
The propositions establish estimation and stability, not content or relational generalisation. Experiments therefore progress from GMMs, where these missing quantities are observable, to CIFAR under controlled image shifts and finally to the natural Product$\rightarrow$Clipart relation in Office-Home. Headline comparisons are paired on semantic task, support draw, target query set and sampling noise where possible. Generation repeats are not independent semantic tasks. CIFAR generation intervals use 10,000 paired class-cluster bootstrap resamples (seed 20261020): each draw samples 20 semantic classes with replacement, retaining all three corruptions within a class. We average corruptions within class, then classes, and report percentile 95\% intervals. Stars mark unrounded paired intervals excluding zero; intervals are pointwise, without multiplicity adjustment. These fixed-checkpoint intervals do not quantify training-seed variation. Because the 20 held-out fine classes belong to only four superclasses, the intervals also do not capture variation across superclasses. Other benchmarks retain their separately specified statistical units. Hyperparameters for target-only adaptation and fine-tuning are selected on validation tasks and frozen before final test evaluation.

\paragraph{Controlled GMM boundary study.}
Two-dimensional Gaussian-mixture tasks vary source-task separation through $\alpha\in\{0,.1,.25,.5,.75,1,1.5\}$ and relation heterogeneity through $\eta\in\{0,.1,.25,.5,1,2\}$. The full $7\times6$ grid is repeated over five seeds, giving 210 trained cells and 13,440 held-out task evaluations at $\KT=0$. A separate shift-magnitude sweep at $\alpha=1,\eta=0$ contributes a further 25 cells and 1{,}600 held-out task evaluations. The primary effect is
\[
G_{\mathrm{source}}=\mathrm{SW}_{\mathrm{wrong}}-\mathrm{SW}_{\mathrm{correct}},
\]
so positive values favour the correct source. Because the data-generating distributions are known, intrinsic Bayes task identification can be measured separately from task identification in the learned coordinate and from held-out affine relation predictability.

\paragraph{CIFAR-100.}
Fine classes are semantic tasks \citep{krizhevsky2009}. Entire superclasses are held out: 60 fine classes from 12 superclasses for meta-training, 20 from four superclasses for validation, and 20 from four disjoint superclasses for final testing. Sources are clean; targets apply Gaussian blur, Gaussian noise or contrast reduction at fixed severity \citep{hendrycks2019}. Source and target pools contain disjoint underlying images. We test zero-target source controls and $\KT\in\{1,5,20\}$ against target-only modelling and validation-selected full fine-tuning. Sliced Wasserstein distance (SW) is the headline distribution metric; energy MMD and class consistency (Appendix~\ref{app:cifar}) are complementary. The appendix retains separate GMM denoising-calibration and historical optimised-coordinate diagnostics; these are not pooled with the image-generation metrics.

\paragraph{Office-Home.}
Product$\rightarrow$Clipart provides a natural relation across 65 object categories. The fixed semantic split is 45/10/10 train/validation/test classes. Real-image audits verify both the semantic evaluator and the Product--Clipart domain distinction before generated images are scored. We first evaluate the same source controls used on CIFAR, then use frozen-checkpoint diagnostics to probe possible semantic limitations in the encoder, transport and coordinate-to-generation path.

\section{Results}
\label{sec:results}
\subsection{GMM: a controlled operating regime}
\label{sec:gmm}
\begin{figure*}[!htbp]
\centering
\begin{minipage}[t]{0.49\textwidth}\centering
\includegraphics[width=\linewidth]{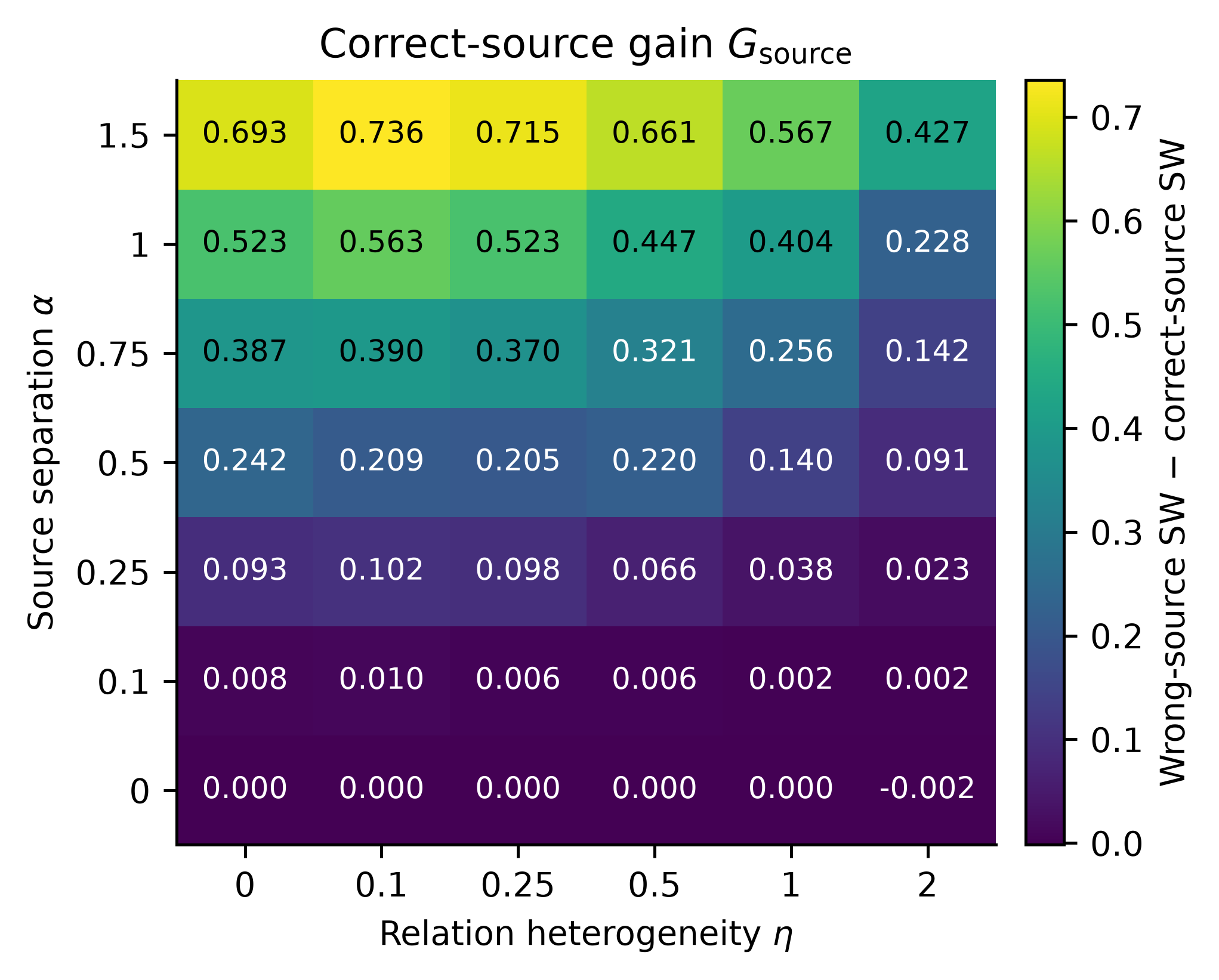}
\end{minipage}\hfill
\begin{minipage}[t]{0.47\textwidth}\centering
\includegraphics[width=\linewidth]{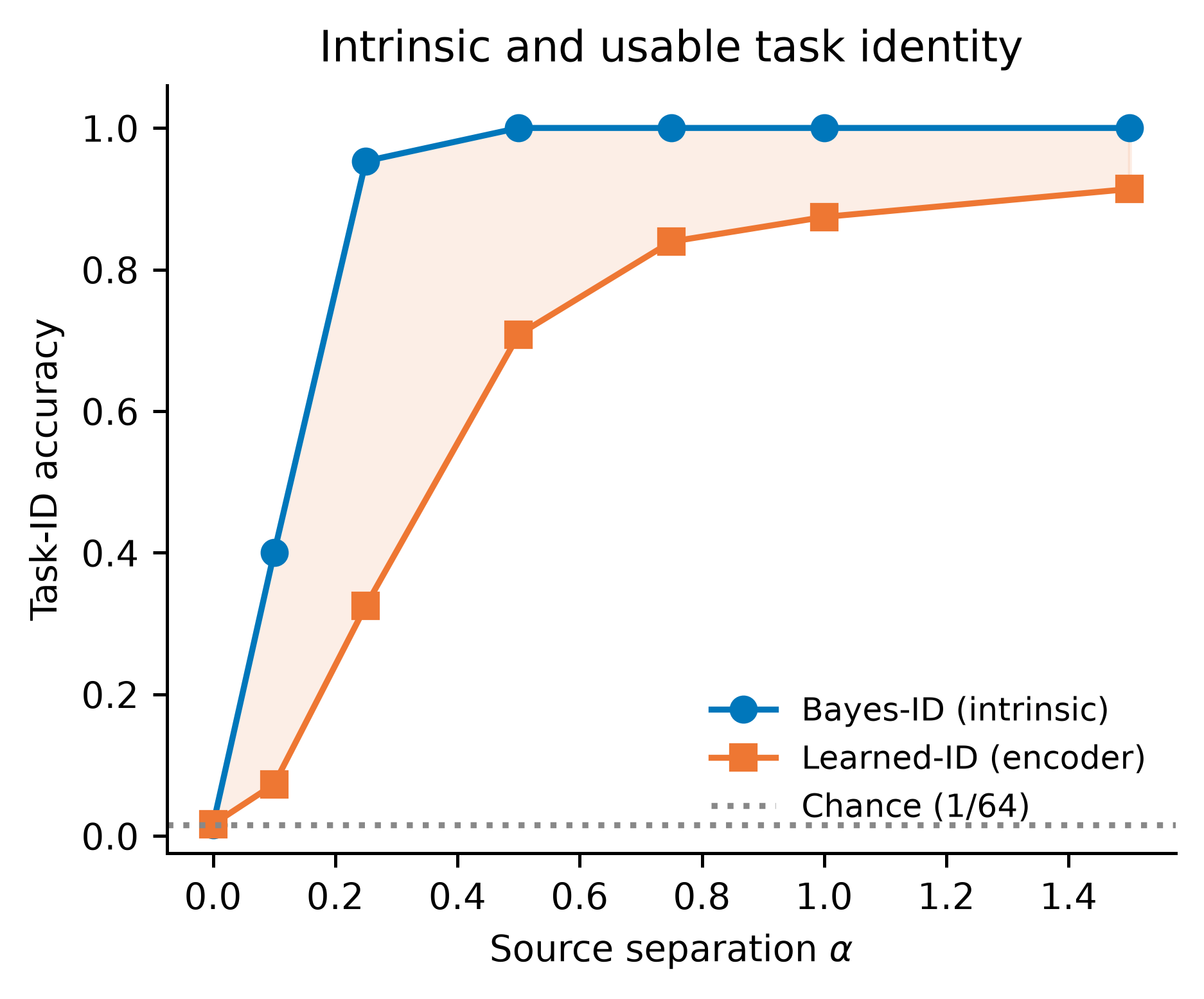}
\end{minipage}
\caption{\textbf{GMM operating regime.} \emph{Left:} correct-source advantage $G_{\mathrm{source}}=\mathrm{SW}_{\mathrm{wrong}}-\mathrm{SW}_{\mathrm{correct}}$ over the full source-identifiability/relation-heterogeneity grid at $\KT=0$. The advantage disappears when source identity is absent and attenuates as the relation becomes less predictable. \emph{Right:} intrinsic Bayes task identification and nearest-centroid identification in the learned encoder coordinate, averaged over relation levels. Nearly perfect intrinsic identifiability can coexist with substantially weaker usable identity. Full cell-level results and paired intervals are in Appendix~\ref{app:gmm}.}
\label{fig:gmm-main}
\end{figure*}

The analytical results guarantee a stable coordinate, not a useful one. GMMs make the missing ingredients observable: we vary source-task separation $\alpha$ and relation heterogeneity $\eta$ independently, then evaluate zero-target transfer over the resulting $7\times6$ grid. Figure~\ref{fig:gmm-main} shows the first two conclusions. When $\alpha=0$, correct and wrong sources are unresolved in all six relation conditions; once the source carries task information, a correct-source advantage emerges, but it decreases as the relation becomes less reusable. In total, 34 of 42 cells have positive paired intervals excluding zero, with the unresolved cells confined to the six $\alpha=0$ conditions and the two most heterogeneous relations at $\alpha=.1$. Transport is not uniformly better than every alternative: at $\eta\geq1$ the abundant-target reference has lower SW in all 14 cells, and source reuse beats transport at $(\alpha,\eta)=(1.5,2)$ (Table~\ref{tab:gmm-sw-paired-references}).

The same experiment distinguishes \emph{intrinsic} from \emph{usable} identifiability. At $\alpha=.25$, the Bayes classifier on the true GMM likelihoods already identifies the source task with accuracy $.953$, yet nearest-centroid identification in the learned coordinate is only $.325$. At $\alpha=1.5$, the corresponding values are $1.000$ and $.914$. Across the 42 cells, $G_{\rm source}$ correlates much more strongly with learned-ID than with Bayes-ID ($\rho=.98$ versus $.89$). Replacing Bayes-ID with learned-ID in the same descriptive interaction model raises the cell-level fit from $R^2=.587$ to $.843$; even within the 24 cells with Bayes-ID $>.99$, learned-ID remains strongly associated with the source-specific gain ($\rho=.97$). We treat these regressions as descriptive summaries of the factorial experiment rather than causal estimates.

Relation predictability is a separate requirement. Along $\alpha=1$, Bayes-ID remains at $1.0$ while increasing $\eta$ from 0 to 2 lowers held-out affine relation $R^2$ from $1.000$ to $.408$ and reduces the correct-over-wrong SW advantage from $.523$ to $.228$. Across the full grid, $\rho(G_{\rm source},R^2_{\rm rel})=.54$. Thus source identifiability alone is insufficient: even when the source task is intrinsically unambiguous, source-specific transfer weakens as the relation becomes less reusable. Full diagnostics are in Appendix~\ref{app:gmm}.

\begin{figure*}[!htbp]
\centering
\begin{minipage}[t]{0.49\textwidth}\centering
\includegraphics[width=.8\linewidth]{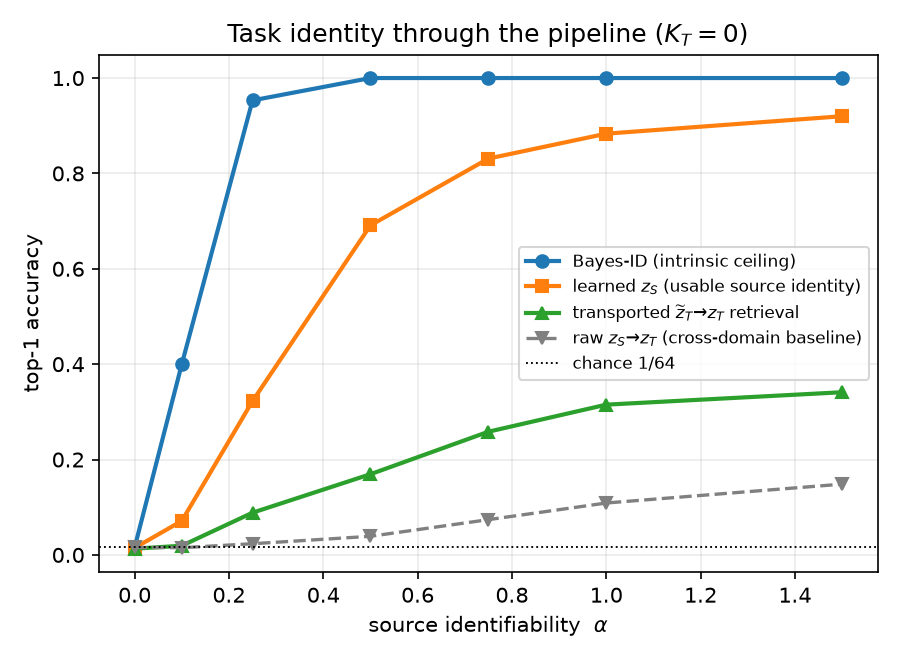}
\end{minipage}\hfill
\begin{minipage}[t]{0.49\textwidth}\centering
\includegraphics[width=.8\linewidth]{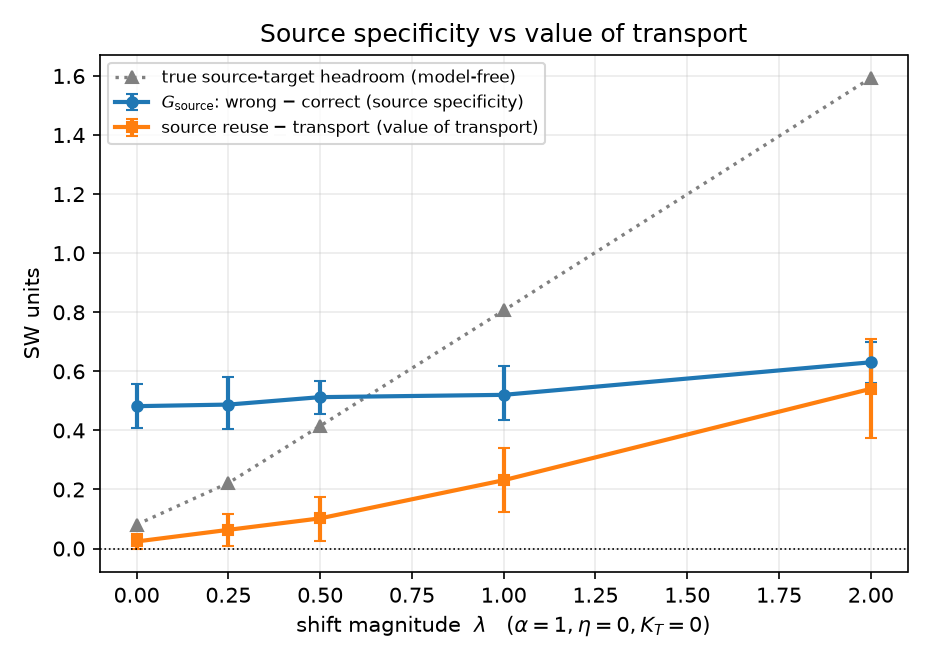}
\end{minipage}
\caption{\textbf{Two additional GMM boundaries.} \emph{Left:} task identity through the pipeline. Bayes-ID is the intrinsic ceiling, learned $z_S$ measures usable source identity, and transported $\widetilde z_T$ measures whether the predicted target coordinate retrieves the correct held-out target task; raw $z_S\!\to z_T$ retrieval is a cross-domain baseline. \emph{Right:} the shift-magnitude sweep separates source specificity from the value of transport, with hierarchical seed/task 95\% intervals. The correct source remains informative as $\lambda\!\to\!0$, whereas the advantage of transport over source reuse grows only when there is nontrivial source--target headroom. Exact means are in Tables~\ref{tab:gmm-identity-chain} and \ref{tab:gmm-shift-means}; the intervals drawn in the right panel are not tabulated.}
\label{fig:gmm-new-mechanism}
\end{figure*}

The coordinate-retrieval analysis makes information loss along the pipeline explicit. For $\alpha\geq.5$, intrinsic identity is perfect, yet learned-$z_S$ retrieval ranges from $.691$ to $.920$ and transported target-coordinate retrieval from only $.169$ to $.341$ (chance $1/64$). Averaged over the grid, transported retrieval is $.172$ versus $.061$ for the raw cross-domain $z_S\!\to z_T$ baseline; the overall transport gain is $+.112$ (paired seed-block interval $[+.106,+.116]$); mean gains are near zero at $\alpha\in\{0,.1\}$ and positive for $\alpha\geq.25$ (cellwise intervals were not computed). Across the grid it correlates with $G_{\rm source}$ at $\rho=.91$, between Bayes-ID ($.89$) and learned-ID ($.98$). Retrieval margins show that the recovery is partial: the learned-$z$ margin rises from $-.002$ to $+.060$ across the sweep, whereas the transported margin remains negative ($-.039$ to $-.079$). Figure~\ref{fig:gmm-new-mechanism} (left) and Appendix~\ref{app:gmm} report the full identity chain.

The shift-magnitude sweep in Figure~\ref{fig:gmm-new-mechanism} (right) isolates a different question: \emph{when is transport itself needed?} We fix $\alpha=1,\eta=0$ and interpolate the same source--target transformation by $\lambda$. The correct-source advantage over a wrong source remains large even at $\lambda=0$ ($G_{\rm source}=.482$), because source identity is still recoverable. In contrast, transport improves over direct source reuse by a mean of only $.024$ SW when there is essentially no shift to correct, but by $.541$ at $\lambda=2$; the independently measured source--target headroom increases from $.08$ to $1.59$ SW over the same sweep. Source specificity and transport utility are therefore different quantities: identifying the right source is not enough to make a transport map useful if source and target are already nearly the same.

The controlled study therefore identifies four empirical conditions, associated with source-specific gains, beyond the two analytical guarantees: task identity must survive the encoder, the relation must generalise, transport must preserve enough identity to reach the target task, and the source--target discrepancy must leave useful headroom over source reuse. CIFAR now asks whether this mechanism survives when the distributions are images rather than known mixtures.

\FloatBarrier
\subsection{CIFAR: source specificity and target-budget comparisons}
\label{sec:cifar}
\begin{figure*}[!htbp]
\centering
\includegraphics[width=.7\linewidth]{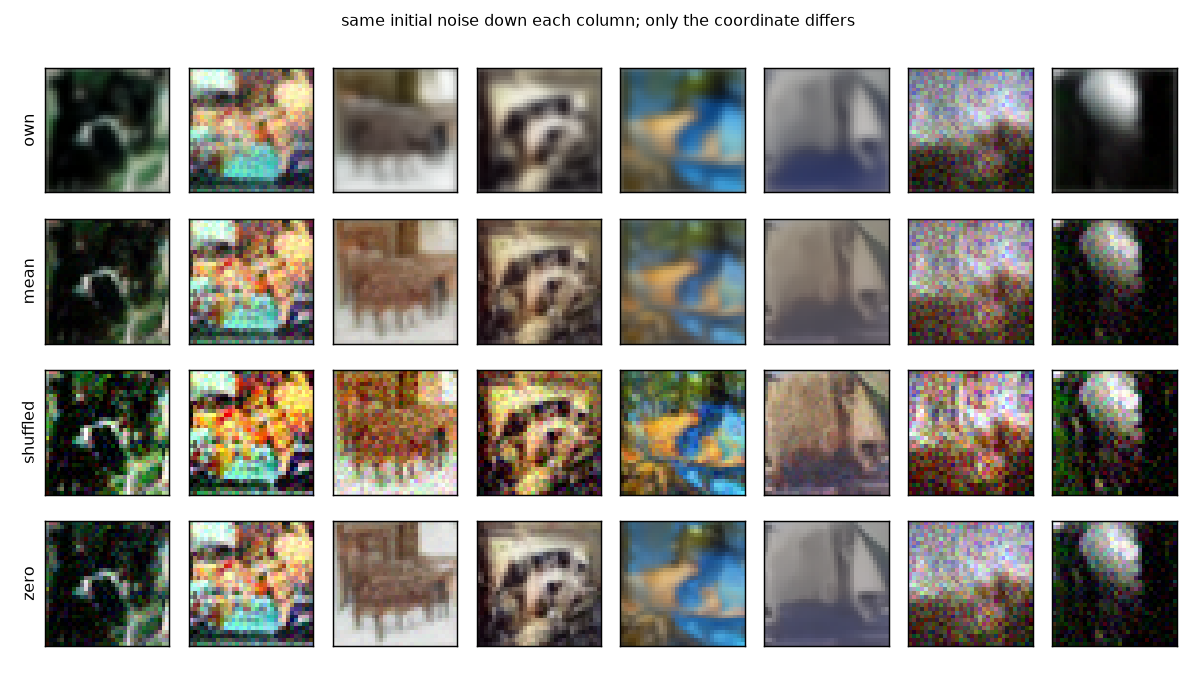}
\caption{\textbf{Qualitative task-coordinate control on held-out CIFAR generations.} Columns share initial diffusion noise; rows differ only in the task coordinate (own, mean, shuffled, or $z=0$). The samples come from the 50k-step CIFAR checkpoint, with matched tasks and paired noise. Own-task coordinates are encoder coordinates, not transported predictions; deployment claims use the separate paired route evaluations in Section~\ref{sec:cifar}.}
\label{fig:cifar-qualitative-early}
\end{figure*}

We first ask whether choosing the correct source class changes the generated target distribution in the expected direction. On 60 held-out validation class--corruption episodes, correct-source transport achieves SW $0.2061$, compared with $0.2393$ using a wrong source and $0.2407$ using only the relation (Table~\ref{tab:cifar-main}). The paired SW and class-consistency differences favour the correct source, although class consistency is not distinguishable from strict source reuse. A separate frozen-coordinate diagnostic on a FiLM checkpoint supports this source dependence before generation: transported coordinates retrieve the correct held-out target class in $62.5\%$, $59.5\%$ and $49.0\%$ of cases for blur, noise and contrast, respectively, versus $17.25\%$, $26.5\%$ and $14.0\%$ for raw source coordinates and at most $2.25\%$ for wrong-source transport (Appendix~\ref{app:cifar}). The separate $K_T=1$ deployment grid in Figure~\ref{fig:cifar-strategy-grid} shows six generated routes and a real-target row with paired generation noise; it should not be conflated with either of these diagnostics. Figure~\ref{fig:cifar-qualitative-early} instead shows how changing the task coordinate affects generation with diffusion noise held fixed, not whether transport succeeds.

The final-test comparison asks whether the target-free route is useful when alternatives can adapt using one, five or twenty target images. At $K_T=1$, mean SW is $0.1781$ for transport, $0.2087$ for target-only coordinate adaptation and $0.2648$ for full fine-tuning. Transport has lower SW than both under the paired class-cluster intervals. Its advantage over full fine-tuning persists at $K_T=5$, but by $K_T=20$ full fine-tuning is better ($0.1483$ versus $0.1844$). The transport--target-only difference is unresolved at $K_T=5$ and $20$. These comparisons use validation-selected adaptation settings that vary with target budget, including both optimisation steps and learning rates. They therefore do not isolate the effect of target-set size. Energy MMD and the paired contrasts are reported in Appendix~\ref{app:cifar}. All full-FT settings were frozen before final-test evaluation. The appendix also contains a separate matched refinement on/off intervention on the transport checkpoint; refinement is not used by the headline route.

\begin{table}[!htbp]
\centering\footnotesize\setlength{\tabcolsep}{5pt}
\caption{\textbf{CIFAR headline comparisons for correct source $+$ transport, $J=0$.} Panel A uses validation classes; Panel B uses separate final-test classes. Distances: $\Delta=\mathrm{comparator}-\mathrm{main}$; consistency: $\Delta=\mathrm{main}-\mathrm{comparator}$. Positive favours the main route. Entries show mean [95\% class-cluster CI]; stars use unrounded paired intervals excluding zero, in either direction.}
\label{tab:cifar-main}
\begin{tabular}{@{}lcccc@{}}
\toprule
\multicolumn{5}{l}{\textbf{A. Zero-target controls: 60 validation episodes}}\\
Route & SW & $\Delta$ SW & Class consistency & $\Delta$ class\\
\midrule
\shortstack[l]{correct source\\$+$ transport} & \shortstack{$0.2061$\\$[0.1789,0.2392]$} & -- & \shortstack{$0.1075$\\$[0.0740,0.1479]$} & --\\
Strict source reuse & \shortstack{$0.2416$\\$[0.2169,0.2703]$} & \shortstack{$0.0355\sigstar$\\$[0.0283,0.0420]$} & \shortstack{$0.1102$\\$[0.0726,0.1554]$} & \shortstack{$-0.0028$\\$[-0.0160,0.0083]$}\\
Wrong source & \shortstack{$0.2393$\\$[0.2085,0.2737]$} & \shortstack{$0.0333\sigstar$\\$[0.0163,0.0525]$} & \shortstack{$0.0470$\\$[0.0366,0.0582]$} & \shortstack{$0.0604\sigstar$\\$[0.0297,0.0977]$}\\
Relation-only & \shortstack{$0.2407$\\$[0.2115,0.2742]$} & \shortstack{$0.0346\sigstar$\\$[0.0224,0.0477]$} & \shortstack{$0.0490$\\$[0.0382,0.0608]$} & \shortstack{$0.0585\sigstar$\\$[0.0293,0.0932]$}\\
\midrule
\multicolumn{5}{l}{\textbf{B. Final-test SW: 180 episodes, 60 per budget}}\\
Route & $\KT=1$ & $\KT=5$ & $\KT=20$ & \\
\midrule
\shortstack[l]{correct source\\$+$ transport} & \shortstack{$0.1781$\\$[0.1627,0.1959]$} & \shortstack{$0.1834$\\$[0.1691,0.1983]$} & \shortstack{$0.1844$\\$[0.1676,0.2030]$} & \\
Target-only & \shortstack{$0.2087$\\$[0.1869,0.2337]$} & \shortstack{$0.1884$\\$[0.1708,0.2078]$} & \shortstack{$0.1799$\\$[0.1677,0.1935]$} & \\
Full fine-tuning & \shortstack{$0.2648$\\$[0.2380,0.2927]$} & \shortstack{$0.2205$\\$[0.2028,0.2386]$} & \shortstack{$0.1483$\\$[0.1395,0.1580]$} & \\
$\Delta$ target-only & \shortstack{$0.0306\sigstar$\\$[0.0152,0.0467]$} & \shortstack{$0.0051$\\$[-0.0028,0.0122]$} & \shortstack{$-0.0044$\\$[-0.0132,0.0036]$} & \\
$\Delta$ full FT & \shortstack{$0.0867\sigstar$\\$[0.0623,0.1108]$} & \shortstack{$0.0371\sigstar$\\$[0.0151,0.0597]$} & \shortstack{$-0.0361\sigstar$\\$[-0.0588,-0.0150]$} & \\
\bottomrule
\end{tabular}
\end{table}

Further experiments test whether source dependence survives changes to the conditioning architecture and transport map. In a separate FiLM validation package, correct-source transport improves SW and class consistency against mean-source, shuffled-source and relation-only controls at every tested budget. This supports source specificity under FiLM, but is not a FiLM full-fine-tuning crossover. Direct matched FiLM--basis comparisons do not resolve a SW ranking. Replacing the linear map with a residual MLP likewise yields no resolved SW difference for the basis model, whereas FiLM with linear transport has lower SW than its nonlinear alternative at all three budgets and lower energy MMD at $K_T=1,5$ (Table~\ref{tab:cifar-ablation}). These are separate, one-training-seed comparisons. The basis and FiLM conditioning modules have 55,344 and 118,784 parameters, respectively; the counts do not establish a runtime or memory advantage.

The stronger source dependence in distributional metrics should not be confused with reliable generation of the requested class. In the headline validation study, class consistency is $0.1075$ for transport and $0.1102$ for strict source reuse, with an unresolved paired difference. A separate direct-$z_T$ diagnostic on the FiLM checkpoint gives held-out generated-class identity of only $3.59\%$, $3.85\%$ and $2.71\%$ for blur, noise and contrast, versus $51.20\%$, $34.50\%$ and $70.90\%$ on its real queries. That diagnostic uses a different support/query partition and semantic evaluator, so its scores are not directly comparable with headline consistency (Appendix~\ref{app:cifar}).

\begin{figure*}[!htbp]
\centering
\begin{minipage}[t]{0.49\textwidth}\centering
\includegraphics[width=\linewidth]{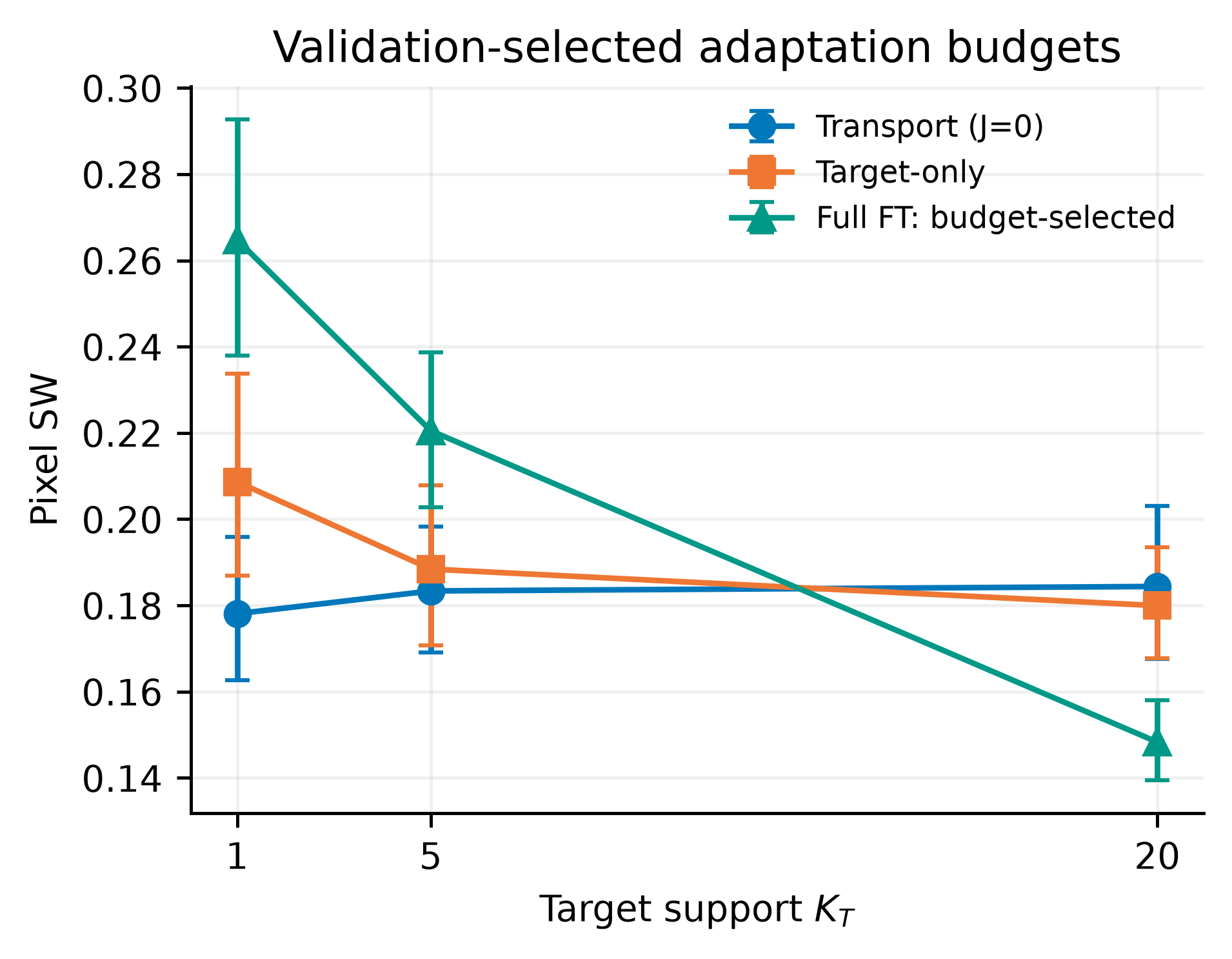}
\end{minipage}\hfill
\begin{minipage}[t]{0.49\textwidth}\centering
\includegraphics[width=\linewidth]{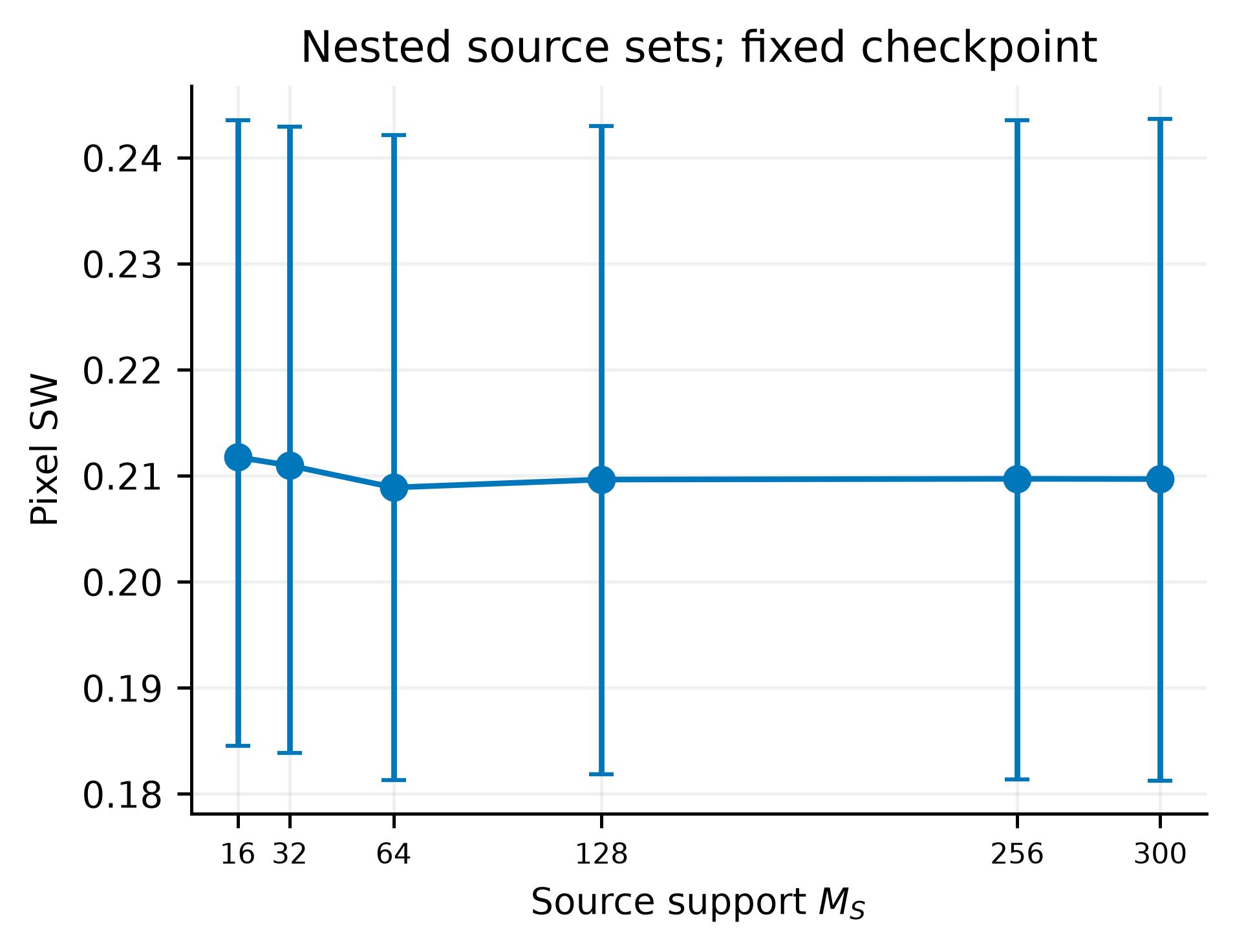}
\end{minipage}
\caption{\textbf{CIFAR target and source budgets.} Left: final-test SW for correct source $+$ transport and target-adapted baselines. The main route uses no target support at any $K_T$; its small changes across budgets reflect different sampled source/query episodes and generation noise. Right: validation source-budget sweep ($K_T=1$, nested source supports, checkpoint trained at $M_S=64$). Error bars show marginal class-cluster percentile 95\% intervals; paired effects and stars are computed separately.}
\label{fig:cifar-budgets}
\end{figure*}

Finally, changing source-set size between $M_S=16$ and 300 produces only small SW differences, all unresolved relative to $M_S=16$ under the paired class-cluster intervals. At $M_S=64$, the difference is $-0.00287\,[-0.00655,0.00089]$. This does not demonstrate an exact plateau or equivalence, and smaller source sets were not tested. Figure~\ref{fig:cifar-budgets} combines the final-test target-budget comparison with the separate validation source-budget sweep. Additional source controls, MMD results, adaptation settings, refinement and task-scarcity experiments appear in Appendix~\ref{app:cifar}.

\begin{table}[!htbp]
\centering\footnotesize
\caption{\textbf{Three separate matched CIFAR packages.} The first row uses the two conditioning runs of correct source $+$ transport; the other rows compare transport structures. Each evaluated transport route uses $J=0$. Signs follow the first column; entries show mean [95\% paired class-cluster CI] and stars use unrounded intervals. One training seed per arm.}
\label{tab:cifar-ablation}
\begin{tabular}{lccc}
\toprule
Paired SW comparison & $\KT=1$ & $\KT=5$ & $\KT=20$\\
\midrule
FiLM $-$ basis & \shortstack{$-0.0094$\\$[-0.0238,0.0046]$} & \shortstack{$-0.0148$\\$[-0.0298,0.0003]$} & \shortstack{$-0.0149$\\$[-0.0300,0.0003]$}\\
MLP $-$ linear (basis) & \shortstack{$-0.0022$\\$[-0.0080,0.0031]$} & \shortstack{$-0.0004$\\$[-0.0068,0.0056]$} & \shortstack{$-0.0036$\\$[-0.0097,0.0019]$}\\
MLP $-$ linear (FiLM) & \shortstack{$0.0093\sigstar$\\$[0.0044,0.0147]$} & \shortstack{$0.0112\sigstar$\\$[0.0057,0.0175]$} & \shortstack{$0.0087\sigstar$\\$[0.0026,0.0151]$}\\
\bottomrule
\end{tabular}
\end{table}

\FloatBarrier
\paragraph{Additional matched controls.}
Because CIFAR's shifts are known pixel transformations, we also test generating from the clean-source coordinate and applying the specified corruption afterwards. On matched test episodes, this simple route achieves SW $0.1688$, compared with $0.1781$ for transport (paired baseline-minus-transport difference $-0.0092\sigstar\,[-0.0138,-0.0045]$). It uses no target images and gives a better distributional match here; task-space transport is not preferable to this known image-space operation (Appendix~\ref{app:new-experiments}, Table~\ref{tab:new-image-baseline}).

To separate target-support size from the number of fine-tuning updates, another experiment fixes full fine-tuning at 100 updates, learning rate $10^{-5}$ and 20 noised examples per update for each $K_T$, selecting the schedule once on validation classes. Full-FT SW falls from $0.3054$ at $K_T=1$ to $0.2123$ at $K_T=5$ and $0.1709$ at $K_T=20$; the paired $20$-minus-$1$ effect is $-0.1345\sigstar\,[-0.1730,-0.0957]$. Under this fixed schedule, transport has lower SW at $K_T=1,5$, but the full-FT-minus-transport difference at $K_T=20$ is unresolved ($-0.0119\,[-0.0429,0.0213]$). This controls the update budget within full fine-tuning, not its compute relative to zero-update transport (Appendix~\ref{app:new-experiments}, Tables~\ref{tab:new-budget-selection}--\ref{tab:new-fixed-effects}).

The matched semantic evaluation gives a more sobering picture. A frozen DINO-100 prototype classifier assigns only $2.53\%$, $2.85\%$ and $2.34\%$ of transport generations to the requested class at $K_T=1,5,20$, compared with $50.49\%$, $49.79\%$ and $50.65\%$ on the real queries. Full fine-tuning reaches $3.68\%$, $9.58\%$ and $10.50\%$, higher than transport under the paired intervals; transport--target-only differences remain unresolved. This instrument uses 100 source-derived class prototypes and is separate from the headline class-consistency metric (Appendix~\ref{app:new-experiments}, Table~\ref{tab:new-cifar-semantics}).

\subsection{Office-Home: natural-domain alignment and unresolved semantic limitations}
\label{sec:office}
CIFAR shows that transport can make use of class-specific source information under controlled image shifts, even without reliable class-correct generation. Office-Home offers a less prescribed test: can a learned Product$\rightarrow$Clipart relation transfer to previously unseen object categories? We first check the evaluators on real images. CLIP recognises held-out Clipart objects with $65.96\%$ top-1 accuracy, DINOv2 distinguishes Product from Clipart with $98.30\%$ accuracy, and same-class cross-domain centroid retrieval achieves $100\%$ top-1 with median rank 1. These checks establish that the evaluators can detect the object and domain distinctions relevant to the test.

\begin{figure*}[!htbp]
\centering
\begin{minipage}[t]{0.48\textwidth}\centering
\includegraphics[width=\linewidth]{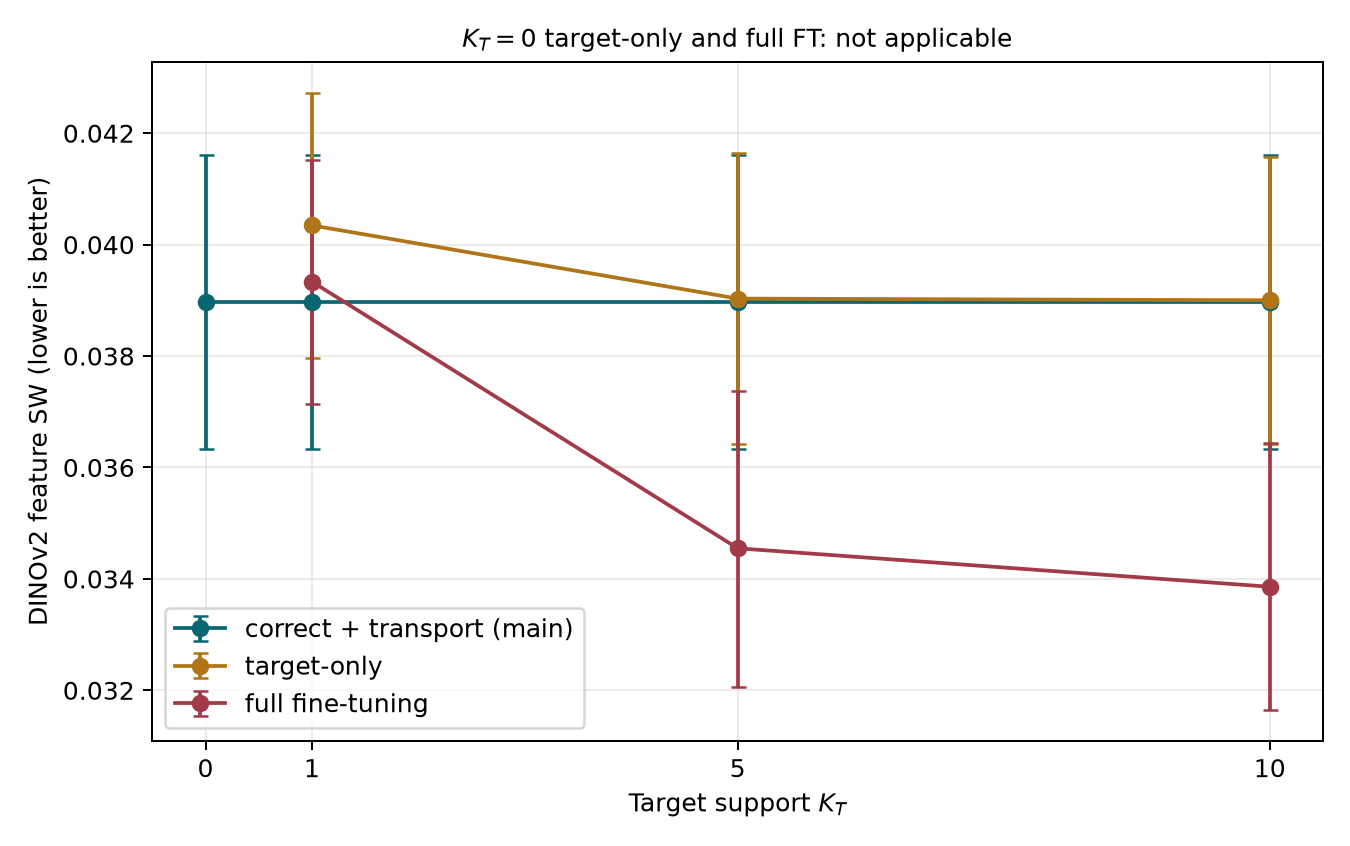}
\end{minipage}\hfill
\begin{minipage}[t]{0.50\textwidth}\centering
\includegraphics[width=\linewidth]{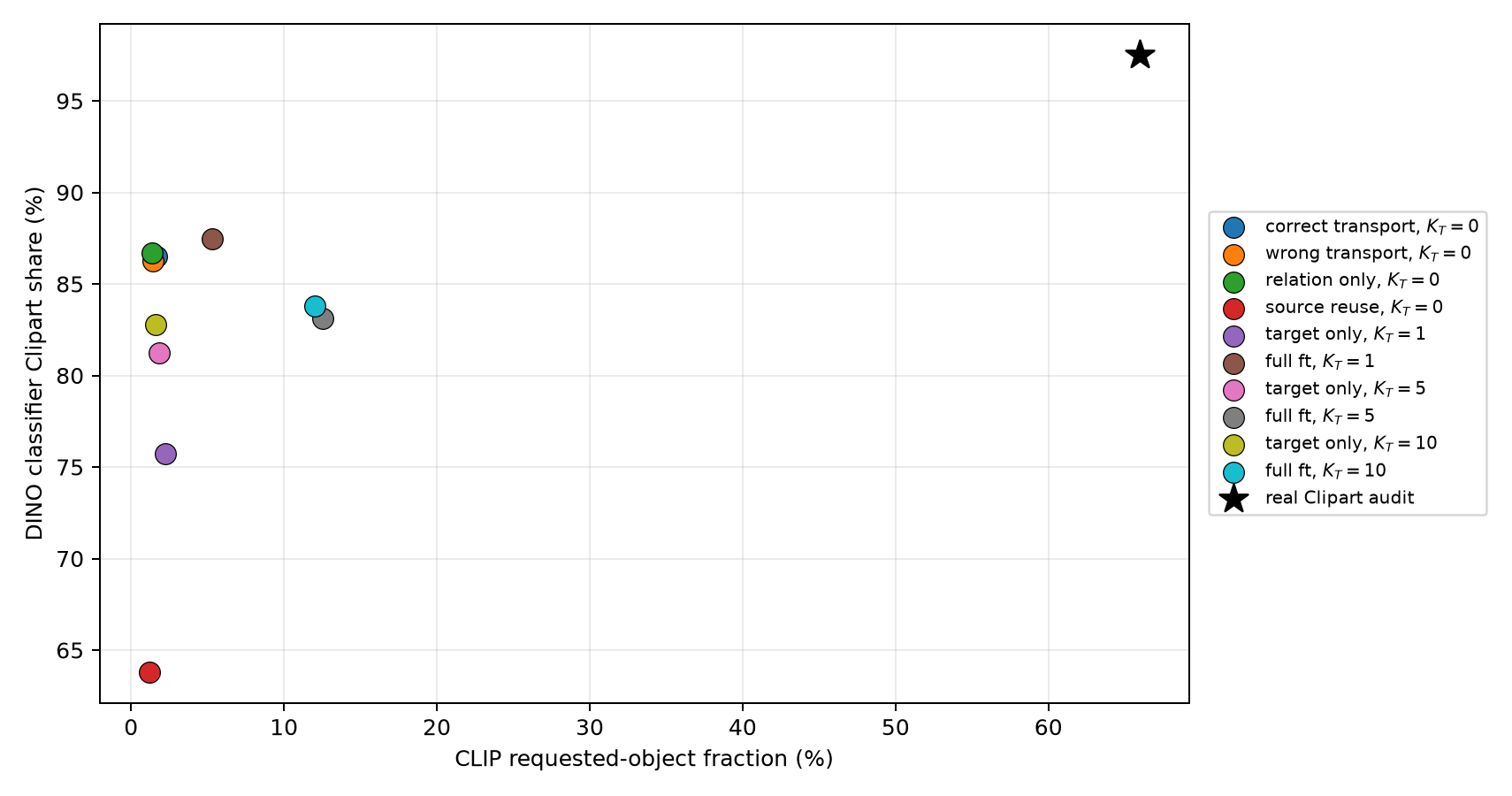}
\end{minipage}
\caption{\textbf{Office-Home exposes a different boundary from CIFAR.} \emph{Left:} correct-source transport is flat in DINO feature-SW because the same zero-target generations are reused at every $\KT$, while full fine-tuning improves as target support grows. \emph{Right:} generated routes can become strongly Clipart-like without preserving the requested object; the star is the held-out real Clipart audit. The natural-domain limitation is therefore semantic, not simply a failure to change domain appearance.}
\label{fig:office-summary}
\end{figure*}

Transport does change the measured domain statistics of generated images. At $K_T=0$, correct-source transport has DINO feature-space SW $.038969$, versus $.039394$ for wrong-source transport, $.039251$ for relation-only prediction and $.041641$ for strict source reuse; all three paired differences exclude zero. Predicted Clipart share rises from $63.8\%$ under source reuse to $86.5\%$ under transport, although wrong-source and relation-only transport produce a similar rise (Table~\ref{tab:office-semantic-paired}). Requested-object identity remains only $1.67\%$, with unresolved differences against wrong-source and relation-only controls. At one target image, transport improves on target-only adaptation in DINO-SW and energy MMD, but its SW difference from full fine-tuning is unresolved. By $K_T=5$ and $10$, full fine-tuning has lower SW under paired intervals (Figure~\ref{fig:office-summary}; Table~\ref{tab:office-distance-paired}).

These results leave two possible sources of error: the model may fail to transport the source coordinate accurately, or the target coordinate may not control the generator strongly enough to preserve object identity. Frozen-checkpoint diagnostics help distinguish them, though not completely. Across 20 resamples of two disjoint 16-image Product support sets per class, learned source coordinates retrieve the correct class among ten held-out candidates with $91.0\%$ $L_2$ top-1 accuracy. Transport raises cross-domain target-coordinate retrieval from $12.0\%$ to $42.0\%$ $L_2$ top-1, although the correct-versus-wrong transport interval remains unresolved across the ten classes.

Bypassing transport does not restore semantic fidelity. A direct target coordinate computed from abundant Clipart support gives only $2.38\%$ requested-class accuracy for seen classes and $1.56\%$ for held-out classes, despite real-query accuracies around $50\%$; some seen-class queries may have appeared in training (Appendix~\ref{app:office}). Correct rather than wrong coordinates slightly reduce denoising MSE across diffusion time, so the generator does respond to the coordinate, but weakly. A separate explicit-class-embedding model reaches $6.90\%$ generated identity on seen classes, still far below the real-image reference. Transport error alone therefore cannot explain the poor object identity. The remaining limitation could lie in the target-coordinate representation, the way the denoiser uses it, or both. The tests do not uniquely localise the problem to the decoder. Complete route matrices, frozen-model diagnostics and validation-only checkpoint selection are in Appendix~\ref{app:office} (Tables~\ref{tab:office-budget-complete}--\ref{tab:office-semantic-paired}, \ref{tab:office-retrieval-paired}--\ref{tab:office-capacity-paired} and \ref{tab:office-selection}).

\FloatBarrier
A final matched comparison uses identical queries and the same CLIP evaluator for transport, target-only adaptation and full fine-tuning. At $K_T=5$, requested-class identity is $1.67\%$, $1.81\%$ and $12.55\%$, respectively, against $52.00\%$ on those real queries. The transport-minus-full-FT difference is $-10.89\sigstar\,[-19.82,-4.14]$ percentage points. Appendix~\ref{app:new-experiments} reports all budgets and the real-query calibration.

\section{Discussion and Limitations}
\label{sec:synthesis}
The experiments make a distinction that is easy to miss: learning a useful source--target relation is not the same as generating the correct object in the target domain. In the Gaussian mixtures, known task distributions expose why transport succeeds or fails. On CIFAR, the correct source improves measured distributional distances, but does not ensure class-correct generations. Office-Home makes this separation clearer still: outputs become more Clipart-like without reliably depicting the requested object. Under their stated assumptions, Propositions~\ref{prop:source-concentration}--\ref{prop:coordinate-stability} establish concentration of source-coordinate estimates and stability of the idealised additive-score process, but neither result guarantees semantic transfer. Table~\ref{tab:mechanism-synthesis} collects the relevant empirical diagnostics without treating their different metrics as interchangeable.

\begin{table}[!htbp]
\centering\scriptsize
\caption{\textbf{Stages of source-informed transfer.} Representative diagnostics associated with transfer and the stage at which it becomes limiting; values are not intended as a common metric across rows.}
\label{tab:mechanism-synthesis}
\begin{tabular}{@{}p{.14\linewidth}p{.42\linewidth}p{.35\linewidth}@{}}
\toprule
Setting & Representation / transport evidence & Generative implication\\
\midrule
GMM & Bayes-ID can be perfect while learned-ID is weaker; at $\alpha=1$, learned-$z_S$ retrieval $.883$ and transported retrieval $.315$. Shift sweep: transport-over-reuse $.024\!\rightarrow\!.541$ as headroom grows.& Correct-vs-wrong SW gain $.523$ at $(1,0)$; usable identity tracks the gain more closely than Bayes-ID\\
CIFAR& Transported target-coordinate retrieval $49$--$62.5\%$ across relations (FiLM checkpoint) & Correct source $+$ transport improves on wrong/relation-only at $\KT=0$; the separate final-test budget comparison is in Table~\ref{tab:cifar-main}\\
Office-Home & Product-$z$ retrieval $91.0\%$; transport raises target-coordinate retrieval $12.0\%\!\rightarrow\!42.0\%$ & Clipart share $86.5\%$; low object ID persists even with abundant direct $z_T$, showing that transport error alone is insufficient to explain the failure.\\
\bottomrule
\end{tabular}
\end{table}

A practical implication is to examine the pipeline before relying on source-informed generation. Does the learned source representation distinguish the tasks? Is the relation sufficiently predictive on held-out tasks, and does transport retain their identity? Is there enough source--target discrepancy for transport to improve on source reuse or a simpler known image-space operation? Finally, does changing the coordinate produce the requested class, rather than merely the right domain statistics? The GMM study examines the representation, relation and headroom factors under known ground truth; Office-Home shows that semantic failure can persist after transport is bypassed. Good coordinate retrieval, low distributional distance and class fidelity are different outcomes.

The transport itself is compact: at $k=16$, a single-relation affine map has 272 active parameters, and the CIFAR implementation uses a shared matrix with three relation-specific biases (304 active parameters), without deployment updates. Source specificity also survives in the separate FiLM conditioning tests, while nonlinear-map and architecture comparisons have their own one-seed scope. The measured SW differences across source sets of 16--300 images are small, but do not establish a plateau. In the final-test comparison, transport has lower mean SW at $K_T=1,5$ under validation-selected, budget-specific adaptation settings, while full fine-tuning has lower SW at $K_T=20$. Since the optimisation settings also change, this is not an isolated target-data effect; the fixed-update experiment provides a complementary comparison.

\paragraph{Limitations.}
Where the CIFAR transformation can be applied directly, generating from the clean source and then applying the known corruption achieves lower SW than learned transport. Neither CIFAR nor Office-Home establishes reliable class-correct generation, and the clinical investigations in Appendix~\ref{app:clinical} show no resolved source-specific advantage. The two propositions, proved in Appendix~\ref{app:theory}, concern coordinate estimation and idealised continuous-time stability, not sufficient conditions for relational generalisation. The test relations were observed during meta-training and their identities were supplied at deployment; transfer to new relations remains untested. CIFAR headline experiments use one training seed. The class-cluster intervals do not capture seed-to-seed variability or variation across the four held-out superclasses. On Office-Home, full fine-tuning is favoured from five target images under the reported protocols.

\FloatBarrier
\section{Conclusion}
When source and target domains recur across categories, paired examples from earlier tasks offer information about how a new category may change between those domains. Meta-Diff explores this idea by learning an affine relation between task coordinates and using the transported source coordinate to condition a shared diffusion model without target-time adaptation. Its theoretical results establish limited estimation and stability properties under explicit assumptions, while the Gaussian-mixture experiments show how usable task identity, predictable relations and source--target headroom affect transfer. On images, correct-source transport improves some distributional comparisons, but a known pixel-space transformation can be better and domain alignment does not guarantee object identity. The contribution is thus a framework for learning reusable relations between task distributions, together with evidence about when those relations help and where semantic generation remains a bottleneck.

\subsection*{AI use statement}
We used generative artificial intelligence (AI) tools for language editing, drafting support, literature search, and to support some implementation and analysis tasks. The authors reviewed all AI-assisted text, technical claims, derivations, experimental specifications, analyses, and code, and take full responsibility for the final content.

\subsection*{Reproducibility statement}
The paper specifies the task-space encoder, conditioning mechanisms, source--target transport map, meta-training objective, evaluation protocols, baselines, and ablations. Full proofs are in Appendix~\ref{app:theory}; complete GMM diagnostics, CIFAR-100 controls and ablations, Office-Home experiments, and clinical stress tests are in Appendices~\ref{app:gmm}, \ref{app:cifar}, \ref{app:office}, and \ref{app:clinical}; supplementary controlled experiments are in Appendix~\ref{app:new-experiments}. Implementation details, data splits, adaptation settings, parameter counts, statistical units, and checkpoint-selection procedures are also reported in the appendix.

\section*{Impact Statement}
This work studies generative modelling under uneven data availability. Source-informed generation may reduce target-specific optimisation where representative target data are difficult to obtain, but synthetic data can also reproduce existing bias or create misleading detail. We therefore evaluate against held-out real data, retain negative stress tests, and do not treat generated clinical images as substitutes for representative collection or clinical validation. Any healthcare deployment would require independent clinical evaluation, governance and privacy review beyond the scope of this work.
\bibliographystyle{iclr2027_conference}
\bibliography{references}

\clearpage
\appendix
\let\MetaDiffAppendixInput\input
\section{Proofs for the Main-Text Guarantees}
\label{app:theory}

The guarantees in Section~\ref{sec:method} justify source-coordinate estimation and stability with respect to coordinate error only. They do not establish low-dimensionality, semantic sufficiency, task identifiability, or cross-task relational generalisation.

\subsection{Proof of Proposition~\ref{prop:source-concentration}}

Let $Y_i=h_\psi(X_i)-\mu$. Then $\E[Y_i]=0$, the $Y_i$ are independent, and
\[
\bar h_M-\mu=\frac{1}{M}\sum_{i=1}^{M}Y_i.
\]
By the $L_\rho$-Lipschitz property of $\rho_\psi$,
\[
\|z_M-\rho_\psi(\mu)\|_2^2
\leq
L_\rho^2\|\bar h_M-\mu\|_2^2.
\]
Taking expectations and expanding,
\begin{align*}
\E\|\bar h_M-\mu\|_2^2
&=
\frac{1}{M^2}
\E\left\|\sum_{i=1}^{M}Y_i\right\|_2^2\\
&=
\frac{1}{M^2}
\left[
\sum_{i=1}^{M}\E\|Y_i\|_2^2
+
2\sum_{i<j}\E\langle Y_i,Y_j\rangle
\right].
\end{align*}
For $i\neq j$, independence and zero mean imply
\[
\E\langle Y_i,Y_j\rangle
=
\langle \E Y_i,\E Y_j\rangle=0.
\]
Hence
\[
\E\|\bar h_M-\mu\|_2^2
=
\frac{1}{M}\E\|h_\psi(X)-\mu\|_2^2
=
\frac{\mathrm{tr}(\mathrm{Cov}[h_\psi(X)])}{M}.
\]
Combining the last two displays yields
\[
\E\|z_M-\rho_\psi(\mu)\|_2^2
\leq
\frac{L_\rho^2\,\mathrm{tr}(\mathrm{Cov}[h_\psi(X)])}{M}.
\]
This proves Proposition~\ref{prop:source-concentration}. The result controls mean-squared estimation error around the population feature coordinate; it does not imply that this coordinate is semantically sufficient.

\subsection{Proof of Proposition~\ref{prop:coordinate-stability}}

Use the increasing reverse-sampling clock of Proposition~\ref{prop:coordinate-stability}, denoted by $t$ here for brevity, and drift $\bar v_z$. Let $X_0^z=X_0^{z'}$ have the same coordinate-independent law $\mu_{\rm init}$ with finite second moment. Couple the well-posed processes with this initial random variable and the same Brownian path, and write
\[
\Delta_t=X_t^z-X_t^{z'}.
\]
Because the diffusion coefficient does not depend on $z$, the stochastic increments cancel under this synchronous coupling. With $\Delta_0=0$,
\[
\Delta_t
=
\int_0^t
\bigl[
\bar v_z(X_s^z,s)-\bar v_{z'}(X_s^{z'},s)
\bigr]\,ds.
\]
Adding and subtracting $\bar v_z(X_s^{z'},s)$ gives
\begin{align*}
\|\Delta_t\|_2
&\leq
\int_0^t
\|\bar v_z(X_s^z,s)-\bar v_z(X_s^{z'},s)\|_2\,ds\\
&\quad+
\int_0^t
\|\bar v_z(X_s^{z'},s)-\bar v_{z'}(X_s^{z'},s)\|_2\,ds.
\end{align*}
The state-Lipschitz assumption bounds the first integrand by $L_x\|\Delta_s\|_2$. For the second,
\[
\bar v_z(x,s)-\bar v_{z'}(x,s)
=
+g(\tau-s)^2B(x,\tau-s)(z-z'),
\]
so
\[
\|\bar v_z(x,s)-\bar v_{z'}(x,s)\|_2
\leq
L_z\|z-z'\|_2.
\]
Therefore
\[
\|\Delta_t\|_2
\leq
L_x\int_0^t\|\Delta_s\|_2\,ds
+
L_z t\|z-z'\|_2.
\]
The integral form of Gr\"onwall's inequality yields
\[
\|\Delta_t\|_2
\leq
L_z\|z-z'\|_2
\int_0^t e^{L_x(t-s)}\,ds
=
L_z\phi_{L_x}(t)\|z-z'\|_2,
\]
where
\[
\phi_{L_x}(t)=
\begin{cases}
(e^{L_xt}-1)/L_x,&L_x>0,\\
t,&L_x=0.
\end{cases}
\]
Since $\phi_{L_x}$ is nondecreasing,
\[
\sup_{t\leq\tau}\|\Delta_t\|_2
\leq
L_z\phi_{L_x}(\tau)\|z-z'\|_2.
\]
Squaring and taking expectations proves the first claim with
$C_\tau=L_z^2\phi_{L_x}(\tau)^2$.

For the Wasserstein statement, the synchronous construction is a valid coupling of the endpoint laws $P_z$ and $P_{z'}$. Hence
\[
W_2(P_z,P_{z'})^2
\leq
\E\|X_\tau^z-X_\tau^{z'}\|_2^2
\leq
C_\tau\|z-z'\|_2^2,
\]
and taking square roots completes the proof.

The proposition establishes continuity of the generated law with respect to $z$ under the stated regularity assumptions and a common initial law coupled identically. Its displayed drift uses additive score conditioning. A general FiLM-conditioned drift would require a corresponding uniform Lipschitz bound in $z$ as an additional assumption; the FiLM experiments alone do not establish that bound. It does not imply that the learned coordinate is low-dimensional, task-identifying, or transferable across unseen semantic tasks.

\FloatBarrier

\section{Gaussian-Mixture Experiments}
\label{app:gmm}

\subsection{Coordinate calibration}
Five separate 2-D Gaussian-mixture families check whether the learned generator responds to task coordinates. We evaluate calibration on each corresponding 80k-step checkpoint using 32 held-out tasks, $K_T=5$ and eight noise draws per task. The mean-minus-own loss interval is positive in 5 of the five families. These conditioning diagnostics remain separate from the factorial boundary study.

\begin{table}[h]
\centering\small
\caption{GMM calibration on five separate 80k-step checkpoints. Each uses 32 held-out tasks; eight noise draws are averaged within task. $\Delta_{\rm task}=\mathcal{L}(\bar z)-\mathcal{L}(z^y)$ has a paired task-level 95\% half-width. This is a coordinate intervention, not a deployment comparison.}
\begin{tabular}{lccc}
\toprule
Task family & Own & Mean & $\Delta_{\rm task}$\\
\midrule
Perturbation .10 & $0.4398$ & $0.5956$ & $0.1558\pm0.0286\sigstar$\\
Perturbation .25 & $0.4912$ & $0.6125$ & $0.1212\pm0.0232\sigstar$\\
Perturbation .50 & $0.5847$ & $0.6729$ & $0.0882\pm0.0272\sigstar$\\
Perturbation 1.00 & $0.6684$ & $0.7591$ & $0.0907\pm0.0423\sigstar$\\
Unrelated & $0.5427$ & $0.6386$ & $0.0959\pm0.0341\sigstar$\\
\bottomrule
\end{tabular}
\end{table}

\begin{figure}[t]
\centering
\includegraphics[width=.98\linewidth]{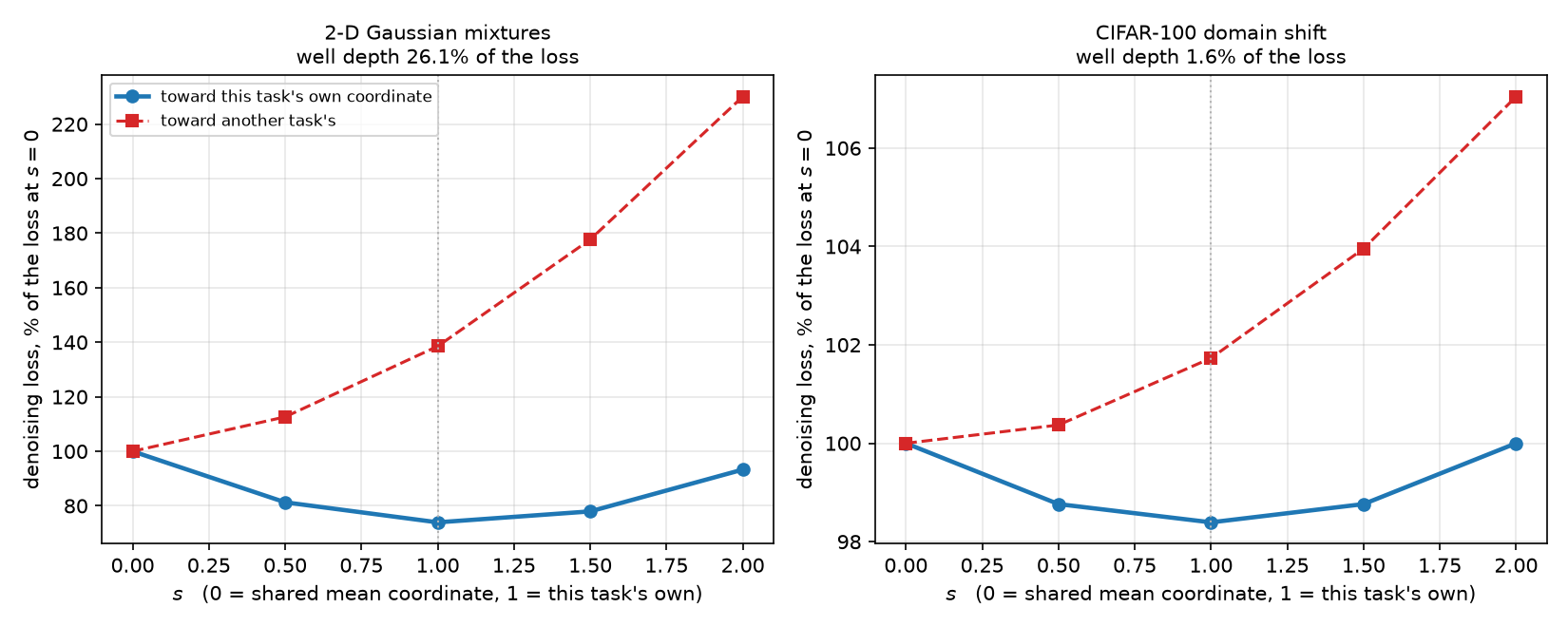}
\caption{\textbf{Task-coordinate interpolation.} The GMM and CIFAR panels use their respective model checkpoints. Losses are normalised by the shared-coordinate loss; well depth is the percentage drop from this loss to the minimum along the own-task direction. Own-task and wrong-task directions keep their diagnostic meanings. Their shapes do not replace the paired deployment-route comparisons.}
\label{fig:coord-profile}
\end{figure}

Figure~\ref{fig:coord-profile} has two different checkpoint sources: the GMM panel uses the 80k-step perturbation-0.1 checkpoint, and the CIFAR panel uses the 50k-step \texttt{cifar\_ds3} checkpoint. They interpolate encoder coordinates and diagnose conditioning; neither panel directly measures correct-source deployment superiority.

\FloatBarrier
\subsection{Complete factorial diagnostics}
The boundary experiment uses 64/16/64 train/validation/test semantic tasks, five model/data seeds, 8,000 training steps, linear transport and $\KT=0$. Each task is a four-component 2-D GMM with fixed mixture weights and covariances; only component means vary across tasks. Let $\bar\mu_{S,j}$ be the component-wise mean over meta-training tasks. Source-task identifiability is controlled by
\[
\mu_{S,j}^{y}(\alpha)
=
\bar\mu_{S,j}
+\alpha\bigl(\mu_{S,j}^{y,0}-\bar\mu_{S,j}\bigr),
\]
where $\mu_{S,j}^{y,0}$ is the original component mean and $\alpha\in\{0,.1,.25,.5,.75,1,1.5\}$. Thus $\alpha=0$ makes source tasks distributionally identical, $\alpha=1$ recovers the original task separation, and $\alpha>1$ increases it.

Targets share the affine base relation
\[
T_0(u)=Qu+\delta,\qquad
Q=1.10
\begin{bmatrix}
\cos25^\circ&-\sin25^\circ\\
\sin25^\circ&\cos25^\circ
\end{bmatrix},
\qquad
\delta=
\begin{bmatrix}
0.75\\-0.50
\end{bmatrix}.
\]
For each task $y$ and component $j$, a perturbation $\xi_j^{y}\sim\mathcal N(0,I_2)$ is drawn once and reused across the heterogeneity sweep. After normalising these perturbations to unit RMS over training tasks and components, target means are
\[
\mu_{T,j}^{y}(\alpha,\eta)
=
T_0\!\left(\mu_{S,j}^{y}(\alpha)\right)
+\eta s_R(\alpha)\xi_j^{y},
\]
where $s_R(\alpha)$ is the RMS magnitude of the common source--target transformation over meta-training components and
$\eta\in\{0,.1,.25,.5,1,2\}$. Hence $\eta=0$ gives a perfectly shared relation, while increasing $\eta$ introduces task-specific displacement that cannot be inferred from other source--target pairs. The same latent tasks and perturbations are reused across neighbouring cells to support paired comparisons.
The component index is $j\in\{1,\ldots,4\}$; it is unrelated to the refinement-step count $J$. The transformation scale depends on the source-separation condition:
\[
s_R(\alpha)^2=\frac{1}{4|\Ttrain|}\sum_{y\in\Ttrain}\sum_{j=1}^{4}
\bigl\|T_0(\mu_{S,j}^{y}(\alpha))-\mu_{S,j}^{y}(\alpha)\bigr\|_2^2.
\]
Training-task centres and normalisation constants are applied unchanged to validation and test tasks. Bayes task identification uses the known source-GMM likelihoods over the 64 held-out candidates; learned identification uses the trained encoder coordinate. Neither identification accuracy is a generated-image fidelity metric. Route SW is the sliced 2-Wasserstein distance over 256 random unit projections between 4,096 generated samples and 10,000 samples from the true target GMM, with sorted projections matched at evenly spaced order statistics; energy MMD uses the first 1,024 points of each set.

\begin{table*}[t]
\centering\scriptsize
\caption{Complete GMM factorial diagnostics. Top: correct-source advantage $G_{\rm source}=\mathrm{SW}_{wrong}-\mathrm{SW}_{correct}$. Middle: learned-coordinate task-ID accuracy. Bottom: held-out true-mean affine relation $R^2$. Each cell averages five seeds; generation effects further average 64 held-out tasks per seed.}
\label{tab:gmm-full-matrices}
\vspace{2pt}\textbf{Correct-source advantage $G_{\rm source}$}\par\vspace{1pt}
\begin{tabular}{ccccccc}
\toprule $\alpha\backslash\eta$ & 0 & 0.1 & 0.25 & 0.5 & 1 & 2 \\ \midrule
0 & 0.000 & 0.000 & -0.000 & 0.000 & -0.000 & -0.002 \\
0.1 & 0.008 & 0.010 & 0.006 & 0.006 & 0.002 & 0.002 \\
0.25 & 0.093 & 0.102 & 0.098 & 0.066 & 0.038 & 0.023 \\
0.5 & 0.242 & 0.209 & 0.205 & 0.220 & 0.140 & 0.091 \\
0.75 & 0.387 & 0.390 & 0.370 & 0.321 & 0.256 & 0.142 \\
1 & 0.523 & 0.563 & 0.523 & 0.447 & 0.404 & 0.228 \\
1.5 & 0.693 & 0.736 & 0.715 & 0.661 & 0.567 & 0.427 \\
\bottomrule\end{tabular}\par\vspace{4pt}
\vspace{2pt}\textbf{Learned-coordinate task ID}\par\vspace{1pt}
\begin{tabular}{ccccccc}
\toprule $\alpha\backslash\eta$ & 0 & 0.1 & 0.25 & 0.5 & 1 & 2 \\ \midrule
0 & 0.014 & 0.016 & 0.016 & 0.020 & 0.018 & 0.013 \\
0.1 & 0.074 & 0.061 & 0.057 & 0.069 & 0.078 & 0.098 \\
0.25 & 0.338 & 0.340 & 0.316 & 0.282 & 0.324 & 0.349 \\
0.5 & 0.718 & 0.737 & 0.717 & 0.728 & 0.670 & 0.678 \\
0.75 & 0.866 & 0.861 & 0.849 & 0.845 & 0.825 & 0.791 \\
1 & 0.873 & 0.894 & 0.891 & 0.886 & 0.871 & 0.830 \\
1.5 & 0.928 & 0.920 & 0.925 & 0.920 & 0.908 & 0.882 \\
\bottomrule\end{tabular}\par\vspace{4pt}
\vspace{2pt}\textbf{True-mean affine relation $R^2$}\par\vspace{1pt}
\begin{tabular}{ccccccc}
\toprule $\alpha\backslash\eta$ & 0 & 0.1 & 0.25 & 0.5 & 1 & 2 \\ \midrule
0 & 1.000 & 0.849 & 0.509 & 0.227 & 0.067 & 0.007 \\
0.1 & 1.000 & 0.918 & 0.652 & 0.327 & 0.105 & 0.019 \\
0.25 & 1.000 & 0.975 & 0.861 & 0.609 & 0.276 & 0.077 \\
0.5 & 1.000 & 0.992 & 0.949 & 0.824 & 0.538 & 0.218 \\
0.75 & 1.000 & 0.995 & 0.971 & 0.892 & 0.672 & 0.333 \\
1 & 1.000 & 0.996 & 0.978 & 0.919 & 0.738 & 0.408 \\
1.5 & 1.000 & 0.997 & 0.984 & 0.939 & 0.794 & 0.487 \\
\bottomrule\end{tabular}\par\vspace{4pt}
\end{table*}

\begin{figure}[t]
\centering
\includegraphics[width=.72\linewidth]{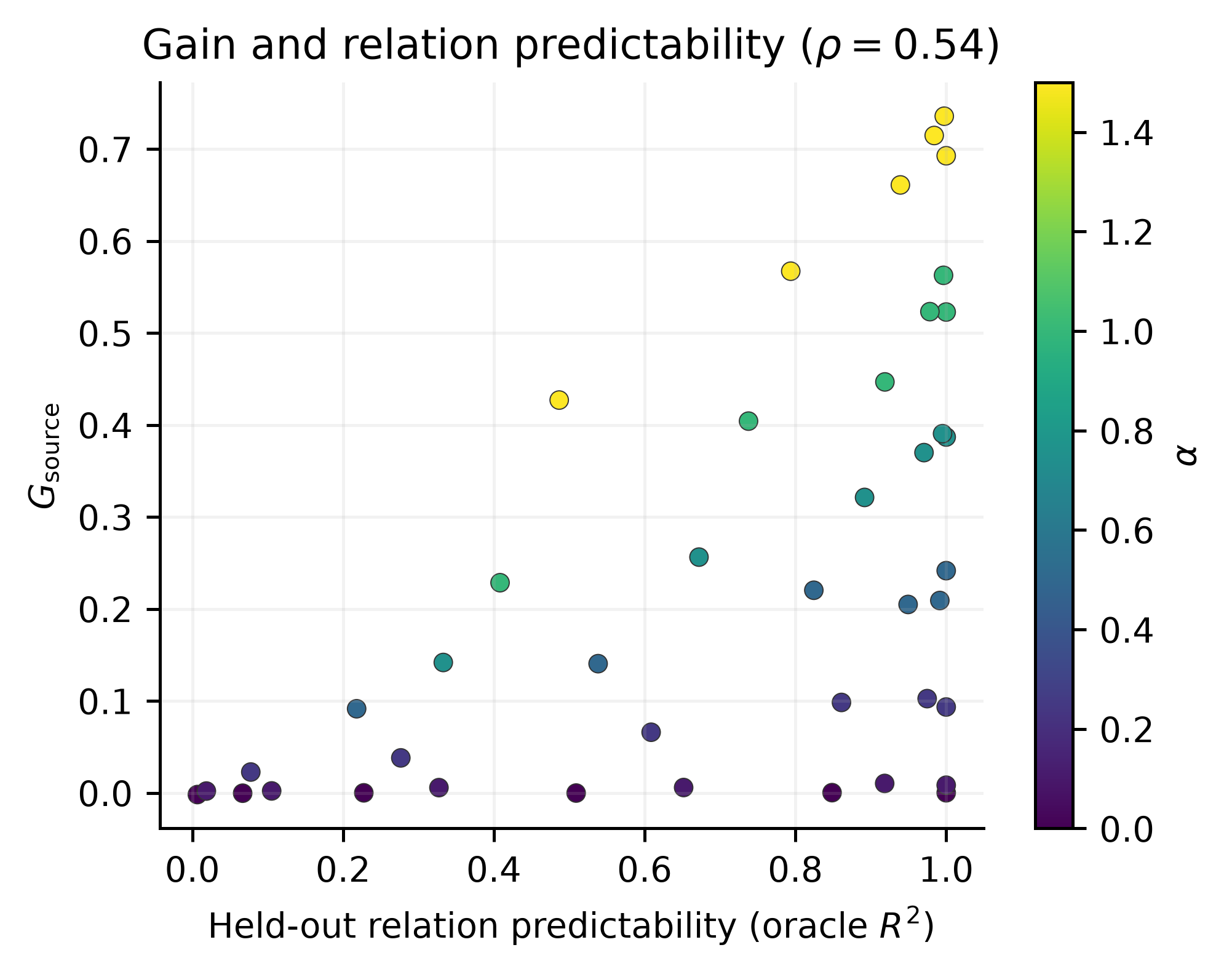}
\caption{\textbf{Relation predictability is a separate boundary from task identifiability.} Each colour fixes source-task separation $\alpha$ while $\eta$ varies. The horizontal axis is the held-out $R^2$ of an affine map fitted to true training-task component means; it is therefore independent of Meta-Diff's learned coordinate. At $\alpha=1$, Bayes-ID remains 1.0 while relation $R^2$ falls from 1.000 to .408 and the correct-over-wrong SW advantage falls from .523 to .228. Over the full grid, $\rho(G_{\rm source},R^2_{\rm rel})=.54$.}
\label{fig:gmm-relation-app}
\end{figure}

The full grid makes three distinctions visible. First, $G_{\rm source}$ is effectively zero at $\alpha=0$ even though a shared relation can still move the source distribution. Second, learned task discrimination increases much more slowly than Bayes identifiability around $\alpha=.1$--$.25$. Third, relation predictability falls with $\eta$ even when Bayes-ID is saturated. The mechanism analysis makes the second point sharper: the rank association with $G_{\rm source}$ is $\rho=.98$ for learned-ID versus $.89$ for Bayes-ID, and replacing Bayes-ID with learned-ID in the same interaction model increases cell-level fit from $R^2=.587$ to $.843$. These regressions are descriptive summaries of the factorial grid rather than causal estimates. Figure~\ref{fig:gmm-relation-app} shows relation predictability, and Figure~\ref{fig:gmm-full-controls} the full-grid controls.

\begin{figure*}[t]
\centering
\resizebox{.96\linewidth}{!}{\input{plots/gmm_full_grid_diagnostics.tex}}
\caption{\textbf{Full-grid GMM controls.} \emph{Left:} relation-only minus correct-source SW; positive values favour the source-conditioned route and stars mark hierarchical paired intervals excluding zero. \emph{Right:} true-source pairwise SW (in units of $10^{-3}$), an outcome-independent source-separation diagnostic that is not on the scale of the route SW in Table~\ref{tab:gmm-sw-absolute}. The source-conditioned route has a resolved advantage over relation-only in 33 of 42 cells; Table~\ref{tab:gmm-sw-paired} also contains 34 resolved correct-versus-wrong cells, while source-reuse and abundant-target-reference comparisons are not uniformly favourable.}
\label{fig:gmm-full-controls}
\end{figure*}

\FloatBarrier
\paragraph{Complete route means and paired uncertainty.}
The following tables give SW/MMD means and paired route contrasts for all 42 factorial cells. Route intervals use the hierarchical seed/task bootstrap and remain separate from the independent pipeline and shift-sweep summaries below.

\begin{table}[htbp]
\centering
\small
\setlength{\tabcolsep}{3pt}
\caption{Complete factorial GMM SW route means. Lower is better; five seeds and 64 held-out tasks per seed.}\label{tab:gmm-sw-absolute}
\begin{tabular}{ccrrrrr}
\toprule
$\alpha$ & $\eta$ & Correct & Wrong & Relation-only & Source reuse & \shortstack{Abundant-target\\reference}\\
\midrule
0 & 0 & 0.09386 & 0.09396 & 0.09361 & 0.64158 & 0.09311\\
0 & 0.1 & 0.14144 & 0.14170 & 0.14129 & 0.70917 & 0.14840\\
0 & 0.25 & 0.16930 & 0.16911 & 0.16933 & 0.63886 & 0.16775\\
0 & 0.5 & 0.32266 & 0.32270 & 0.32317 & 0.77757 & 0.31470\\
0 & 1 & 0.58670 & 0.58635 & 0.58467 & 0.71809 & 0.52292\\
0 & 2 & 1.04219 & 1.04036 & 1.04108 & 1.16237 & 0.90496\\
0.1 & 0 & 0.18620 & 0.19456 & 0.19078 & 0.64653 & 0.19528\\
0.1 & 0.1 & 0.16964 & 0.17991 & 0.17319 & 0.71509 & 0.18359\\
0.1 & 0.25 & 0.23343 & 0.23926 & 0.23713 & 0.68183 & 0.23469\\
0.1 & 0.5 & 0.32876 & 0.33447 & 0.33166 & 0.76656 & 0.34392\\
0.1 & 1 & 0.61856 & 0.62068 & 0.62086 & 0.77128 & 0.56109\\
0.1 & 2 & 1.13177 & 1.13375 & 1.13227 & 1.19641 & 0.99374\\
0.25 & 0 & 0.26717 & 0.36053 & 0.32118 & 0.70593 & 0.29249\\
0.25 & 0.1 & 0.25727 & 0.35972 & 0.31534 & 0.68822 & 0.25517\\
0.25 & 0.25 & 0.30459 & 0.40290 & 0.35736 & 0.73701 & 0.31244\\
0.25 & 0.5 & 0.38146 & 0.44737 & 0.41840 & 0.73248 & 0.34137\\
0.25 & 1 & 0.66614 & 0.70417 & 0.69010 & 0.82028 & 0.55004\\
0.25 & 2 & 1.16372 & 1.18639 & 1.17351 & 1.23343 & 0.98578\\
0.5 & 0 & 0.43521 & 0.67676 & 0.57230 & 0.75540 & 0.48756\\
0.5 & 0.1 & 0.46706 & 0.67618 & 0.59320 & 0.66103 & 0.45364\\
0.5 & 0.25 & 0.50701 & 0.71183 & 0.62122 & 0.76954 & 0.52467\\
0.5 & 0.5 & 0.53815 & 0.75841 & 0.64413 & 0.78782 & 0.52228\\
0.5 & 1 & 0.83608 & 0.97654 & 0.91079 & 0.89080 & 0.68141\\
0.5 & 2 & 1.31188 & 1.40333 & 1.36229 & 1.32739 & 1.08919\\
0.75 & 0 & 0.63087 & 1.01769 & 0.85555 & 0.87930 & 0.68385\\
0.75 & 0.1 & 0.59480 & 0.98529 & 0.82756 & 0.83604 & 0.63539\\
0.75 & 0.25 & 0.65450 & 1.02438 & 0.87596 & 0.87217 & 0.68470\\
0.75 & 0.5 & 0.77528 & 1.09644 & 0.95094 & 0.93380 & 0.73713\\
0.75 & 1 & 0.98062 & 1.23690 & 1.13105 & 1.05599 & 0.85252\\
0.75 & 2 & 1.57659 & 1.71834 & 1.66407 & 1.47961 & 1.30518\\
1 & 0 & 0.83601 & 1.35871 & 1.14405 & 1.00989 & 0.87475\\
1 & 0.1 & 0.77718 & 1.33978 & 1.10764 & 1.01242 & 0.83143\\
1 & 0.25 & 0.84056 & 1.36367 & 1.14931 & 1.06450 & 0.88271\\
1 & 0.5 & 0.98325 & 1.42997 & 1.24690 & 1.04953 & 0.93214\\
1 & 1 & 1.19519 & 1.59930 & 1.40758 & 1.25568 & 1.07821\\
1 & 2 & 1.83706 & 2.06552 & 1.95703 & 1.68288 & 1.53973\\
1.5 & 0 & 1.30175 & 1.99432 & 1.71676 & 1.58434 & 1.45669\\
1.5 & 0.1 & 1.32602 & 2.06157 & 1.75072 & 1.56571 & 1.39372\\
1.5 & 0.25 & 1.32210 & 2.03668 & 1.74836 & 1.58367 & 1.39352\\
1.5 & 0.5 & 1.38495 & 2.04579 & 1.77314 & 1.57787 & 1.41911\\
1.5 & 1 & 1.66776 & 2.23484 & 2.02376 & 1.67006 & 1.50820\\
1.5 & 2 & 2.47424 & 2.90125 & 2.69910 & 2.32351 & 2.11900\\
\bottomrule
\end{tabular}
\end{table}

\begin{table}[htbp]
\centering
\small
\setlength{\tabcolsep}{3pt}
\caption{Complete factorial GMM SW source controls: comparator minus correct-source transport, with hierarchical seed/task 95\% intervals. Positive favours correct-source transport.}\label{tab:gmm-sw-paired}
\begin{tabular}{ccll}
\toprule
$\alpha$ & $\eta$ & Wrong $-$ correct & Relation-only $-$ correct\\
\midrule
0 & 0 & $+0.00010\,[-0.00035,+0.00059]$ & $-0.00025\,[-0.00086,+0.00036]$\\
0 & 0.1 & $+0.00025\,[-0.00054,+0.00093]$ & $-0.00015\,[-0.00087,+0.00044]$\\
0 & 0.25 & $-0.00019\,[-0.00106,+0.00062]$ & $+0.00003\,[-0.00105,+0.00102]$\\
0 & 0.5 & $+0.00004\,[-0.00138,+0.00160]$ & $+0.00051\,[-0.00113,+0.00227]$\\
0 & 1 & $-0.00035\,[-0.00319,+0.00254]$ & $-0.00203\,[-0.00437,+0.00043]$\\
0 & 2 & $-0.00182\,[-0.00695,+0.00300]$ & $-0.00110\,[-0.00633,+0.00385]$\\
0.1 & 0 & $+0.00835\sigstar\,[+0.00423,+0.01379]$ & $+0.00458\sigstar\,[+0.00175,+0.00758]$\\
0.1 & 0.1 & $+0.01027\sigstar\,[+0.00584,+0.01483]$ & $+0.00355\sigstar\,[+0.00065,+0.00663]$\\
0.1 & 0.25 & $+0.00583\sigstar\,[+0.00176,+0.01033]$ & $+0.00370\sigstar\,[+0.00007,+0.00708]$\\
0.1 & 0.5 & $+0.00572\sigstar\,[+0.00102,+0.01033]$ & $+0.00291\sigstar\,[+0.00013,+0.00566]$\\
0.1 & 1 & $+0.00213\,[-0.00186,+0.00623]$ & $+0.00231\,[-0.00150,+0.00582]$\\
0.1 & 2 & $+0.00198\,[-0.00546,+0.00931]$ & $+0.00051\,[-0.00778,+0.00873]$\\
0.25 & 0 & $+0.09336\sigstar\,[+0.07494,+0.11153]$ & $+0.05401\sigstar\,[+0.03773,+0.07027]$\\
0.25 & 0.1 & $+0.10245\sigstar\,[+0.07687,+0.12837]$ & $+0.05807\sigstar\,[+0.04549,+0.06963]$\\
0.25 & 0.25 & $+0.09831\sigstar\,[+0.07986,+0.11555]$ & $+0.05277\sigstar\,[+0.04184,+0.06411]$\\
0.25 & 0.5 & $+0.06591\sigstar\,[+0.04754,+0.08641]$ & $+0.03694\sigstar\,[+0.01948,+0.05338]$\\
0.25 & 1 & $+0.03803\sigstar\,[+0.02468,+0.05266]$ & $+0.02396\sigstar\,[+0.01248,+0.03724]$\\
0.25 & 2 & $+0.02266\sigstar\,[+0.00366,+0.04314]$ & $+0.00979\,[-0.01160,+0.03503]$\\
0.5 & 0 & $+0.24155\sigstar\,[+0.21516,+0.26783]$ & $+0.13709\sigstar\,[+0.10877,+0.16139]$\\
0.5 & 0.1 & $+0.20912\sigstar\,[+0.16496,+0.25024]$ & $+0.12614\sigstar\,[+0.09744,+0.15530]$\\
0.5 & 0.25 & $+0.20482\sigstar\,[+0.15696,+0.25306]$ & $+0.11421\sigstar\,[+0.08465,+0.14390]$\\
0.5 & 0.5 & $+0.22026\sigstar\,[+0.17131,+0.27742]$ & $+0.10598\sigstar\,[+0.07229,+0.13699]$\\
0.5 & 1 & $+0.14047\sigstar\,[+0.08743,+0.20163]$ & $+0.07471\sigstar\,[+0.04927,+0.10087]$\\
0.5 & 2 & $+0.09145\sigstar\,[+0.05889,+0.12524]$ & $+0.05041\sigstar\,[+0.00794,+0.08689]$\\
0.75 & 0 & $+0.38682\sigstar\,[+0.31401,+0.46122]$ & $+0.22468\sigstar\,[+0.18242,+0.26639]$\\
0.75 & 0.1 & $+0.39049\sigstar\,[+0.32193,+0.44730]$ & $+0.23276\sigstar\,[+0.18354,+0.27327]$\\
0.75 & 0.25 & $+0.36989\sigstar\,[+0.30159,+0.44673]$ & $+0.22146\sigstar\,[+0.17257,+0.26753]$\\
0.75 & 0.5 & $+0.32116\sigstar\,[+0.24744,+0.40265]$ & $+0.17567\sigstar\,[+0.11925,+0.23146]$\\
0.75 & 1 & $+0.25629\sigstar\,[+0.18494,+0.34726]$ & $+0.15043\sigstar\,[+0.09492,+0.20354]$\\
0.75 & 2 & $+0.14175\sigstar\,[+0.09699,+0.18694]$ & $+0.08748\sigstar\,[+0.04723,+0.13009]$\\
1 & 0 & $+0.52270\sigstar\,[+0.44979,+0.59553]$ & $+0.30804\sigstar\,[+0.25922,+0.35910]$\\
1 & 0.1 & $+0.56260\sigstar\,[+0.47423,+0.66587]$ & $+0.33046\sigstar\,[+0.27791,+0.38118]$\\
1 & 0.25 & $+0.52310\sigstar\,[+0.44510,+0.59904]$ & $+0.30874\sigstar\,[+0.24722,+0.36915]$\\
1 & 0.5 & $+0.44672\sigstar\,[+0.32666,+0.56506]$ & $+0.26366\sigstar\,[+0.20126,+0.33012]$\\
1 & 1 & $+0.40410\sigstar\,[+0.31186,+0.50894]$ & $+0.21238\sigstar\,[+0.15675,+0.27564]$\\
1 & 2 & $+0.22846\sigstar\,[+0.13314,+0.34093]$ & $+0.11997\sigstar\,[+0.05312,+0.18728]$\\
1.5 & 0 & $+0.69257\sigstar\,[+0.56014,+0.82813]$ & $+0.41501\sigstar\,[+0.33127,+0.50103]$\\
1.5 & 0.1 & $+0.73555\sigstar\,[+0.63170,+0.84237]$ & $+0.42470\sigstar\,[+0.35280,+0.49942]$\\
1.5 & 0.25 & $+0.71458\sigstar\,[+0.60007,+0.83214]$ & $+0.42626\sigstar\,[+0.33255,+0.51560]$\\
1.5 & 0.5 & $+0.66084\sigstar\,[+0.53810,+0.81305]$ & $+0.38819\sigstar\,[+0.31138,+0.47313]$\\
1.5 & 1 & $+0.56708\sigstar\,[+0.43503,+0.71278]$ & $+0.35600\sigstar\,[+0.25317,+0.46499]$\\
1.5 & 2 & $+0.42700\sigstar\,[+0.28685,+0.58549]$ & $+0.22485\sigstar\,[+0.14329,+0.30859]$\\
\bottomrule
\end{tabular}
\end{table}

\begin{table}[htbp]
\centering
\small
\setlength{\tabcolsep}{3pt}
\caption{Complete factorial GMM SW reuse/reference contrasts: comparator minus correct-source transport, with hierarchical seed/task 95\% intervals. Positive favours correct-source transport. The abundant-target coordinate reference encodes 512 target samples with the frozen encoder and uses the same frozen generator, without target-time coordinate optimisation. It is a reference route, not a performance upper bound.}\label{tab:gmm-sw-paired-references}
\begin{tabular}{ccll}
\toprule
$\alpha$ & $\eta$ & Source reuse $-$ correct & Abundant-target ref. $-$ correct\\
\midrule
0 & 0 & $+0.54772\sigstar\,[+0.47915,+0.61436]$ & $-0.00075\,[-0.02123,+0.01295]$\\
0 & 0.1 & $+0.56772\sigstar\,[+0.50726,+0.63460]$ & $+0.00695\,[-0.00126,+0.01545]$\\
0 & 0.25 & $+0.46955\sigstar\,[+0.40729,+0.51238]$ & $-0.00156\,[-0.01588,+0.01273]$\\
0 & 0.5 & $+0.45492\sigstar\,[+0.37794,+0.53592]$ & $-0.00796\,[-0.02093,+0.00691]$\\
0 & 1 & $+0.13139\,[-0.01521,+0.26379]$ & $-0.06379\sigstar\,[-0.09976,-0.02633]$\\
0 & 2 & $+0.12019\,[-0.02686,+0.29146]$ & $-0.13723\sigstar\,[-0.18348,-0.08521]$\\
0.1 & 0 & $+0.46033\sigstar\,[+0.40389,+0.52582]$ & $+0.00908\sigstar\,[+0.00413,+0.01490]$\\
0.1 & 0.1 & $+0.54545\sigstar\,[+0.47017,+0.62192]$ & $+0.01395\sigstar\,[+0.00489,+0.02307]$\\
0.1 & 0.25 & $+0.44839\sigstar\,[+0.37088,+0.51481]$ & $+0.00125\,[-0.00893,+0.00999]$\\
0.1 & 0.5 & $+0.43781\sigstar\,[+0.32473,+0.53652]$ & $+0.01517\,[-0.02963,+0.07979]$\\
0.1 & 1 & $+0.15272\sigstar\,[+0.03401,+0.24160]$ & $-0.05746\sigstar\,[-0.08008,-0.03720]$\\
0.1 & 2 & $+0.06465\,[-0.14536,+0.25890]$ & $-0.13803\sigstar\,[-0.18973,-0.08234]$\\
0.25 & 0 & $+0.43875\sigstar\,[+0.34914,+0.52549]$ & $+0.02532\,[-0.01838,+0.06959]$\\
0.25 & 0.1 & $+0.43095\sigstar\,[+0.33990,+0.51491]$ & $-0.00210\,[-0.01978,+0.01578]$\\
0.25 & 0.25 & $+0.43242\sigstar\,[+0.31778,+0.56028]$ & $+0.00784\,[-0.01576,+0.03842]$\\
0.25 & 0.5 & $+0.35102\sigstar\,[+0.25553,+0.44285]$ & $-0.04009\,[-0.07272,+0.00166]$\\
0.25 & 1 & $+0.15415\sigstar\,[+0.01199,+0.25514]$ & $-0.11609\sigstar\,[-0.14699,-0.08798]$\\
0.25 & 2 & $+0.06971\,[-0.06061,+0.19539]$ & $-0.17795\sigstar\,[-0.23321,-0.13095]$\\
0.5 & 0 & $+0.32019\sigstar\,[+0.26384,+0.36867]$ & $+0.05235\sigstar\,[+0.02156,+0.08589]$\\
0.5 & 0.1 & $+0.19397\sigstar\,[+0.08944,+0.29378]$ & $-0.01342\,[-0.07453,+0.04412]$\\
0.5 & 0.25 & $+0.26253\sigstar\,[+0.16573,+0.34602]$ & $+0.01765\,[-0.02270,+0.06038]$\\
0.5 & 0.5 & $+0.24967\sigstar\,[+0.12643,+0.35329]$ & $-0.01587\,[-0.06285,+0.03293]$\\
0.5 & 1 & $+0.05472\,[-0.10061,+0.15118]$ & $-0.15467\sigstar\,[-0.22629,-0.09391]$\\
0.5 & 2 & $+0.01552\,[-0.13468,+0.15717]$ & $-0.22269\sigstar\,[-0.29576,-0.16079]$\\
0.75 & 0 & $+0.24843\sigstar\,[+0.12217,+0.36943]$ & $+0.05298\,[-0.01921,+0.13013]$\\
0.75 & 0.1 & $+0.24124\sigstar\,[+0.12439,+0.34690]$ & $+0.04059\,[-0.01339,+0.08357]$\\
0.75 & 0.25 & $+0.21767\sigstar\,[+0.12208,+0.31442]$ & $+0.03021\,[-0.02956,+0.09356]$\\
0.75 & 0.5 & $+0.15852\sigstar\,[+0.05433,+0.24901]$ & $-0.03815\,[-0.07614,+0.00502]$\\
0.75 & 1 & $+0.07538\,[-0.05367,+0.16205]$ & $-0.12810\sigstar\,[-0.20445,-0.06870]$\\
0.75 & 2 & $-0.09698\,[-0.30044,+0.04797]$ & $-0.27141\sigstar\,[-0.39106,-0.17787]$\\
1 & 0 & $+0.17388\sigstar\,[+0.07600,+0.27364]$ & $+0.03874\,[-0.02206,+0.09591]$\\
1 & 0.1 & $+0.23524\sigstar\,[+0.14588,+0.32692]$ & $+0.05425\,[-0.00352,+0.10887]$\\
1 & 0.25 & $+0.22394\sigstar\,[+0.08812,+0.33918]$ & $+0.04214\,[-0.02872,+0.10773]$\\
1 & 0.5 & $+0.06628\,[-0.12114,+0.20964]$ & $-0.05111\,[-0.15309,+0.03730]$\\
1 & 1 & $+0.06048\,[-0.00260,+0.12129]$ & $-0.11699\sigstar\,[-0.18859,-0.04980]$\\
1 & 2 & $-0.15418\,[-0.33414,+0.00775]$ & $-0.29733\sigstar\,[-0.43268,-0.17561]$\\
1.5 & 0 & $+0.28258\sigstar\,[+0.17342,+0.39528]$ & $+0.15494\sigstar\,[+0.10056,+0.21521]$\\
1.5 & 0.1 & $+0.23969\sigstar\,[+0.09724,+0.38696]$ & $+0.06770\,[-0.07017,+0.19717]$\\
1.5 & 0.25 & $+0.26157\sigstar\,[+0.15161,+0.36427]$ & $+0.07142\sigstar\,[+0.00223,+0.14246]$\\
1.5 & 0.5 & $+0.19292\sigstar\,[+0.10353,+0.28814]$ & $+0.03416\,[-0.02510,+0.08968]$\\
1.5 & 1 & $+0.00230\,[-0.08436,+0.09102]$ & $-0.15957\sigstar\,[-0.24140,-0.07676]$\\
1.5 & 2 & $-0.15073\sigstar\,[-0.29265,-0.04305]$ & $-0.35525\sigstar\,[-0.46329,-0.24579]$\\
\bottomrule
\end{tabular}
\end{table}

\begin{table}[htbp]
\centering
\small
\setlength{\tabcolsep}{3pt}
\caption{Complete factorial GMM MMD route means. Lower is better; five seeds and 64 held-out tasks per seed.}\label{tab:gmm-mmd-absolute}
\begin{tabular}{ccrrrrr}
\toprule
$\alpha$ & $\eta$ & Correct & Wrong & Relation-only & Source reuse & \shortstack{Abundant-target\\reference}\\
\midrule
0 & 0 & 0.01827 & 0.01820 & 0.01813 & 0.88275 & 0.01782\\
0 & 0.1 & 0.03767 & 0.03772 & 0.03767 & 1.01943 & 0.04387\\
0 & 0.25 & 0.06638 & 0.06640 & 0.06653 & 0.87212 & 0.06448\\
0 & 0.5 & 0.14388 & 0.14391 & 0.14401 & 1.02813 & 0.13898\\
0 & 1 & 0.32266 & 0.32176 & 0.32035 & 0.71514 & 0.24480\\
0 & 2 & 0.72649 & 0.72400 & 0.72402 & 1.19824 & 0.52956\\
0.1 & 0 & 0.07334 & 0.08149 & 0.07906 & 0.83916 & 0.07990\\
0.1 & 0.1 & 0.05986 & 0.06983 & 0.06498 & 0.97429 & 0.06869\\
0.1 & 0.25 & 0.09329 & 0.10050 & 0.09806 & 0.91302 & 0.09706\\
0.1 & 0.5 & 0.16318 & 0.17106 & 0.16820 & 0.95430 & 0.18661\\
0.1 & 1 & 0.34343 & 0.34776 & 0.34619 & 0.80935 & 0.25908\\
0.1 & 2 & 0.75629 & 0.75893 & 0.75859 & 1.22893 & 0.54214\\
0.25 & 0 & 0.10815 & 0.23019 & 0.17891 & 0.78865 & 0.13855\\
0.25 & 0.1 & 0.10634 & 0.23746 & 0.17717 & 0.80646 & 0.10232\\
0.25 & 0.25 & 0.12515 & 0.25525 & 0.19413 & 0.81587 & 0.12691\\
0.25 & 0.5 & 0.18401 & 0.28003 & 0.23306 & 0.81571 & 0.14977\\
0.25 & 1 & 0.37590 & 0.43806 & 0.41686 & 0.81969 & 0.25679\\
0.25 & 2 & 0.78925 & 0.82407 & 0.81257 & 1.25444 & 0.56455\\
0.5 & 0 & 0.17616 & 0.49284 & 0.35407 & 0.59608 & 0.23008\\
0.5 & 0.1 & 0.22647 & 0.52662 & 0.39873 & 0.45728 & 0.20212\\
0.5 & 0.25 & 0.23612 & 0.52913 & 0.40286 & 0.59270 & 0.25045\\
0.5 & 0.5 & 0.25410 & 0.57246 & 0.40274 & 0.62153 & 0.23771\\
0.5 & 1 & 0.49623 & 0.72537 & 0.62030 & 0.73690 & 0.33088\\
0.5 & 2 & 0.82885 & 0.98155 & 0.91753 & 1.21317 & 0.55722\\
0.75 & 0 & 0.28914 & 0.83530 & 0.59069 & 0.60096 & 0.33253\\
0.75 & 0.1 & 0.27496 & 0.82089 & 0.58035 & 0.56537 & 0.29812\\
0.75 & 0.25 & 0.31701 & 0.85119 & 0.62212 & 0.60848 & 0.35899\\
0.75 & 0.5 & 0.39967 & 0.87885 & 0.64763 & 0.62969 & 0.34415\\
0.75 & 1 & 0.53643 & 0.93519 & 0.77855 & 0.75011 & 0.41116\\
0.75 & 2 & 0.99672 & 1.24134 & 1.15750 & 1.17534 & 0.68976\\
1 & 0 & 0.38453 & 1.13998 & 0.81764 & 0.61778 & 0.42747\\
1 & 0.1 & 0.34852 & 1.17486 & 0.82466 & 0.62534 & 0.37613\\
1 & 0.25 & 0.40165 & 1.17876 & 0.84802 & 0.69373 & 0.42724\\
1 & 0.5 & 0.53272 & 1.20578 & 0.95292 & 0.63116 & 0.45816\\
1 & 1 & 0.64184 & 1.28125 & 0.99301 & 0.83097 & 0.54703\\
1 & 2 & 1.13301 & 1.52550 & 1.35384 & 1.22301 & 0.82498\\
1.5 & 0 & 0.66734 & 1.76367 & 1.28258 & 1.04101 & 0.82003\\
1.5 & 0.1 & 0.68992 & 1.82891 & 1.34389 & 1.01392 & 0.73695\\
1.5 & 0.25 & 0.64330 & 1.74992 & 1.28208 & 1.00599 & 0.71198\\
1.5 & 0.5 & 0.68368 & 1.67984 & 1.25541 & 0.96404 & 0.69489\\
1.5 & 1 & 0.86344 & 1.75000 & 1.38795 & 0.94367 & 0.67324\\
1.5 & 2 & 1.62377 & 2.34254 & 1.99685 & 1.68128 & 1.17444\\
\bottomrule
\end{tabular}
\end{table}

\begin{table}[htbp]
\centering
\small
\setlength{\tabcolsep}{3pt}
\caption{Complete factorial GMM MMD source controls: comparator minus correct-source transport, with hierarchical seed/task 95\% intervals. Positive favours correct-source transport.}\label{tab:gmm-mmd-paired}
\begin{tabular}{ccll}
\toprule
$\alpha$ & $\eta$ & Wrong $-$ correct & Relation-only $-$ correct\\
\midrule
0 & 0 & $-0.00007\,[-0.00025,+0.00007]$ & $-0.00014\,[-0.00033,+0.00003]$\\
0 & 0.1 & $+0.00005\,[-0.00008,+0.00018]$ & $+0.00000\,[-0.00013,+0.00012]$\\
0 & 0.25 & $+0.00003\,[-0.00022,+0.00026]$ & $+0.00015\,[-0.00009,+0.00044]$\\
0 & 0.5 & $+0.00004\,[-0.00072,+0.00090]$ & $+0.00014\,[-0.00066,+0.00098]$\\
0 & 1 & $-0.00090\,[-0.00334,+0.00176]$ & $-0.00231\sigstar\,[-0.00389,-0.00061]$\\
0 & 2 & $-0.00249\,[-0.00761,+0.00164]$ & $-0.00247\,[-0.00659,+0.00100]$\\
0.1 & 0 & $+0.00815\sigstar\,[+0.00443,+0.01245]$ & $+0.00573\sigstar\,[+0.00352,+0.00875]$\\
0.1 & 0.1 & $+0.00998\sigstar\,[+0.00621,+0.01473]$ & $+0.00512\sigstar\,[+0.00289,+0.00797]$\\
0.1 & 0.25 & $+0.00720\sigstar\,[+0.00217,+0.01418]$ & $+0.00476\sigstar\,[+0.00159,+0.00878]$\\
0.1 & 0.5 & $+0.00788\sigstar\,[+0.00284,+0.01387]$ & $+0.00502\sigstar\,[+0.00219,+0.00798]$\\
0.1 & 1 & $+0.00433\,[-0.00037,+0.00924]$ & $+0.00277\,[-0.00149,+0.00691]$\\
0.1 & 2 & $+0.00264\,[-0.00755,+0.01284]$ & $+0.00230\,[-0.00857,+0.01487]$\\
0.25 & 0 & $+0.12203\sigstar\,[+0.09047,+0.15798]$ & $+0.07076\sigstar\,[+0.05145,+0.08979]$\\
0.25 & 0.1 & $+0.13112\sigstar\,[+0.09258,+0.18448]$ & $+0.07083\sigstar\,[+0.05726,+0.08734]$\\
0.25 & 0.25 & $+0.13011\sigstar\,[+0.10825,+0.15320]$ & $+0.06898\sigstar\,[+0.05395,+0.08484]$\\
0.25 & 0.5 & $+0.09601\sigstar\,[+0.07087,+0.12419]$ & $+0.04904\sigstar\,[+0.02913,+0.06673]$\\
0.25 & 1 & $+0.06216\sigstar\,[+0.04336,+0.08356]$ & $+0.04096\sigstar\,[+0.02735,+0.05778]$\\
0.25 & 2 & $+0.03481\sigstar\,[+0.00644,+0.06557]$ & $+0.02331\,[-0.00206,+0.05358]$\\
0.5 & 0 & $+0.31668\sigstar\,[+0.27528,+0.36086]$ & $+0.17791\sigstar\,[+0.15063,+0.20589]$\\
0.5 & 0.1 & $+0.30015\sigstar\,[+0.22862,+0.37459]$ & $+0.17226\sigstar\,[+0.13303,+0.22478]$\\
0.5 & 0.25 & $+0.29300\sigstar\,[+0.23177,+0.36024]$ & $+0.16674\sigstar\,[+0.13550,+0.19907]$\\
0.5 & 0.5 & $+0.31836\sigstar\,[+0.24775,+0.40319]$ & $+0.14864\sigstar\,[+0.10246,+0.19293]$\\
0.5 & 1 & $+0.22914\sigstar\,[+0.16015,+0.30839]$ & $+0.12407\sigstar\,[+0.08790,+0.15940]$\\
0.5 & 2 & $+0.15270\sigstar\,[+0.09963,+0.20514]$ & $+0.08868\sigstar\,[+0.03489,+0.13962]$\\
0.75 & 0 & $+0.54616\sigstar\,[+0.46194,+0.63855]$ & $+0.30155\sigstar\,[+0.24394,+0.37747]$\\
0.75 & 0.1 & $+0.54593\sigstar\,[+0.44574,+0.63694]$ & $+0.30539\sigstar\,[+0.23962,+0.37953]$\\
0.75 & 0.25 & $+0.53418\sigstar\,[+0.44127,+0.64401]$ & $+0.30511\sigstar\,[+0.23772,+0.38154]$\\
0.75 & 0.5 & $+0.47917\sigstar\,[+0.38579,+0.57822]$ & $+0.24796\sigstar\,[+0.16661,+0.33848]$\\
0.75 & 1 & $+0.39875\sigstar\,[+0.30187,+0.52259]$ & $+0.24212\sigstar\,[+0.16768,+0.31574]$\\
0.75 & 2 & $+0.24462\sigstar\,[+0.17113,+0.31805]$ & $+0.16078\sigstar\,[+0.09541,+0.22461]$\\
1 & 0 & $+0.75545\sigstar\,[+0.65833,+0.85396]$ & $+0.43311\sigstar\,[+0.34737,+0.53878]$\\
1 & 0.1 & $+0.82634\sigstar\,[+0.68290,+0.99393]$ & $+0.47615\sigstar\,[+0.38587,+0.56505]$\\
1 & 0.25 & $+0.77711\sigstar\,[+0.66190,+0.89134]$ & $+0.44637\sigstar\,[+0.35150,+0.57146]$\\
1 & 0.5 & $+0.67306\sigstar\,[+0.52330,+0.82360]$ & $+0.42019\sigstar\,[+0.31515,+0.52882]$\\
1 & 1 & $+0.63941\sigstar\,[+0.49330,+0.81341]$ & $+0.35117\sigstar\,[+0.25545,+0.46088]$\\
1 & 2 & $+0.39249\sigstar\,[+0.24244,+0.57382]$ & $+0.22082\sigstar\,[+0.11081,+0.32567]$\\
1.5 & 0 & $+1.09634\sigstar\,[+0.91938,+1.28449]$ & $+0.61525\sigstar\,[+0.49514,+0.75308]$\\
1.5 & 0.1 & $+1.13899\sigstar\,[+0.97726,+1.31227]$ & $+0.65397\sigstar\,[+0.55980,+0.75650]$\\
1.5 & 0.25 & $+1.10662\sigstar\,[+0.96072,+1.25630]$ & $+0.63877\sigstar\,[+0.49773,+0.79731]$\\
1.5 & 0.5 & $+0.99616\sigstar\,[+0.82566,+1.20117]$ & $+0.57174\sigstar\,[+0.44409,+0.70192]$\\
1.5 & 1 & $+0.88656\sigstar\,[+0.68842,+1.10578]$ & $+0.52451\sigstar\,[+0.37423,+0.69403]$\\
1.5 & 2 & $+0.71877\sigstar\,[+0.48858,+0.97687]$ & $+0.37308\sigstar\,[+0.23704,+0.50955]$\\
\bottomrule
\end{tabular}
\end{table}

\begin{table}[htbp]
\centering
\small
\setlength{\tabcolsep}{3pt}
\caption{Complete factorial GMM MMD reuse/reference contrasts: comparator minus correct-source transport, with hierarchical seed/task 95\% intervals. Positive favours correct-source transport. The abundant-target coordinate reference encodes 512 target samples with the frozen encoder and uses the same frozen generator, without target-time coordinate optimisation. It is a reference route, not a performance upper bound.}\label{tab:gmm-mmd-paired-references}
\begin{tabular}{ccll}
\toprule
$\alpha$ & $\eta$ & Source reuse $-$ correct & Abundant-target ref. $-$ correct\\
\midrule
0 & 0 & $+0.86448\sigstar\,[+0.62391,+1.08397]$ & $-0.00044\,[-0.01105,+0.00643]$\\
0 & 0.1 & $+0.98176\sigstar\,[+0.86073,+1.10203]$ & $+0.00620\sigstar\,[+0.00008,+0.01270]$\\
0 & 0.25 & $+0.80574\sigstar\,[+0.62634,+0.92682]$ & $-0.00190\,[-0.01245,+0.00877]$\\
0 & 0.5 & $+0.88426\sigstar\,[+0.69128,+1.05914]$ & $-0.00489\,[-0.02169,+0.01614]$\\
0 & 1 & $+0.39248\sigstar\,[+0.18649,+0.59469]$ & $-0.07786\sigstar\,[-0.11287,-0.04096]$\\
0 & 2 & $+0.47175\sigstar\,[+0.24454,+0.74321]$ & $-0.19693\sigstar\,[-0.24489,-0.15161]$\\
0.1 & 0 & $+0.76582\sigstar\,[+0.64899,+0.88055]$ & $+0.00656\sigstar\,[+0.00010,+0.01308]$\\
0.1 & 0.1 & $+0.91444\sigstar\,[+0.73172,+1.06691]$ & $+0.00883\sigstar\,[+0.00461,+0.01360]$\\
0.1 & 0.25 & $+0.81973\sigstar\,[+0.60398,+1.01282]$ & $+0.00377\,[-0.00503,+0.01209]$\\
0.1 & 0.5 & $+0.79112\sigstar\,[+0.62793,+0.91909]$ & $+0.02343\,[-0.02890,+0.10372]$\\
0.1 & 1 & $+0.46593\sigstar\,[+0.31219,+0.61078]$ & $-0.08435\sigstar\,[-0.11503,-0.05602]$\\
0.1 & 2 & $+0.47265\sigstar\,[+0.17378,+0.73851]$ & $-0.21415\sigstar\,[-0.24325,-0.18304]$\\
0.25 & 0 & $+0.68050\sigstar\,[+0.53245,+0.83404]$ & $+0.03039\,[-0.01902,+0.07962]$\\
0.25 & 0.1 & $+0.70012\sigstar\,[+0.56395,+0.82675]$ & $-0.00402\,[-0.02492,+0.01340]$\\
0.25 & 0.25 & $+0.69072\sigstar\,[+0.54898,+0.83160]$ & $+0.00176\,[-0.02138,+0.03217]$\\
0.25 & 0.5 & $+0.63169\sigstar\,[+0.45808,+0.82498]$ & $-0.03424\,[-0.07724,+0.01442]$\\
0.25 & 1 & $+0.44378\sigstar\,[+0.25424,+0.62415]$ & $-0.11912\sigstar\,[-0.14697,-0.09456]$\\
0.25 & 2 & $+0.46519\sigstar\,[+0.34597,+0.59740]$ & $-0.22470\sigstar\,[-0.27629,-0.17115]$\\
0.5 & 0 & $+0.41992\sigstar\,[+0.34712,+0.49157]$ & $+0.05392\sigstar\,[+0.02198,+0.09153]$\\
0.5 & 0.1 & $+0.23082\sigstar\,[+0.09143,+0.36996]$ & $-0.02435\,[-0.09469,+0.04308]$\\
0.5 & 0.25 & $+0.35657\sigstar\,[+0.20655,+0.49022]$ & $+0.01433\,[-0.03347,+0.06629]$\\
0.5 & 0.5 & $+0.36743\sigstar\,[+0.19529,+0.50937]$ & $-0.01639\,[-0.07890,+0.05097]$\\
0.5 & 1 & $+0.24067\sigstar\,[+0.01953,+0.39879]$ & $-0.16535\sigstar\,[-0.24434,-0.09497]$\\
0.5 & 2 & $+0.38432\sigstar\,[+0.25790,+0.51440]$ & $-0.27164\sigstar\,[-0.32902,-0.21952]$\\
0.75 & 0 & $+0.31182\sigstar\,[+0.13990,+0.47463]$ & $+0.04339\,[-0.01995,+0.10980]$\\
0.75 & 0.1 & $+0.29041\sigstar\,[+0.14437,+0.42873]$ & $+0.02316\,[-0.02126,+0.06039]$\\
0.75 & 0.25 & $+0.29147\sigstar\,[+0.15950,+0.42208]$ & $+0.04198\,[-0.02242,+0.10815]$\\
0.75 & 0.5 & $+0.23002\sigstar\,[+0.10975,+0.34290]$ & $-0.05552\sigstar\,[-0.09451,-0.01572]$\\
0.75 & 1 & $+0.21367\sigstar\,[+0.05158,+0.32043]$ & $-0.12527\sigstar\,[-0.20447,-0.06212]$\\
0.75 & 2 & $+0.17862\,[-0.02082,+0.34877]$ & $-0.30697\sigstar\,[-0.42875,-0.19865]$\\
1 & 0 & $+0.23325\sigstar\,[+0.10184,+0.37085]$ & $+0.04294\,[-0.01879,+0.10034]$\\
1 & 0.1 & $+0.27682\sigstar\,[+0.15706,+0.40179]$ & $+0.02761\,[-0.03696,+0.08544]$\\
1 & 0.25 & $+0.29208\sigstar\,[+0.11990,+0.45826]$ & $+0.02559\,[-0.04356,+0.09289]$\\
1 & 0.5 & $+0.09844\,[-0.10099,+0.24785]$ & $-0.07456\,[-0.18434,+0.01108]$\\
1 & 1 & $+0.18914\sigstar\,[+0.11140,+0.27233]$ & $-0.09481\sigstar\,[-0.16812,-0.02574]$\\
1 & 2 & $+0.08999\,[-0.09839,+0.25686]$ & $-0.30804\sigstar\,[-0.45103,-0.16747]$\\
1.5 & 0 & $+0.37367\sigstar\,[+0.21737,+0.53198]$ & $+0.15269\sigstar\,[+0.08160,+0.23430]$\\
1.5 & 0.1 & $+0.32400\sigstar\,[+0.10381,+0.55118]$ & $+0.04703\,[-0.12328,+0.20882]$\\
1.5 & 0.25 & $+0.36268\sigstar\,[+0.20650,+0.52492]$ & $+0.06868\,[-0.01017,+0.16233]$\\
1.5 & 0.5 & $+0.28036\sigstar\,[+0.14932,+0.42798]$ & $+0.01121\,[-0.05211,+0.07335]$\\
1.5 & 1 & $+0.08023\,[-0.01360,+0.17687]$ & $-0.19020\sigstar\,[-0.24430,-0.13708]$\\
1.5 & 2 & $+0.05751\,[-0.06773,+0.17574]$ & $-0.44933\sigstar\,[-0.60695,-0.29172]$\\
\bottomrule
\end{tabular}
\end{table}

\FloatBarrier
\subsection{Usable identity tracks the source-specific gain}

The mechanism analysis compares intrinsic Bayes task identification with nearest-centroid identification in the learned source coordinate. Bayes-ID saturates rapidly, while learned-ID catches up only at much larger $\alpha$. Averaging across relation levels gives the following identity profile. All factorial summaries and regressions in the main text and appendix use the same 42 unrounded five-seed cell means; Spearman correlations use average ranks for ties.

\begin{table}[h]
\centering\small
\caption{Intrinsic versus usable source-task identifiability. Chance is $1/64=.0156$. The collapse gap is Bayes-ID minus learned-ID.}
\begin{tabular}{cccc}
\toprule
$\alpha$ & Bayes-ID & learned-ID & Collapse gap\\
\midrule
0.00 & .016 & .016 & $-.001$\\
0.10 & .400 & .073 & .327\\
0.25 & .953 & .325 & \textbf{.628}\\
0.50 & 1.000 & .708 & .292\\
0.75 & 1.000 & .839 & .161\\
1.00 & 1.000 & .874 & .126\\
1.50 & 1.000 & .914 & .086\\
\bottomrule
\end{tabular}
\end{table}

The source-specific gain tracks usable learned identity more closely than the intrinsic ceiling. Across the 42 cells, $\rho(G_{\rm source},I_{\rm learned})=.98$ versus $\rho(G_{\rm source},I_{\rm Bayes})=.89$. Restricting to the 24 cells with Bayes-ID $>.99$ leaves $\rho(G_{\rm source},I_{\rm learned})=.97$. The following table reports descriptive cell-level regressions.

\begin{table}[h]
\centering\small
\caption{Descriptive GMM regressions predicting $G_{\rm source}$. OLS fits use the 42 unrounded seed-mean cells. Displayed coefficient intervals are the conventional homoskedastic normal approximation $\widehat\beta\pm1.96\,\mathrm{SE}_{\rm OLS}$ with residual variance estimated using $42-p$ degrees of freedom ($p=4$ including the intercept). These descriptive cell-level intervals do not account for matched-seed dependence across cells and are not the hierarchical seed/task intervals for route comparisons.}
\begin{tabular}{lccc}
\toprule
Predictors & Fit $R^2$ & Key coefficient & 95\% CI\\
\midrule
Bayes-ID $+R^2_{\rm rel}+$ interaction & .587 & $I_{\rm Bayes}\!\times R^2_{\rm rel}=+.54$ & $[+.17,+.92]$\\
learned-ID $+R^2_{\rm rel}+$ interaction & \textbf{.843} & $I_{\rm learned}\!\times R^2_{\rm rel}=+.61$ & $[+.33,+.89]$\\
Bayes-ID $+$ learned-ID $+R^2_{\rm rel}$ & .814 & $I_{\rm learned}=+.73$ & $[+.55,+.91]$\\
& & $I_{\rm Bayes}=-.27$ & $[-.44,-.10]$\\
\bottomrule
\end{tabular}
\end{table}

The negative Bayes coefficient in the joint descriptive regression should not be interpreted causally; Bayes-ID and learned-ID share strong variation with $\alpha$. The robust point is that variation remaining in the learned coordinate continues to track the transfer benefit after intrinsic identification has saturated.

The dedicated identity-pipeline retrieval uses an independently sampled retrieval protocol, so its learned-$z_S$ values below differ slightly from the phase-diagram learned-ID summary above; the qualitative ordering is unchanged.

\subsection{Identity through the transport pipeline}

To locate where task information is lost, we retrieve the correct target task at successive stages of the pipeline. Transported-coordinate retrieval asks whether $\widetilde z_T=Az_S+b$ is closest to the correct held-out target coordinate. Raw $z_S\!\to z_T$ retrieval is a cross-domain baseline that omits transport.

\begin{table}[h]
\centering\small
\caption{Mean top-1 task identity through the GMM pipeline, averaged over relation levels and seeds. Chance is $1/64=.0156$.}\label{tab:gmm-identity-chain}
\begin{tabular}{ccccc}
\toprule
$\alpha$ & Bayes-ID & learned $z_S$ & transported $\widetilde z_T$ & raw $z_S\!\to z_T$\\
\midrule
0.00 & .016 & .015 & .013 & .016\\
0.10 & .400 & .071 & .019 & .015\\
0.25 & .953 & .323 & .089 & .023\\
0.50 & 1.000 & .691 & .169 & .039\\
0.75 & 1.000 & .831 & .258 & .074\\
1.00 & 1.000 & .883 & .315 & .109\\
1.50 & 1.000 & .920 & .341 & .148\\
\bottomrule
\end{tabular}
\end{table}

Identity is therefore compressed at both learned stages. Averaged over the grid, transported retrieval is $.172$, compared with $.061$ for raw cross-domain retrieval. The reported overall gain is $+.112$ (paired seed-block interval $[+.106,+.116]$). Mean transported retrieval is close to chance at $\alpha=0$ and $.1$. These summaries do not provide cellwise retrieval intervals. Across the 42 cells, $\rho(G_{\rm source},I_{\widetilde z_T})=.91$. The nearest-centroid margin for learned $z_S$ grows from $-.002$ at $\alpha=0$ to $+.060$ at $\alpha=1.5$, while the transported margin remains negative ($-.039$ to $-.079$), showing that transport improves task alignment without fully recovering the correct target coordinate.

\subsection{Source--target shift magnitude and transport headroom}

The phase diagram varies task identity and relation consistency. A separate sweep holds $\alpha=1,\eta=0$ fixed and changes only the magnitude of the shared transformation,
\[
T_\lambda(u)=u+\lambda(T_0(u)-u),
\qquad
\lambda\in\{0,.25,.5,1,2\}.
\]
This separates the value of choosing the correct source from the value of transporting that source.

\begin{table}[h]
\centering\small
\caption{GMM shift-magnitude sweep: route means averaged over tasks and five seeds. $G_{\rm source}=\mathrm{SW}_{\rm wrong}-\mathrm{SW}_{\rm correct}$; transport-over-reuse is $\mathrm{SW}_{\rm reuse}-\mathrm{SW}_{\rm correct}$.}\label{tab:gmm-shift-means}
\begin{tabular}{ccccccc}
\toprule
$\lambda$ & Correct & Wrong & Relation-only & Source reuse & $G_{\rm source}$ & Transport over reuse\\
\midrule
0.00 & .759 & 1.241 & 1.062 & .783 & +.482 & +.024\\
0.25 & .765 & 1.252 & 1.048 & .828 & +.487 & +.063\\
0.50 & .802 & 1.315 & 1.115 & .905 & +.513 & +.102\\
1.00 & .835 & 1.355 & 1.157 & 1.066 & +.520 & +.231\\
2.00 & 1.037 & 1.667 & 1.443 & 1.577 & +.631 & +.541\\
\bottomrule
\end{tabular}
\end{table}

The source-specific gain remains large across the sweep because the semantic source task is identifiable even when $\lambda=0$. The mean transport-over-reuse advantage grows from $.024$ at $\lambda=0$ to $.541$ at $\lambda=2$. These shift-sweep comparisons are descriptive mean differences. The $\lambda=1$ condition repeats the $(\alpha=1,\eta=0)$ factorial cell with independently trained models, so its means differ slightly from Table~\ref{tab:gmm-sw-absolute}. The independent model-free source--target SW headroom increases from approximately $.08$ to $1.59$ over the same sweep. This dissociation shows that source specificity and transport utility answer different questions: a correct source can matter even when there is almost nothing to transport.

\subsection{Historical representation-capacity and relatedness diagnostics}
The earlier GMM suite (run before the final protocol) is retained because it probes the representational ceiling, but it is not used as evidence for the final correct source $+$ transport route. Its aggregate summaries lack the paired task-level records needed to recover new confidence intervals. Here the historical oracle is an optimised coordinate, not the frozen-encoder reference in the factorial experiment. In the relatedness reference protocol, it uses 2,048 target samples, starts at the transported coordinate, and applies 500 Adam updates at learning rate $0.01$ without a coordinate prior, keeping the generator frozen. The abundant-target full-FT reference updates the generator and reports the smallest mean score-field error over $(N_{\rm FT},\mathrm{lr})\in\{(500,3{\times}10^{-4}),(2000,3{\times}10^{-4}),(8000,3{\times}10^{-4}),(8000,10^{-4})\}$. This reference protocol uses six held-out tasks and 1,024 score-evaluation points per task. These are privileged, budget-selected diagnostics, not guaranteed performance upper bounds.

\begin{table}[h]
\centering\small
\caption{Historical GMM representation-capacity diagnostic. The oracle fits an abundant-target coordinate; captured fraction is $(e_0-e_{\rm oracle})/(e_0-e_{\rm FT})$.}
\begin{tabular}{lrrrr}
\toprule
Configuration & No adapt. & Oracle & Full FT ref. & Captured\\
\midrule
$k=16$, 80k & .8821 & .6320 & .1276 & 33.2\%\\
$k=16$, 320k & .8741 & .5965 & .1508 & 38.4\%\\
$k=64$, 80k & .8868 & .5396 & .1267 & 45.7\%\\
$k=64$, 320k & .8779 & .4690 & .1565 & 56.7\%\\
\bottomrule
\end{tabular}
\end{table}

\begin{table}[h]
\centering\small
\caption{Historical GMM task-relatedness diagnostic at $k=16$, 80k steps. Larger template perturbations produce less related tasks.}
\begin{tabular}{lrrrr}
\toprule
Task family & No adapt. & Oracle & Best-grid FT & Captured\\
\midrule
Perturbation .10 & .7513 & .3835 & .1179 & 58.1\%\\
Perturbation .25 & .8910 & .5583 & .1220 & 43.3\%\\
Perturbation .50 & .9516 & .6949 & .1359 & 31.5\%\\
Perturbation 1.00 & .9679 & .7808 & .1239 & 22.2\%\\
\bottomrule
\end{tabular}
\end{table}

These diagnostics support the intuitive claim that a compact shared representation has a higher ceiling when tasks are more closely related. Because the runs predate the final deployment protocol and report aggregate rather than paired route statistics, they are deliberately separated from the main claim.

\section{CIFAR-100: Full Controls and Ablations}
\label{app:cifar}

\subsection{Protocol}
The official CIFAR-100 training and test images are pooled before the custom semantic-task split; original split and row identifiers are retained for provenance. Entire superclasses are assigned to train, validation or test so that close sibling classes do not cross the split. Twelve superclasses (60 fine classes) train the model, four superclasses (20 classes) are used for validation and hyperparameter selection, and four superclasses (20 classes) are held for final evaluation. A fixed permutation creates disjoint source and target pools: 300 source-support images, 50 source-query images, 20 reserved target-support images and 230 target-query images per class. Target transformations are Gaussian blur ($\sigma=1.0$), additive Gaussian noise (standard deviation $0.18$ on the normalised scale), and contrast reduction (factor $0.45$ toward the image mean), all at severity 3. The 300/50/20/230 counts are pool capacities. The main generation evaluations produce 96 samples and compare against 128 sampled target queries per episode. Pixel SW uses 256 unit-normal Gaussian projections with projection seed 0 on flattened $[-1,1]$ images. Energy MMD is the energy-distance V-statistic $2\overline{\|a-b\|}-\overline{\|a-a'\|}-\overline{\|b-b'\|}$, including within-sample diagonals, using the first 64 generated and 64 query images. Class consistency is the share of generated samples that a separately trained 20-way classifier assigns to the episode's own fine class. The classifier is a small convolutional network (three strided convolutional blocks with GroupNorm, global average pooling and a linear head) trained on real images of the 20 fine classes of the evaluated split, each batch shown clean or under one of the three corruptions; on 50 withheld real images per class its fine-class top-1 accuracy is 38--40\% for the validation classes (chance 5\%). Gaussian-noise draws are deterministic functions of image identifiers. Wrong-source controls choose the next distinct class within the same corruption's episode order, retaining the relation. Protocols for independent diagnostics are specified separately below.

\paragraph{Meaning of the evaluation packages.}
In all headline generation packages, $J=0$ denotes no target-time coordinate-refinement steps for correct-source transport; $\KT$ is the target-support budget made available to adapting baselines, not a refinement-step count. Table~\ref{tab:cifar-main}A uses 60 validation class--transformation episodes at $\KT=0$; Table~\ref{tab:cifar-main}B uses 60 final-test episodes at each of $\KT=1,5,20$. The source-dependence sweep is a separate 180-episode validation package. The conditioning comparison and the two transport-structure comparisons are three separately matched packages with one training seed per arm; the FiLM conditioning arm and the FiLM-linear arm are the same trained model, whereas the two basis arms are trained separately. Rows are not paired across packages. The interpolation profile, deployment grid and refinement diagnostics use the same 50k \texttt{cifar\_ds3} checkpoint as the headline packages. The frozen FiLM retrieval/direct-target diagnostic below is another independent package and uses its own support resampling, post-hoc target split and semantic evaluator.
For the generation tables, intervals are percentile 95\% class-cluster bootstrap intervals (10,000 resamples; seed 20261020). Each resample retains the three corruptions of a sampled class, averages within class and then across 20 classes; paired contrasts use the same sampled classes. Entries report mean [lower, upper], rather than symmetric half-widths. A star marks an unrounded paired interval excluding zero in either direction. These pointwise intervals are not multiplicity-adjusted and do not cover training-seed variation. Since test classes are drawn from only four superclasses, these intervals also do not capture across-superclass variation. Independent retrieval and direct-target diagnostics retain their explicitly stated resampling units.

\subsection{Source specificity from zero to twenty targets}
This independent validation package adds mean-source and shuffled-source controls that the final-test comparison (Table~\ref{tab:cifar-main}B) lacks. 9 of nine main-relative source-control SW intervals favour correct source $+$ transport. Target-only comparisons favour transport at $K_T=1$; remain unresolved at $K_T=5,20$. Rows labelled source-initialised adaptation optimise a source-initialised coordinate using target support; they do not implement direct target-free reuse.

\begin{table}[h]
\centering\small
\caption{CIFAR validation source-dependence sweep: 180 episodes, 60 per budget. Main transport and mean/shuffled/relation controls use $J=0$. Source-initialised adaptation uses target support and is not strict reuse. SW is lower-is-better; paired $\Delta$ is comparator minus main. Entries show mean [95\% class-cluster CI] with unrounded paired stars.}
\begin{tabular}{lccc}
\toprule
Route / comparison & $\KT=1$ & $\KT=5$ & $\KT=20$\\
\midrule
\shortstack[l]{correct source\\$+$ transport} & \shortstack{$0.2071$\\$[0.1797,0.2421]$} & \shortstack{$0.2077$\\$[0.1831,0.2380]$} & \shortstack{$0.2129$\\$[0.1854,0.2471]$}\\
Source-init. adaptation & \shortstack{$0.2436$\\$[0.2186,0.2756]$} & \shortstack{$0.2351$\\$[0.2108,0.2652]$} & \shortstack{$0.2203$\\$[0.1901,0.2597]$}\\
Target-only & \shortstack{$0.2255$\\$[0.1986,0.2605]$} & \shortstack{$0.2130$\\$[0.1858,0.2469]$} & \shortstack{$0.2087$\\$[0.1781,0.2481]$}\\
Mean source & \shortstack{$0.2309$\\$[0.2019,0.2667]$} & \shortstack{$0.2347$\\$[0.2073,0.2663]$} & \shortstack{$0.2379$\\$[0.2080,0.2731]$}\\
Shuffled source & \shortstack{$0.2417$\\$[0.2113,0.2779]$} & \shortstack{$0.2453$\\$[0.2165,0.2774]$} & \shortstack{$0.2469$\\$[0.2155,0.2827]$}\\
Relation-only & \shortstack{$0.2413$\\$[0.2126,0.2764]$} & \shortstack{$0.2427$\\$[0.2158,0.2739]$} & \shortstack{$0.2476$\\$[0.2183,0.2818]$}\\
\midrule
$\Delta$ Source-init. adaptation & \shortstack{$0.0366\sigstar$\\$[0.0303,0.0427]$} & \shortstack{$0.0274\sigstar$\\$[0.0221,0.0330]$} & \shortstack{$0.0074$\\$[-0.0019,0.0168]$}\\
$\Delta$ Target-only & \shortstack{$0.0184\sigstar$\\$[0.0080,0.0294]$} & \shortstack{$0.0053$\\$[-0.0016,0.0123]$} & \shortstack{$-0.0043$\\$[-0.0157,0.0076]$}\\
$\Delta$ Mean source & \shortstack{$0.0239\sigstar$\\$[0.0135,0.0351]$} & \shortstack{$0.0270\sigstar$\\$[0.0155,0.0398]$} & \shortstack{$0.0250\sigstar$\\$[0.0136,0.0374]$}\\
$\Delta$ Shuffled source & \shortstack{$0.0347\sigstar$\\$[0.0192,0.0518]$} & \shortstack{$0.0376\sigstar$\\$[0.0197,0.0573]$} & \shortstack{$0.0340\sigstar$\\$[0.0171,0.0530]$}\\
$\Delta$ Relation-only & \shortstack{$0.0342\sigstar$\\$[0.0231,0.0462]$} & \shortstack{$0.0350\sigstar$\\$[0.0239,0.0474]$} & \shortstack{$0.0347\sigstar$\\$[0.0224,0.0475]$}\\
\bottomrule
\end{tabular}
\end{table}

\begin{figure}[t]
\centering
\includegraphics[width=.88\linewidth]{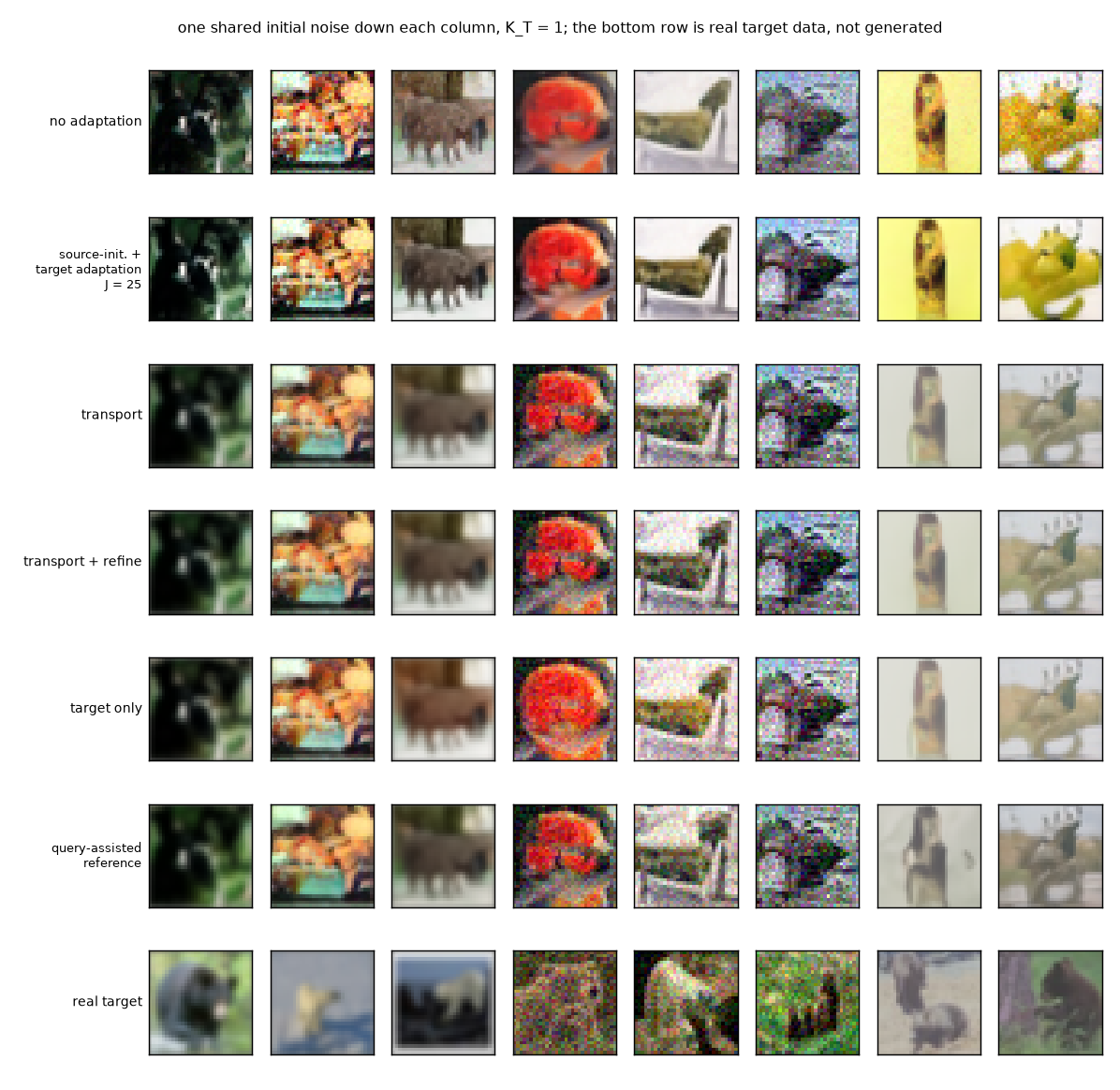}
\caption{CIFAR $K_T=1$ deployment grid from the same 50k-step checkpoint as the headline packages. The seven rows show a zero-coordinate control ($z=0$, no adaptation), source-initialised target adaptation ($J=25$), correct source $+$ transport ($J=0$), transport plus refinement, target-only, a query-assisted target-coordinate reference, and real target. The source-initialised route uses target support and is not strict reuse. The query-assisted reference starts at the transported coordinate and optimises the coordinate on the same target-query set used as its reference, without a coordinate prior, for 25 steps; its frozen generator is unchanged. It is a privileged diagnostic, not a deployable baseline or a performance upper bound. Generated rows retain the original samples and paired noise; the bottom row is real target data.}
\label{fig:cifar-strategy-grid}
\end{figure}

The final-test package of Table~\ref{tab:cifar-main}B gives energy MMD $2.3306,2.3386,2.4041$ for correct source $+$ transport, $3.4434,2.5351,2.4137$ for target-only, and $5.8971,4.0219,1.9086$ for full fine-tuning at $K_T=1,5,20$. Paired full-FT-minus-main differences are $3.5666\, [2.5424,4.6245]\sigstar$, $1.6833\, [0.9979,2.3836]\sigstar$, $-0.4955\, [-1.0792,0.0386]$, favouring transport at $K_T=1,5$ and remaining unresolved at $K_T=20$. Paired target-only-minus-main differences are $1.1129\, [0.5787,1.7474]\sigstar$, $0.1965\, [-0.0146,0.4128]$, $0.0096\, [-0.2037,0.2111]$, favouring transport at $K_T=1$ and remaining unresolved at $K_T=5,20$. These are paired class-cluster intervals on the same 60 episodes per budget and are not inferred from SW.

\subsection{Validation-frozen full fine-tuning and coordinate refinement}
\label{app:cifar-selection}

\paragraph{Identity of the current CIFAR result packages.}
The archived source-dependence (A1), source-budget (A4), zero-target (A6) and final-test (A9) outputs identify the same 50k-step basis-conditioned \texttt{cifar\_ds3} EMA checkpoint, with linear transport and $k=16$ (recorded SHA256 prefix \texttt{be2b86b158cf}). Its training seed is 12345; these evaluations use seed 4321, 96 generated samples per episode and 50-step DDIM. A1 and A6 use $M_S=64$ with independently drawn source sets; A4 varies $M_S$ using nested prefixes. The separate frozen FiLM diagnostics below use their own checkpoint and protocol.

\paragraph{Coordinate updates in the validation controls.}
The A1 and A4 reports log $J=25$, coordinate learning rate $0.01$ and coordinate-prior scale $\beta_0=1.0$ (penalty $(\beta_0/\KT)\|z-z_0\|_2^2$ added to the support denoising loss, where $z_0$ is the route's initial coordinate) for their target-using coordinate routes; the corresponding checkpoint configuration records a noise-batch setting of 16. Target-only starts from the target-support encoding, while source-initialised adaptation starts from $z_S$. Correct-source transport remains at $J=0$. These validation-control settings are separate from the full-FT budgets below.

The full-fine-tuning settings were selected on validation tasks and frozen before correct source $+$ transport was evaluated on the final-test classes. They remain 25 steps at learning rate $10^{-5}$ for $\KT=1$, 25 at $10^{-4}$ for $\KT=5$, and 400 at $10^{-5}$ for $\KT=20$. No test result selects a new budget or model.

\begin{table}[h]
\centering\small
\caption{CIFAR validation-selected full-fine-tuning settings frozen before the matched final-test evaluation.}
\label{tab:cifar-ft-selection}
\begin{tabular}{ccc}
\toprule
Target budget & Full-FT steps & Learning rate\\
\midrule
$\KT=1$ & 25 & $10^{-5}$\\
$\KT=5$ & 25 & $10^{-4}$\\
$\KT=20$ & 400 & $10^{-5}$\\
\bottomrule
\end{tabular}
\end{table}

The five-budget A3 refinement experiment compares matched on/off conditions on the 50k-step \texttt{cifar\_ds3} checkpoint. It uses 300 validation episodes (60 at each $K_T$), matched target support/query sets and paired refinement/generation noise. The metric is transformation-statistic distance. Gain is transport alone minus transport plus 25 refinement steps; positive favours refinement. This separate diagnostic is not pooled with final-test SW or MMD.

\begin{table}[h]
\centering\small
\caption{Matched refinement ablation on the correct source $+$ transport checkpoint. $J=0$ remains the main route; the ablation uses $J=25$, learning rate $0.01$ and the same prior/noise budget. Entries report paired 95\% class-cluster intervals; stars use unrounded gain intervals.}
\label{tab:cifar-refinement}
\begin{tabular}{cc}
\toprule
$\KT$ & Refinement gain [95\% CI]\\
\midrule
1 & $-0.0244\, [-0.0365,-0.0120]\sigstar$\\
2 & $-0.0071\, [-0.0249,0.0116]$\\
5 & $-0.0267\, [-0.0468,-0.0061]\sigstar$\\
10 & $0.0144\, [-0.0169,0.0459]$\\
20 & $-0.0065\, [-0.0511,0.0380]$\\
\bottomrule
\end{tabular}
\end{table}

The gain intervals are negative at $K_T=1,5$: refinement worsens transformation-statistic distance in those comparisons. The other three intervals contain zero; this establishes neither equivalence nor universal absence of benefit. The separate source-initialised target-adaptation baseline starts from $z_S$ and uses target evidence. Its behaviour cannot be inferred from the post-transport gains in this table, and it must not be called strict direct reuse.

\subsection{Source budget and conditioning generality}
The measured source-budget changes between $M_S=16$ and 300 are small, but the paired SW intervals relative to 16 all contain zero (Table~\ref{tab:cifar-source-budget}). At $M_S=64$, the difference is $-0.00287\, [-0.00655,0.00089]$. This does not establish equivalence or an exact plateau; no $M_S<16$ was tested. The model was trained with $M_S=64$ and evaluated at $K_T=1$ with nested source prefixes and matched query/noise. Proposition~\ref{prop:source-concentration} concerns coordinate estimation error, not equivalence of generated SW at all source sizes.

\begin{table}[h]
\centering\small
\caption{CIFAR validation source-budget sweep for correct source $+$ transport ($\KT=1$, $J=0$, 60 episodes; nested source prefixes; checkpoint trained at $M_S=64$). Entries are mean SW [95\% class-cluster CI]; the paired difference is the given $M_S$ minus $M_S=16$ on the same episodes, queries and generation noise. No paired interval excludes zero.}
\label{tab:cifar-source-budget}
\begin{tabular}{ccc}
\toprule
$M_S$ & SW & Paired difference vs.\ $M_S=16$\\
\midrule
16 & $0.2118\,[0.1846,0.2435]$ & --\\
32 & $0.2110\,[0.1839,0.2429]$ & $-0.00080\,[-0.00372,0.00224]$\\
64 & $0.2089\,[0.1813,0.2421]$ & $-0.00287\,[-0.00655,0.00089]$\\
128 & $0.2096\,[0.1818,0.2430]$ & $-0.00212\,[-0.00601,0.00207]$\\
256 & $0.2097\,[0.1814,0.2435]$ & $-0.00205\,[-0.00660,0.00299]$\\
300 & $0.2097\,[0.1812,0.2436]$ & $-0.00207\,[-0.00674,0.00305]$\\
\bottomrule
\end{tabular}
\end{table}

In the independent 180-episode FiLM validation package, SW for correct source $+$ transport is $0.1973, 0.1942, 0.1987$ and class consistency is $0.1306, 0.1278, 0.1311$ at $K_T=1,5,20$. Mean-source SW is $0.2417, 0.2441, 0.2462$, shuffled-source $0.2567, 0.2578, 0.2616$, and relation-only $0.2366, 0.2373, 0.2414$. All 9/9 SW and 9/9 semantic main-relative intervals favour the correct source. Paired consistency gains are $0.0740\, [0.0410,0.1161]\sigstar$, $0.0786\, [0.0431,0.1257]\sigstar$, $0.0780\, [0.0441,0.1217]\sigstar$ against mean source, $0.0835\, [0.0517,0.1252]\sigstar$, $0.0799\, [0.0436,0.1286]\sigstar$, $0.0835\, [0.0488,0.1292]\sigstar$ against shuffled source, and $0.0757\, [0.0451,0.1141]\sigstar$, $0.0773\, [0.0443,0.1219]\sigstar$, $0.0773\, [0.0470,0.1149]\sigstar$ against relation-only. These source controls do not test the full FiLM target-only/full-FT crossover.

\begin{table}[h]
\centering\small
\caption{Three separately matched CIFAR SW packages; the FiLM conditioning and FiLM-linear rows report the same trained model. Upper block: correct source $+$ transport with two conditioning architectures. Lower blocks: transport-structure ablations, with direct $J=0$ generation in both compared arms. Pairing is within block only; intervals resample validation classes at one training seed per arm.}
\begin{tabular}{lccc}
\toprule
Configuration / comparison & $\KT=1$ & $\KT=5$ & $\KT=20$\\
\midrule
\shortstack[l]{Basis: correct source\\$+$ transport} & \shortstack{$0.2067$\\$[0.1793,0.2414]$} & \shortstack{$0.2090$\\$[0.1834,0.2391]$} & \shortstack{$0.2136$\\$[0.1864,0.2469]$}\\
\shortstack[l]{FiLM: correct source\\$+$ transport} & \shortstack{$0.1973$\\$[0.1689,0.2355]$} & \shortstack{$0.1942$\\$[0.1709,0.2239]$} & \shortstack{$0.1987$\\$[0.1720,0.2334]$}\\
FiLM $-$ basis & \shortstack{$-0.0094$\\$[-0.0238,0.0046]$} & \shortstack{$-0.0148$\\$[-0.0298,0.0003]$} & \shortstack{$-0.0149$\\$[-0.0300,0.0003]$}\\
\midrule
Basis residual MLP & \shortstack{$0.2102$\\$[0.1839,0.2451]$} & \shortstack{$0.2109$\\$[0.1879,0.2392]$} & \shortstack{$0.2138$\\$[0.1870,0.2481]$}\\
Basis linear & \shortstack{$0.2124$\\$[0.1829,0.2516]$} & \shortstack{$0.2113$\\$[0.1859,0.2442]$} & \shortstack{$0.2174$\\$[0.1886,0.2550]$}\\
MLP $-$ linear & \shortstack{$-0.0022$\\$[-0.0080,0.0031]$} & \shortstack{$-0.0004$\\$[-0.0068,0.0056]$} & \shortstack{$-0.0036$\\$[-0.0097,0.0019]$}\\
\midrule
FiLM residual MLP & \shortstack{$0.2066$\\$[0.1759,0.2489]$} & \shortstack{$0.2054$\\$[0.1795,0.2395]$} & \shortstack{$0.2074$\\$[0.1777,0.2465]$}\\
FiLM linear & \shortstack{$0.1973$\\$[0.1689,0.2355]$} & \shortstack{$0.1942$\\$[0.1709,0.2239]$} & \shortstack{$0.1987$\\$[0.1720,0.2334]$}\\
MLP $-$ linear & \shortstack{$0.0093\sigstar$\\$[0.0044,0.0147]$} & \shortstack{$0.0112\sigstar$\\$[0.0057,0.0175]$} & \shortstack{$0.0087\sigstar$\\$[0.0026,0.0151]$}\\
\bottomrule
\end{tabular}
\end{table}

In the retained FiLM transport-structure ablation, residual-MLP-minus-linear MMD differences are $0.2479\, [0.0585,0.5412]\sigstar$, $0.2810\, [0.0976,0.5549]\sigstar$, $0.1958\, [-0.0144,0.4870]$, favouring linear transport at $K_T=1,5$ and unresolved at $K_T=20$. They are paired effects, not raw MMD values. The matched conditioning SW comparison remains unresolved at all three budgets. The basis and FiLM conditioning counts remain 55,344 and 118,784 (46.6\% ratio). Their evaluated score paths have 27,798,195 and 27,861,635 parameters, respectively; the shared transport module adds the same 352 registered parameters (304 active). No latency, FLOP or memory conclusion is inferred.

\subsection{Meta-training-task scarcity}
Both task-scarcity arms use correct source $+$ transport and $J=0$, with 50\% (30/60 classes, all 12 superclasses) and 25\% training subsets. At 50\%, FiLM-minus-basis SW effects are $-0.0098\, [-0.0268,0.0062]$, $-0.0102\, [-0.0280,0.0067]$, $-0.0104\, [-0.0293,0.0078]$; all contain zero. MMD effects are $-0.3484\, [-0.7712,0.0456]$, $-0.5448\, [-1.0850,-0.0527]\sigstar$, $-0.5468\, [-1.1630,0.0169]$, favouring FiLM at $K_T=5$ only. Superclass-consistency effects are $0.0363\, [0.0050,0.0675]\sigstar$, $0.0380\, [0.0050,0.0693]\sigstar$, $0.0335\, [0.0030,0.0635]\sigstar$, favouring FiLM at all three budgets. At 25\%, SW effects are $0.0027\, [-0.0079,0.0144]$, $0.0007\, [-0.0108,0.0129]$, $-0.0003\, [-0.0118,0.0113]$, all unresolved. These SW intervals establish neither an architecture ranking nor a ranking reversal. Each arm has one training seed; the class-cluster intervals do not cover training-seed uncertainty.

\subsection{Coordinate retrieval and direct-target diagnostics}
The frozen 50k EMA FiLM model with affine transport (checkpoint hash prefix \texttt{f7c1369c}) was probed on 60 seen (meta-training) and 20 held-out (not used in meta-training) fine classes. Retrieval uses 20 fixed support resamples, with 64 source and 64 target images per class in each resample; candidates are confined to the corresponding seen or held-out class group. This diagnostic support pool is distinct from the default 20-image target-support pool used for the headline budget comparison. Its bootstrap resamples the 20 support draws conditional on the same classes, so significance describes stability to support choice and does not establish population-level class significance. With 20 held-out candidates, correct-source transport achieves L2 retrieval of 62.5\%, 59.5\%, and 49.0\% for blur, noise, and contrast, respectively; raw $z_S$ achieves 17.25\%, 26.5\%, and 14.0\%, and wrong-source transport achieves 1.25\%, .75\%, and 2.25\%. Tables~\ref{tab:cifar-retrieval-full} and \ref{tab:cifar-retrieval-paired} give retrieval rates, margins and paired contrasts for the reported class groups and transformations.

\begin{table}[htbp]
\centering\scriptsize\setlength{\tabcolsep}{3pt}
\caption{Complete CIFAR frozen-coordinate retrieval. Each cell is top-1 percentage / mean nearest-wrong margin. Positive margins favour the requested class; relation-only chance/tie behaviour is not task discrimination.}\label{tab:cifar-retrieval-full}
\begin{tabular}{lllrrrr}
\toprule
Group & Relation & Metric & Raw $z_S$ & Main & Wrong-source + transport & Relation only\\
\midrule

Seen & Blur & L2 & $36.50/-0.01050$ & $71.50/+0.02241$ & $0.42/-0.24324$ & $1.67/-0.13592$\\
 & & Cosine & $26.58/-0.06849$ & $72.58/+0.01675$ & $0.42/-0.24228$ & $1.67/-0.10330$\\
Seen & Noise & L2 & $56.83/+0.00548$ & $74.50/+0.02872$ & $0.33/-0.26493$ & $1.67/-0.13956$\\
 & & Cosine & $30.50/-0.04843$ & $78.08/+0.01791$ & $0.25/-0.24565$ & $1.67/-0.09906$\\
Seen & Contrast & L2 & $36.00/-0.01447$ & $68.83/+0.02719$ & $0.50/-0.26657$ & $1.67/-0.12781$\\
 & & Cosine & $25.58/-0.07178$ & $72.08/+0.02048$ & $0.67/-0.28837$ & $1.67/-0.11249$\\
\midrule
Held-out & Blur & L2 & $17.25/-0.02427$ & $62.50/+0.01114$ & $1.25/-0.12900$ & $5.00/-0.08812$\\
 & & Cosine & $16.75/-0.13231$ & $64.50/+0.00806$ & $1.25/-0.11517$ & $5.00/-0.07663$\\
Held-out & Noise & L2 & $26.50/-0.01098$ & $59.50/+0.00985$ & $0.75/-0.11476$ & $5.00/-0.08370$\\
 & & Cosine & $15.50/-0.09900$ & $60.00/+0.00693$ & $0.00/-0.09392$ & $5.00/-0.06372$\\
Held-out & Contrast & L2 & $14.00/-0.04431$ & $49.00/-0.00423$ & $2.25/-0.09975$ & $5.00/-0.07584$\\
 & & Cosine & $10.25/-0.16097$ & $48.25/-0.00313$ & $2.00/-0.09921$ & $5.00/-0.07595$\\
\bottomrule
\end{tabular}
\end{table}
\begin{table}[htbp]
\centering\scriptsize\setlength{\tabcolsep}{3pt}
\caption{CIFAR L2 retrieval gains in percentage points, main minus comparator. Intervals bootstrap the 20 paired support resamples conditional on the fixed class set; positive favours correct-source transport.}
\label{tab:cifar-retrieval-paired}
\begin{tabular}{llrr}
\toprule
Group & Relation & Main $-$ raw source (pp) [95\% CI] & Main $-$ wrong source (pp) [95\% CI]\\
\midrule
Seen & Blur & $+35.00\,[+32.25,+37.67]\sigstar$ & $+71.08\,[+69.08,+73.08]\sigstar$\\
Seen & Noise & $+17.67\,[+15.33,+20.00]\sigstar$ & $+74.17\,[+72.25,+75.92]\sigstar$\\
Seen & Contrast & $+32.83\,[+30.67,+35.08]\sigstar$ & $+68.33\,[+66.58,+70.08]\sigstar$\\
Held-out & Blur & $+45.25\,[+42.25,+48.25]\sigstar$ & $+61.25\,[+58.25,+64.00]\sigstar$\\
Held-out & Noise & $+33.00\,[+27.25,+38.25]\sigstar$ & $+58.75\,[+54.50,+63.25]\sigstar$\\
Held-out & Contrast & $+35.00\,[+30.25,+39.50]\sigstar$ & $+46.75\,[+43.00,+50.50]\sigstar$\\
\bottomrule
\end{tabular}
\end{table}
\paragraph{Direct-target generation and timestep probes.}
For direct-$z_T$ generation, a separate post-hoc repartition supplies 200 target-support images and 50 disjoint real target queries per class; it does not reuse the headline 20/230 target-support/query partition. Each class has 96 generated samples using 50-step DDIM. The frozen DINO nearest-prototype evaluator ranks all 80 seen-plus-held-out classes and is calibrated on the diagnostic real queries. These identity percentages are not directly comparable with headline semantic consistency because both the partition and the identity instrument differ. Generation and timestep intervals resample classes, not the fixed support draws used for retrieval. Tables~\ref{tab:cifar-direct-target} and \ref{tab:cifar-direct-target-contrast} report the raw metrics and the descriptive held-out-minus-seen identity contrasts.
\begin{table}[htbp]
\centering\scriptsize\setlength{\tabcolsep}{3pt}
\caption{CIFAR direct-target generation using abundant target support. Identity rates are percentages; SW and MMD use the diagnostic DINO feature space, not the headline pixel-space metrics.}
\label{tab:cifar-direct-target}
\begin{tabular}{llrrrr}
\toprule
Group & Relation & Generated identity (\%) & Real identity (\%) & DINO-SW $\downarrow$ & DINO-MMD $\downarrow$\\
\midrule
Seen & Blur & 5.52 & 50.80 & 0.02176 & 0.15937\\
Seen & Noise & 3.78 & 29.27 & 0.01671 & 0.09777\\
Seen & Contrast & 6.84 & 78.10 & 0.02983 & 0.27962\\
Held-out & Blur & 3.59 & 51.20 & 0.02338 & 0.18884\\
Held-out & Noise & 3.85 & 34.50 & 0.01888 & 0.13402\\
Held-out & Contrast & 2.71 & 70.90 & 0.03059 & 0.28967\\
\bottomrule
\end{tabular}
\end{table}
\begin{table}[htbp]
\centering\scriptsize\setlength{\tabcolsep}{3pt}
\caption{CIFAR direct-target generated-identity differences, held-out minus seen, in percentage points. The two different class groups are independently resampled; these are descriptive group contrasts, not paired causal effects.}
\label{tab:cifar-direct-target-contrast}
\begin{tabular}{lr}
\toprule
Relation & Identity $\Delta$ (pp) [95\% CI]\\
\midrule
Blur & $-1.93\,[-6.70,+4.11]$\\
Noise & $+0.07\,[-4.34,+6.18]$\\
Contrast & $-4.13\,[-7.99,+0.14]$\\
\bottomrule
\end{tabular}
\end{table}
\FloatBarrier
The timestep probe pairs correct and cyclic wrong-class target coordinates at identical real query, noise and timestep. The five bins contain 200 diffusion steps each, with 800--999 the highest-noise bin. In Table~\ref{tab:cifar-time-probe}, positive $\Delta_z=\mathcal L_\epsilon(z_{\rm wrong})-\mathcal L_\epsilon(z_{\rm correct})$ favours the correct target coordinate; this does not specifically test transport. MSE values are unscaled. Stars follow the source report's unrounded class-bootstrap limits, including limits displayed as zero after rounding.
\begin{table}[htbp]
\centering\scriptsize\setlength{\tabcolsep}{3pt}
\caption{CIFAR coordinate use across diffusion time: absolute denoising MSE and wrong-minus-correct differences with class-bootstrap 95\% intervals.}\label{tab:cifar-time-probe}
\begin{tabular}{lllrrl}
\toprule
Group & Relation & Timesteps & Correct MSE & Wrong MSE & $\Delta_z$ [95\% CI]\\
\midrule

Seen & Blur & 0--199 & 0.069118 & 0.069843 & $+0.00073\,[+0.00058,+0.00089]\sigstar$\\
 & & 200--399 & 0.025385 & 0.026199 & $+0.00081\,[+0.00065,+0.00099]\sigstar$\\
 & & 400--599 & 0.014950 & 0.015695 & $+0.00074\,[+0.00061,+0.00089]\sigstar$\\
 & & 600--799 & 0.008320 & 0.008864 & $+0.00054\,[+0.00045,+0.00065]\sigstar$\\
 & & 800--999 & 0.002333 & 0.002566 & $+0.00023\,[+0.00019,+0.00028]\sigstar$\\
\midrule
Seen & Noise & 0--199 & 0.575592 & 0.576098 & $+0.00051\,[+0.00040,+0.00062]\sigstar$\\
 & & 200--399 & 0.149464 & 0.150564 & $+0.00110\,[+0.00087,+0.00135]\sigstar$\\
 & & 400--599 & 0.055730 & 0.056779 & $+0.00105\,[+0.00083,+0.00129]\sigstar$\\
 & & 600--799 & 0.020958 & 0.021725 & $+0.00077\,[+0.00063,+0.00092]\sigstar$\\
 & & 800--999 & 0.004284 & 0.004563 & $+0.00028\,[+0.00023,+0.00033]\sigstar$\\
\midrule
Seen & Contrast & 0--199 & 0.097140 & 0.098230 & $+0.00109\,[+0.00084,+0.00137]\sigstar$\\
 & & 200--399 & 0.025000 & 0.026042 & $+0.00104\,[+0.00082,+0.00130]\sigstar$\\
 & & 400--599 & 0.011904 & 0.012583 & $+0.00068\,[+0.00055,+0.00083]\sigstar$\\
 & & 600--799 & 0.005251 & 0.005642 & $+0.00039\,[+0.00032,+0.00047]\sigstar$\\
 & & 800--999 & 0.001187 & 0.001322 & $+0.00014\,[+0.00011,+0.00016]\sigstar$\\
\midrule
Held-out & Blur & 0--199 & 0.079225 & 0.079248 & $+0.00002\,[-0.00008,+0.00012]$\\
 & & 200--399 & 0.030886 & 0.030951 & $+0.00006\,[+0.00000,+0.00013]\sigstar$\\
 & & 400--599 & 0.018033 & 0.018145 & $+0.00011\,[+0.00005,+0.00018]\sigstar$\\
 & & 600--799 & 0.009747 & 0.009893 & $+0.00015\,[+0.00005,+0.00026]\sigstar$\\
 & & 800--999 & 0.002668 & 0.002732 & $+0.00006\,[+0.00002,+0.00012]\sigstar$\\
\midrule
Held-out & Noise & 0--199 & 0.584385 & 0.584397 & $+0.00001\,[-0.00004,+0.00006]$\\
 & & 200--399 & 0.157766 & 0.157801 & $+0.00004\,[-0.00005,+0.00013]$\\
 & & 400--599 & 0.060890 & 0.060990 & $+0.00010\,[+0.00005,+0.00016]\sigstar$\\
 & & 600--799 & 0.023057 & 0.023172 & $+0.00012\,[+0.00004,+0.00021]\sigstar$\\
 & & 800--999 & 0.004634 & 0.004695 & $+0.00006\,[+0.00002,+0.00011]\sigstar$\\
\midrule
Held-out & Contrast & 0--199 & 0.111840 & 0.111861 & $+0.00002\,[-0.00011,+0.00016]$\\
 & & 200--399 & 0.030179 & 0.030231 & $+0.00005\,[-0.00006,+0.00015]$\\
 & & 400--599 & 0.013888 & 0.013959 & $+0.00007\,[-0.00000,+0.00015]$\\
 & & 600--799 & 0.005905 & 0.005997 & $+0.00009\,[+0.00003,+0.00017]\sigstar$\\
 & & 800--999 & 0.001368 & 0.001395 & $+0.00003\,[+0.00001,+0.00005]\sigstar$\\
\bottomrule
\end{tabular}
\end{table}

Bypassing transport does not guarantee semantic generation. Direct target coordinates estimated from 200 target images give held-out generated identity of 3.59\% (blur), 3.85\% (noise), and 2.71\% (contrast), compared with real-query identity of 51.2\%, 34.5\%, and 70.9\%. Diffusion-time probes nevertheless show that correct target coordinates lower denoising MSE in 9 of 15 held-out relation--timestep bins, including the highest-noise bin (800--999) for all three relations. The evidence therefore supports coordinate use without implying that coordinate use alone guarantees class-correct generation.

\section{Office-Home: Full Natural-Domain Stress Test}
\label{app:office}

\subsection{Protocol and evaluator audit}
The Office-Home experiment uses the Product$\rightarrow$Clipart relation, 45/10/10 train/validation/test semantic classes, $64\times64$ images, FiLM conditioning, linear transport and the effective-batch-64 training run. Ten held-out classes are the independent inference units; eight generation repeats and wrong-source assignments are averaged within class before the 10,000-draw class bootstrap.

\begin{table}[h]
\centering\small
\caption{Office-Home real-image evaluator audit.}
\begin{tabular}{lc}
\toprule
Check & Result\\
\midrule
CLIP top-1, Product / Clipart & 89.03\% / 65.96\%\\
Test classes with Clipart top-1 $\geq30\%$ & 10/10\\
DINOv2 held-out domain accuracy & 98.30\%\\
Same-class centroid top-1 / median rank & 100.0\% / 1\\
Same-minus-different centroid cosine gap & .5077\\
\bottomrule
\end{tabular}
\end{table}

\subsection{Validation-only checkpoint and adaptation selection}

All Office-Home model and adapting-baseline choices were fixed on the ten validation classes before the outer test. Validation DINO-SW decreased monotonically across the saved checkpoint sequence, and the 50,000-step checkpoint was selected for the final route comparisons.

\begin{table}[h]
\centering\small
\caption{Office-Home validation-only selection record. The upper block reports the main-route checkpoint trajectory. The lower block reports the budget-specific target-only and full-fine-tuning settings subsequently frozen for final testing. These validation values are not pooled with test results.}
\label{tab:office-selection}
\begin{tabular}{rc}
\toprule
Checkpoint step & Main $\KT=0$ DINO-SW\\
\midrule
20,000 & .040525\\
25,000 & .039609\\
30,000 & .038712\\
35,000 & .038019\\
40,000 & .037572\\
45,000 & .037351\\
50,000 & .037283\\
\bottomrule
\end{tabular}

\vspace{4pt}
\begin{tabular}{crrrrrr}
\toprule
$\KT$ & Target steps & Target lr & Target val SW & Full-FT steps & Full-FT lr & Full-FT val SW\\
\midrule
1 & 5 & .030 & .037660 & 100 & $3{\times}10^{-5}$ & .036521\\
5 & 5 & .010 & .038021 & 400 & $3{\times}10^{-5}$ & .034206\\
10 & 5 & .030 & .037397 & 400 & $3{\times}10^{-5}$ & .032314\\
\bottomrule
\end{tabular}
\end{table}

A separate validation-only zero-target code-path check (a separate sampling run; Table~\ref{tab:office-selection} reports .037283 for the same checkpoint) gave DINO-SW .037239 for correct-source transport, .038168 for wrong-source transport, .037950 for relation-only prediction and .040021 for source reuse. These values are selection/audit evidence only; all Office-Home headline superiority claims use the held-out test route matrices below.

\subsection{Zero-target and target-budget results}
\begin{table}[h]
\centering\small
\caption{Office-Home $\KT=0$ route matrix. Distance metrics are lower-is-better; requested-class and Clipart fractions are higher-is-better.}
\begin{tabular}{lrrrr}
\toprule
Route & DINO-SW & DINO-MMD & Object ID (\%) & Clipart (\%)\\
\midrule
Correct $+$ transport & .038969 & .407846 & 1.67 & 86.50\\
Wrong $+$ transport & .039394 & .419016 & 1.44 & 86.27\\
Relation-only & .039251 & .415280 & 1.37 & 86.71\\
Source reuse & .041641 & .457524 & 1.18 & 63.80\\
\bottomrule
\end{tabular}
\end{table}

The paired SW gain over wrong-source transport is $.000425\sigstar$ $[.000213,.000643]$; over relation-only it is $.000282\sigstar$ $[.000109,.000456]$; over source reuse it is $.002673\sigstar$ $[.002270,.003036]$. Requested-class differences from wrong and relation-only are unresolved. Thus the zero-target claim is distributional/domain alignment rather than reliable semantic transfer.

\begin{table}[htbp]
\centering
\scriptsize
\setlength{\tabcolsep}{3pt}
\caption{Office-Home target-budget route means. The same zero-target correct-source generations are reused across budgets; adapting baselines receive the stated target support. SW and MMD are lower-is-better; identity and Clipart rates are percentages.}\label{tab:office-budget-complete}
\begin{tabular}{clrrrr}
\toprule
$\KT$ & Route & DINO-SW & DINO-MMD & Object ID (\%) & Clipart (\%)\\
\midrule
1 & Correct $+$ transport & 0.038969 & 0.407846 & 1.67 & 86.50\\
1 & Target-only & 0.040346 & 0.439860 & 2.23 & 75.73\\
1 & Full FT & 0.039336 & 0.468223 & 5.33 & 87.49\\
5 & Correct $+$ transport & 0.038969 & 0.407846 & 1.67 & 86.50\\
5 & Target-only & 0.039031 & 0.409734 & 1.81 & 81.22\\
5 & Full FT & 0.034550 & 0.349838 & 12.55 & 83.11\\
10 & Correct $+$ transport & 0.038969 & 0.407846 & 1.67 & 86.50\\
10 & Target-only & 0.039001 & 0.406858 & 1.61 & 82.79\\
10 & Full FT & 0.033859 & 0.321982 & 12.01 & 83.82\\
\bottomrule
\end{tabular}
\end{table}

\begin{figure}[t]
\centering
\includegraphics[width=.98\linewidth]{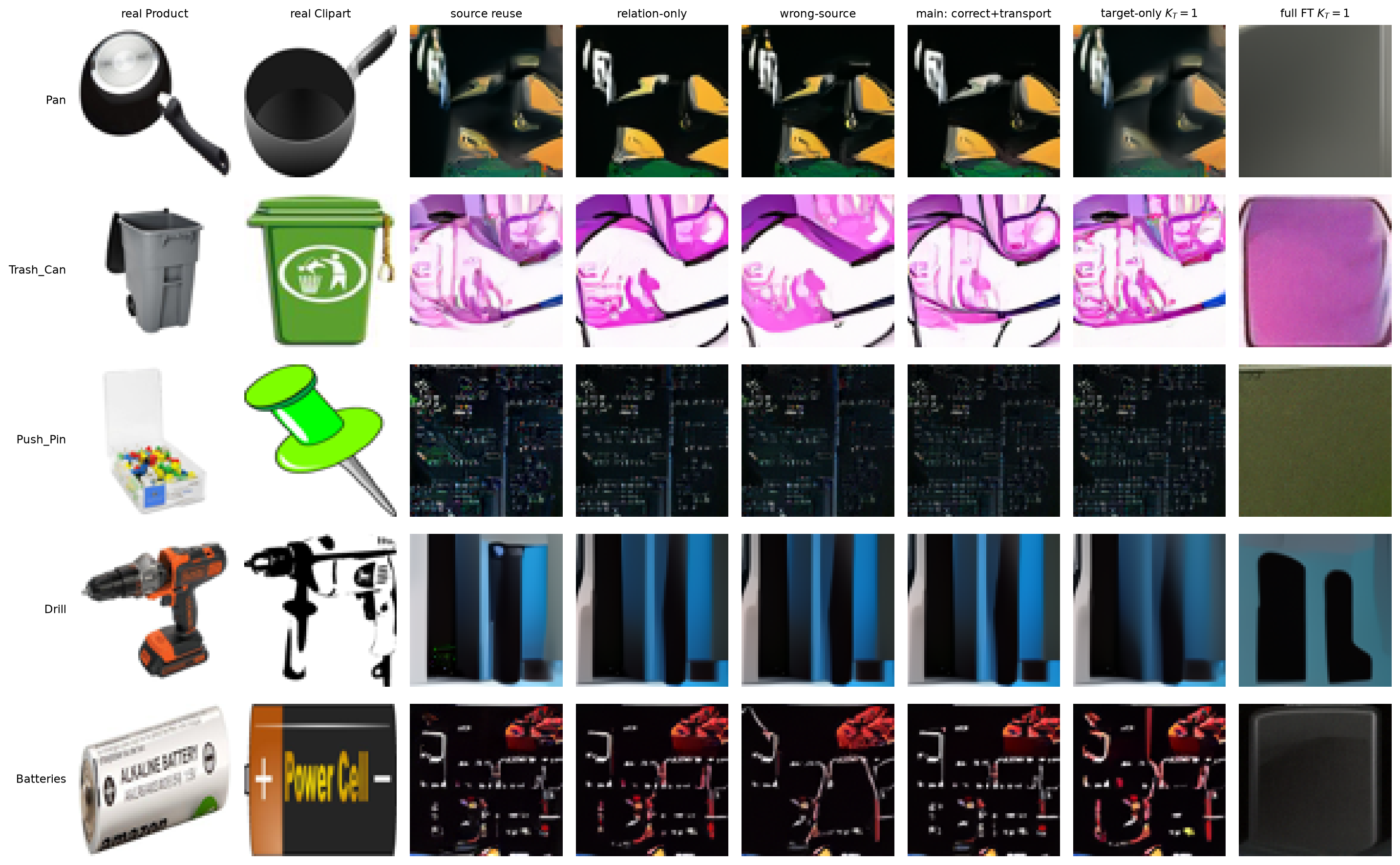}
\caption{Office-Home generation grid. Rows are held-out classes. Columns contain real Product, real Clipart, source reuse, relation-only, wrong-source transport, correct-source transport, one-shot target-only and one-shot full fine-tuning. Generated routes share DDIM noise within class.}
\label{fig:office-grid}
\end{figure}

At $\KT=1$, transport improves DINO-SW over target-only with paired difference $.001377$ $[.000568,.002311]$; its SW comparison with full fine-tuning is unresolved, although its MMD is better. At $\KT=5$ and $10$, full fine-tuning is better in SW, MMD and requested-object accuracy, with paired intervals excluding zero. Figure~\ref{fig:office-grid} shows example generations.

\FloatBarrier
\paragraph{Complete main-relative uncertainty.}
For distance metrics, $\Delta$ is comparator minus main; for requested identity and Clipart share it is main minus comparator. Thus positive values always favour correct-source transport in Tables~\ref{tab:office-distance-paired} and \ref{tab:office-semantic-paired}. Eight generation repeats and wrong-source assignments are averaged within each of the ten held-out classes before a paired 10,000-draw class bootstrap. Identity/domain differences are percentage points. Stars mark unrounded 95\% intervals excluding zero in either direction; the tests are exploratory and unadjusted for multiplicity.
\begin{table}[htbp]
\centering
\scriptsize
\setlength{\tabcolsep}{3pt}
\caption{Office-Home paired distributional gains with 95\% class-bootstrap intervals. Positive favours correct-source transport; negative favours the comparator.}\label{tab:office-distance-paired}
\begin{tabular}{clll}
\toprule
$\KT$ & Comparator & $\Delta$ DINO-SW [95\% CI] & $\Delta$ DINO-MMD [95\% CI]\\
\midrule
0 & Wrong + transport & $0.000425^{*}$ [0.000213, 0.000643] & $0.011170^{*}$ [0.006870, 0.015706]\\
0 & Relation-only & $0.000282^{*}$ [0.000109, 0.000456] & $0.007434^{*}$ [0.002360, 0.012075]\\
0 & Source reuse & $0.002673^{*}$ [0.002270, 0.003036] & $0.049678^{*}$ [0.039741, 0.060430]\\
1 & Target-only & $0.001377^{*}$ [0.000568, 0.002311] & $0.032014^{*}$ [0.015413, 0.047276]\\
1 & Full FT & $0.000367$ [-0.001578, 0.002312] & $0.060376^{*}$ [0.019137, 0.101637]\\
5 & Target-only & $0.000062$ [-0.000754, 0.000800] & $0.001888$ [-0.019453, 0.018779]\\
5 & Full FT & $-0.004419^{*}$ [-0.006279, -0.002922] & $-0.058008^{*}$ [-0.097989, -0.024569]\\
10 & Target-only & $0.000032$ [-0.000685, 0.000719] & $-0.000988$ [-0.021510, 0.016291]\\
10 & Full FT & $-0.005110^{*}$ [-0.006639, -0.003887] & $-0.085864^{*}$ [-0.124581, -0.053092]\\
\bottomrule
\end{tabular}
\end{table}
\begin{table}[htbp]
\centering
\scriptsize
\setlength{\tabcolsep}{3pt}
\caption{Office-Home paired identity and domain-share gains with 95\% class-bootstrap intervals, in percentage points. Positive favours correct-source transport.}\label{tab:office-semantic-paired}
\begin{tabular}{clll}
\toprule
$\KT$ & Comparator & $\Delta$ object ID [95\% CI] & $\Delta$ Clipart [95\% CI]\\
\midrule
0 & Wrong + transport & $0.230$ [-0.043, 0.543] & $0.226$ [-0.438, 0.955]\\
0 & Relation-only & $0.299$ [-0.078, 0.885] & $-0.208$ [-0.885, 0.443]\\
0 & Source reuse & $0.482$ [-0.221, 1.406] & $22.695^{*}$ [20.820, 24.661]\\
1 & Target-only & $-0.56$ [-1.98, 0.42] & $10.768^{*}$ [2.760, 19.062]\\
1 & Full FT & $-3.66$ [-8.57, 0.18] & $-0.990$ [-4.844, 2.734]\\
5 & Target-only & $-0.14$ [-0.70, 0.30] & $5.273^{*}$ [1.667, 9.532]\\
5 & Full FT & $-10.89^{*}$ [-19.82, -4.14] & $3.385$ [-1.289, 8.581]\\
10 & Target-only & $0.05$ [-0.64, 0.57] & $3.711^{*}$ [0.482, 7.161]\\
10 & Full FT & $-10.34^{*}$ [-17.88, -3.84] & $2.682$ [-1.289, 8.815]\\
\bottomrule
\end{tabular}
\end{table}

\subsection{Semantic limitations after bypassing transport}
Three frozen-checkpoint probes and a separately trained class-conditioning capacity control probe where object identity is lost.

\begin{table}[h]
\centering\small
\caption{Office-Home target-coordinate retrieval on ten held-out classes.}
\begin{tabular}{lrrrr}
\toprule
Query route & L2 top-1 & Cosine top-1 & L2 margin & Cosine margin\\
\midrule
Raw $z_S$ & 12.0\% & 26.5\% & -.0500 & -.1749\\
Correct source $+$ transport & 42.0\% & 44.5\% & -.0153 & -.0620\\
Wrong-source transport & 20.0\% & 19.0\% & -.0897 & -.2654\\
Relation-only & 10.0\% & 10.0\% & -.0545 & -.1510\\
\bottomrule
\end{tabular}
\end{table}

The cross-domain retrieval probe uses 20 fixed support resamples over the ten held-out classes, with 16 Product images and at most 32 disjoint Clipart support images per class. Retrieval is among ten target-class coordinates. L2 margins compare the correct target distance with the nearest wrong target; cosine margins compare corresponding similarities, with positive margins favourable in both cases. Table~\ref{tab:office-retrieval-paired} averages support resamples within class and reports paired class-bootstrap contrasts. This target-coordinate task is distinct from the within-Product retrieval summary in the main text.
\begin{table}[htbp]
\centering\scriptsize\setlength{\tabcolsep}{3pt}
\caption{Office-Home target-coordinate retrieval gains, main minus comparator, in percentage points with paired class-bootstrap 95\% intervals.}
\label{tab:office-retrieval-paired}
\begin{tabular}{lrr}
\toprule
Comparator & L2 $\Delta$ [95\% CI] & Cosine $\Delta$ [95\% CI]\\
\midrule
Raw $z_S$ & $+30.00\,[+4.00,+58.50]\sigstar$ & $+18.00\,[-2.50,+43.50]$\\
Wrong-source transport & $+22.00\,[-11.50,+56.00]$ & $+25.50\,[-10.00,+60.00]$\\
Relation only & $+32.00\,[+5.50,+59.50]\sigstar$ & $+34.50\,[+7.00,+61.50]\sigstar$\\
\bottomrule
\end{tabular}
\end{table}

Correct transport improves L2 retrieval over raw $z_S$ and both L2/cosine retrieval over relation-only (intervals exclude zero), but the correct-versus-wrong transport class intervals include zero. This is partial coordinate alignment rather than a resolved source-specific retrieval result.

\begin{table}[h]
\centering\small
\caption{Office-Home direct-$z_T$ generation using abundant non-query Clipart support.}
\begin{tabular}{lrrrrr}
\toprule
Group & Generated ID & Real ID & DINO-SW & DINO-MMD & Clipart\\
\midrule
Seen (45 classes) & 2.38\% & 50.33\% & .03679 & .35175 & 87.34\%\\
Held-out (10 classes) & 1.56\% & 52.00\% & .03902 & .40642 & 88.33\%\\
\bottomrule
\end{tabular}
\end{table}

For direct-target generation, all available Clipart images outside 20 fixed real queries per class form the target support. The frozen EMA checkpoint (hash prefix \texttt{44e2710a}) generates 96 images per class using 50-step DDIM. Seen classes were used in original Meta-Diff training, and their query images were not necessarily excluded from that training. This is a target-coordinate diagnostic, not the zero-target main route. The seen and held-out groups contain different classes, so their identity contrast independently resamples the two groups rather than pairing them.

Bypassing transport does not restore object identity; target-coordinate representation and/or generator use remain possible limitations, and these results do not isolate the decoder alone. The held-out-minus-seen generated-identity contrast is $-.82$ percentage points with 95\% interval $[-2.45,.76]$.

The diffusion-time probe replaces the correct target coordinate with a cyclic wrong-class coordinate while holding query image, noise and timestep fixed. On seen classes, wrong-minus-correct MSE is positive in all five timestep bins; on held-out classes it is positive with intervals excluding zero in three of five bins. The highest-noise differences are only $5.5\times10^{-5}$ (seen) and $1.7\times10^{-5}$ (held-out), showing weak but nonzero coordinate use.

\begin{table}[htbp]
\centering\scriptsize\setlength{\tabcolsep}{3pt}
\caption{Office-Home direct-target coordinate use across diffusion time. MSEs use identical query/noise/timestep pairs; positive wrong-minus-correct differences favour the correct coordinate. Intervals resample classes.}
\label{tab:office-time-probe}
\begin{tabular}{llrrl}
\toprule
Group & Timesteps & Correct MSE & Wrong MSE & $\Delta_z$ [95\% CI]\\
\midrule
Seen & 0--199 & 0.056513 & 0.056724 & $+0.000211\,[+0.000141,+0.000294]\sigstar$\\
 & 200--399 & 0.019655 & 0.019785 & $+0.000131\,[+0.000086,+0.000183]\sigstar$\\
 & 400--599 & 0.013028 & 0.013164 & $+0.000136\,[+0.000089,+0.000189]\sigstar$\\
 & 600--799 & 0.008825 & 0.008963 & $+0.000138\,[+0.000095,+0.000187]\sigstar$\\
 & 800--999 & 0.003287 & 0.003342 & $+0.000055\,[+0.000039,+0.000072]\sigstar$\\
\midrule
Held-out & 0--199 & 0.059910 & 0.059928 & $+0.000018\,[-0.000044,+0.000073]$\\
 & 200--399 & 0.022160 & 0.022180 & $+0.000020\,[-0.000070,+0.000098]$\\
 & 400--599 & 0.014911 & 0.014948 & $+0.000037\,[+0.000006,+0.000068]\sigstar$\\
 & 600--799 & 0.009441 & 0.009505 & $+0.000063\,[+0.000006,+0.000144]\sigstar$\\
 & 800--999 & 0.003140 & 0.003157 & $+0.000017\,[+2.15\!\times\!10^{-7},+0.000034]\sigstar$\\
\bottomrule
\end{tabular}
\end{table}

Finally, an explicit 16-D learned class embedding provides a separately trained capacity control with the same FiLM interface. On seen Clipart classes, generated identity rises from 2.38\% with direct learned $z_T$ to 6.90\%, a paired gain of 4.51 percentage points $[2.85,6.37]$; DINO-SW and MMD also improve. Because 6.90\% remains far below the 50.33\% real-image calibration, the control demonstrates partial semantic conditioning capacity rather than a complete solution.

The capacity control is newly initialised, has 45 learned 16-D training-class embeddings, and uses neither the set encoder nor transport. It completes 50,000 updates with 64 independently noised observations per update; evaluation uses the same 20 fixed real queries and 96 generated images per class with DDIM-50. Those query IDs are explicitly withheld from this new control, whereas they may have appeared during training of the original direct-$z_T$ model. This exposure difference limits causal attribution of the gain to the conditioning representation alone. Tables~\ref{tab:office-capacity-raw} and \ref{tab:office-capacity-paired} report the raw metrics and contrasts; positive effects in the latter favour the explicit class control, not Meta-Diff's main route.
\begin{table}[htbp]
\centering\scriptsize\setlength{\tabcolsep}{3pt}
\caption{Office-Home separately trained class-conditioning capacity control on 45 seen classes. Identity and Clipart columns are percentages.}
\label{tab:office-capacity-raw}
\begin{tabular}{lrrrrr}
\toprule
Seen-class route & Generated identity (\%) & Real identity (\%) & DINO-SW $\downarrow$ & DINO-MMD $\downarrow$ & Clipart (\%)\\
\midrule
Original direct $z_T$ & 2.38 & 50.33 & 0.03679 & 0.35175 & 87.34\\
Explicit class embedding & 6.90 & 50.33 & 0.03507 & 0.32768 & 87.45\\
\bottomrule
\end{tabular}
\end{table}
\begin{table}[htbp]
\centering\scriptsize\setlength{\tabcolsep}{3pt}
\caption{Office-Home capacity-control effects with 95\% intervals paired over 45 classes. Identity/domain gains are control minus direct-$z_T$; SW/MMD reductions are direct-$z_T$ minus control.}
\label{tab:office-capacity-paired}
\begin{tabular}{lr}
\toprule
Metric & Favor-control $\Delta$ [95\% CI]\\
\midrule
Requested-class identity & $+4.51\,\mathrm{pp}\,[+2.85,+6.37]\sigstar$\\
DINO-SW reduction & $+0.00172\,[+0.00121,+0.00229]\sigstar$\\
DINO-MMD reduction & $+0.02407\,[+0.01332,+0.03671]\sigstar$\\
Clipart-share increase & $+0.12\,\mathrm{pp}\,[-1.30,+1.53]$\\
\bottomrule
\end{tabular}
\end{table}

\FloatBarrier
\section{Clinical and External-Domain Stress Tests}
\label{app:clinical}
These studies are retained for transparency and boundary analysis. They are not used as positive evidence for clinical or demographic source-specific transfer.

\subsection{ChestX-ray14 age-group transfer}
Tasks are single-positive thoracic findings, source ages are 45--60, and target ages are 0--30, 30--45 and 60--75. The fixed 4/2/3 finding split uses whole-patient allocation. Evaluation covers three held-out findings, three age relations and four resamples, with $\MS=64$.

\begin{table}[h]
\centering\scriptsize
\caption{ChestX-ray14 SW route means. Lower is better. The independently reported wrong-minus-correct paired interval at $\KT=1$ is $7.7\times10^{-6}\pm2.3\times10^{-5}$; other displayed route differences lack paired covariance information.}
\resizebox{\linewidth}{!}{%
\begin{tabular}{lccccc}
\toprule
Method & $\KT=1$ & $2$ & $5$ & $10$ & $20$\\
\midrule
No adaptation & .3696 & .3696 & .3696 & .3696 & .3696\\
Target-only & .1397 & .1406 & .1398 & .1405 & .1408\\
Relation-only & .3680 & .3680 & .3680 & .3680 & .3680\\
Wrong-source transport & .1407 & .1407 & .1407 & .1407 & .1407\\
Correct source $+$ transport & .1407 & .1407 & .1407 & .1407 & .1407\\
Full fine-tuning & .2981 & .2802 & .2304 & .1800 & .1591\\
\bottomrule
\end{tabular}}
\end{table}

Correct-source transport reaches the broad target-image distribution far better than no adaptation or relation-only, but wrong-source transport has the same rounded mean and the available paired correct-versus-wrong contrast is unresolved. The finding classifier has only 12.1\% accuracy and an abundant-target age diagnostic has slope $.039\pm.064$, so neither image distance nor predictor output validates the requested age/finding identity. This branch therefore establishes broad distribution matching, not source-specific demographic transfer.

\subsection{Fitzpatrick17k disease and pathology-family transfer}
All 16,577 official rows were recovered from the SFU renamed archive, matched to the official CSV by MD5, and filtered through the published cleaned allow-list. The retained set contains 11,394 images, of which 11,098 have Fitzpatrick type I--VI. The disease-level split is 8/4/4 train/validation/test diseases. Pathology-family follow-ups use six test families for I--III$\rightarrow$IV--VI and five for I--III$\rightarrow$V--VI.

\begin{table}[h]
\centering\scriptsize
\caption{Fitzpatrick zero-target route means. Invalidated phenotype/semantic instruments are omitted here; statistic distance, SW and energy MMD are lower-is-better.}
\resizebox{\linewidth}{!}{%
\begin{tabular}{llrrr}
\toprule
Protocol & Route & Statistic & SW & MMD\\
\midrule
Disease IV--VI & Source reuse & 2.5248 & .3642 & 7.1765\\
& Wrong disease, same family & 2.5020 & .3638 & 7.1466\\
& Wrong disease, different family & 2.5025 & .3639 & 7.1444\\
& Correct source $+$ transport & 2.5030 & .3638 & 7.1429\\
\midrule
Family IV--VI & Source reuse & 1.7452 & .3081 & 5.7946\\
& Wrong-family transport & 1.7185 & .3064 & 5.7468\\
& Correct source $+$ transport & 1.7188 & .3063 & 5.7436\\
\bottomrule
\end{tabular}}
\end{table}

Disease-level correct-source and wrong-source routes show no resolved difference on the valid distributional metrics under the reported intervals; this is not an equivalence result. The same conclusion holds at pathology-family resolution: correct source has no resolved advantage over wrong-family transport. The stronger I--III$\rightarrow$V--VI follow-up also fails to recover source specificity; the wrong-task-minus-correct paired effects are $-.00068\pm.00273$ for statistic distance, $-.00003\pm.00134$ for SW and $-.00107\pm.03907$ for MMD.

The interpretation is further constrained by the evaluators. The pathology-family semantic gate fails on every outer fold in the stronger split, and an exploratory phenotype audit finds real-target recall of 0\% for IV--VI and 0--6.3\% for V--VI despite high aggregate accuracy. A source-only disease-separability analysis reports $\rho=-.775$ over four diseases, but three of the four semantic effects are exactly zero and only 6,793/10,000 bootstrap draws have defined rank variance. We therefore do not use task separability as the central explanation for the Fitzpatrick result. Downstream demographic augmentation was not run because the pre-specified positive family-semantic gate was not met.

\subsection{CRC histology preflight}
CRC uses NCT-CRC-HE-100K as source and CRC-VAL-HE-7K as external target over nine tissue classes and three outer folds. A prespecified gate required both a valid tissue evaluator and sufficient source--target domain separation before any Meta-Diff training.

\begin{table}[h]
\centering\small
\caption{CRC real-image preflight and stopping decision.}
\begin{tabular}{lc}
\toprule
Check & Result\\
\midrule
Target tissue balanced accuracy & 88.25\%\\
Minimum target tissue recall & 60.57\%\\
Domain balanced accuracy, fold 1 & 59.01\%\\
Domain balanced accuracy, fold 2 & 58.24\%\\
Domain balanced accuracy, fold 3 & 57.00\%\\
Same-tissue centroid retrieval & 100\% top-1 in each fold\\
Diffusion training / generation & Not run: structural gate failed\\
\bottomrule
\end{tabular}
\end{table}

The tissue evaluator and cross-domain tissue alignment pass, but all three domain-separation folds fall below the locked 70\% threshold. No CRC generation result exists. The stop therefore concerns benchmark suitability under the frozen representation and provides no evidence for or against Meta-Diff generation performance.

\section{Supplementary Controlled Experiments}
\label{app:new-experiments}
These post-hoc additions address a known-corruption image-space baseline, a controlled full-FT budget across target-set sizes, and semantic fidelity under a common evaluator. The CIFAR experiments use the same frozen 50,000-update EMA model as the main generation comparison and its fixed semantic-class split. All results are conditional on one trained model; training-seed variability is not estimated. Each test budget contains 20 held-out classes with three severity-3 corruptions per class; each route produces 96 samples and is evaluated against 128 target queries. Source support contains 64 images. SW and energy MMD use the definitions in Appendix~\ref{app:cifar}. Intervals resample 20 classes 10,000 times with seed 20261020, retaining all three corruptions within each class; they are pointwise and unadjusted.

Routes share target queries and DDIM initial noise within an episode. In these new comparisons, denoising noise and full-FT dropout use fixed per-episode random seeds. The full-FT measurements below are a separate seeded replication of the budget-specific comparison in Table~\ref{tab:cifar-main}B; stochastic adaptation means they need not reproduce its earlier point estimates exactly.

\subsection{Known-corruption image-space baseline}
The image-space baseline encodes the correct clean-source support, generates using that untransported coordinate, and then applies the known target corruption at severity 3. It uses neither target images nor target-time optimisation. Blur, noise and contrast have the same definitions and scales as the target-data corruptions; Gaussian corruption noise is independent of the real-query noise. We evaluate the $K_T=1$ slice of the regenerated test episodes; $K_T$ is only an episode index for these two zero-target routes. The baseline and transport share source support, target queries, sampling noise, preprocessing and evaluator.

\begin{table}[ht]
\centering\small
\caption{Known-corruption baseline on 60 matched test episodes. Entries are mean [95\% class-cluster CI]. For SW/MMD, the paired effect is image-space baseline minus transport; for identity, it is transport minus image-space baseline in percentage points. Positive favours transport; stars mark paired intervals excluding zero.}
\label{tab:new-image-baseline}
\begin{tabular}{lccc}
\toprule
Route & SW & Energy MMD & DINO-100 (\%)\\
\midrule
Correct source $+$ transport & \shortstack{$0.1781$\\$[0.1627,0.1959]$} & \shortstack{$2.3306$\\$[1.9624,2.7620]$} & \shortstack{$2.53$\\$[0.54,5.73]$}\\
\shortstack[l]{Clean-source generation\\$+$ corruption} & \shortstack{$0.1688$\\$[0.1528,0.1867]$} & \shortstack{$2.3336$\\$[1.9307,2.8087]$} & \shortstack{$2.62$\\$[0.57,5.71]$}\\
Paired effect & \shortstack{$-0.0092\sigstar$\\$[-0.0138,-0.0045]$} & \shortstack{$0.0030$\\$[-0.0748,0.0827]$} & \shortstack{$-0.09$\\$[-0.59,0.35]$}\\
\bottomrule
\end{tabular}
\end{table}

The simple image-space route has lower SW, so these results do not establish an advantage for task-space transport over this baseline.

\subsection{Fixed full-FT update budget across target-set sizes}
The validation subset contains four fixed validation classes (fine IDs 3, 6, 7 and 9), with all three corruptions and $K_T=1,5,20$. The grid is $N_{\rm FT}\in\{25,100\}$ and weight learning rate $\{10^{-5},10^{-4}\}$. A single setting minimises equally weighted pixel SW over all 36 validation episodes, and is then frozen for all three test budgets. The chosen setting is $N_{\rm FT}=100$, learning rate $10^{-5}$. Each update processes exactly 20 noised examples, repeated from the available support set as necessary; architecture and precision are held fixed. This matches update count and processed-example count within full-FT across $K_T$, not wall-clock time or compute between full-FT and coordinate-only routes.

Test episodes have nested target supports, identical source and query images, and identical DDIM initial noise across $K_T$. This panel uses a separate matched draw from the original budget-specific comparison; route comparisons are made within each panel. Zero-target transport therefore gives the same output at all three budgets in this panel. The target-only reference uses 25 coordinate updates, learning rate $0.01$, prior scale $\beta_0=1$ and 20 examples per update in this new panel. Its coordinate penalty follows the original $\beta_0/K_T$ rule; the controlled full-FT schedule does not use this coordinate prior.

\begin{table}[ht]
\centering\small
\caption{Validation selection for the common full-FT schedule. Lower mean SW is better; no test outcome selects the setting.}
\label{tab:new-budget-selection}
\begin{tabular}{ccc}
\toprule
Updates & Learning rate & Mean validation SW\\
\midrule
25 & $10^{-5}$ & 0.218506\\
25 & $10^{-4}$ & 0.259794\\
100 & $10^{-5}$ & 0.217296\\
100 & $10^{-4}$ & 0.310748\\
\bottomrule
\end{tabular}
\end{table}

\begin{table}[ht]
\centering\small
\caption{Controlled-budget test panel, with 60 matched episodes per budget. Mean [95\% class-cluster CI]. The full-FT update schedule is constant across columns; zero-target transport has identical samples across columns.}
\label{tab:new-fixed-budget}
\begin{tabular}{lccc}
\toprule
Route and metric & $K_T=1$ & $K_T=5$ & $K_T=20$\\
\midrule
Transport: SW & \shortstack{$0.1828$\\$[0.1634,0.2036]$} & \shortstack{$0.1828$\\$[0.1634,0.2036]$} & \shortstack{$0.1828$\\$[0.1634,0.2036]$}\\
Transport: MMD & \shortstack{$2.5036$\\$[2.0193,3.0320]$} & \shortstack{$2.5036$\\$[2.0193,3.0320]$} & \shortstack{$2.5036$\\$[2.0193,3.0320]$}\\
Transport: DINO-100 (\%) & \shortstack{$2.64$\\$[0.40,5.97]$} & \shortstack{$2.64$\\$[0.40,5.97]$} & \shortstack{$2.64$\\$[0.40,5.97]$}\\
Target-only: SW & \shortstack{$0.2108$\\$[0.1857,0.2381]$} & \shortstack{$0.1905$\\$[0.1690,0.2138]$} & \shortstack{$0.1774$\\$[0.1615,0.1937]$}\\
Target-only: MMD & \shortstack{$3.5534$\\$[2.7894,4.4358]$} & \shortstack{$2.7992$\\$[2.2504,3.3998]$} & \shortstack{$2.4628$\\$[2.0898,2.8443]$}\\
Target-only: DINO-100 (\%) & \shortstack{$2.64$\\$[0.50,5.99]$} & \shortstack{$2.74$\\$[0.47,6.18]$} & \shortstack{$2.69$\\$[0.52,5.95]$}\\
Full FT: SW & \shortstack{$0.3054$\\$[0.2749,0.3378]$} & \shortstack{$0.2123$\\$[0.1902,0.2359]$} & \shortstack{$0.1709$\\$[0.1533,0.1919]$}\\
Full FT: MMD & \shortstack{$7.4178$\\$[6.0127,8.9525]$} & \shortstack{$3.5963$\\$[2.9610,4.2934]$} & \shortstack{$2.4053$\\$[1.9574,2.9350]$}\\
Full FT: DINO-100 (\%) & \shortstack{$5.28$\\$[1.23,11.34]$} & \shortstack{$5.26$\\$[1.82,9.83]$} & \shortstack{$5.73$\\$[2.33,10.14]$}\\
\bottomrule
\end{tabular}
\end{table}

\begin{table}[ht]
\centering\small
\caption{Paired SW effects for the controlled-budget panel. The first two rows compare routes at each budget; positive favours transport. The last two rows report changes in full-FT SW as $K_T$ increases; negative favours the larger support set.}
\label{tab:new-fixed-effects}
\begin{tabular}{lccc}
\toprule
Comparison & $K_T=1$ / change & $K_T=5$ & $K_T=20$\\
\midrule
Target-only $-$ transport: SW & \shortstack{$0.0280\sigstar$\\$[0.0127,0.0448]$} & \shortstack{$0.0077\sigstar$\\$[0.0021,0.0134]$} & \shortstack{$-0.0054$\\$[-0.0121,0.0014]$}\\
Full FT $-$ transport: SW & \shortstack{$0.1226\sigstar$\\$[0.0899,0.1568]$} & \shortstack{$0.0294\sigstar$\\$[0.0033,0.0570]$} & \shortstack{$-0.0119$\\$[-0.0429,0.0213]$}\\
Full FT: $K_T=5$ minus 1 & \shortstack{$-0.0932\sigstar$\\$[-0.1270,-0.0564]$} & -- & --\\
Full FT: $K_T=20$ minus 1 & \shortstack{$-0.1345\sigstar$\\$[-0.1730,-0.0957]$} & -- & --\\
\bottomrule
\end{tabular}
\end{table}

\begin{figure}[ht]
\centering
\includegraphics[width=.70\linewidth]{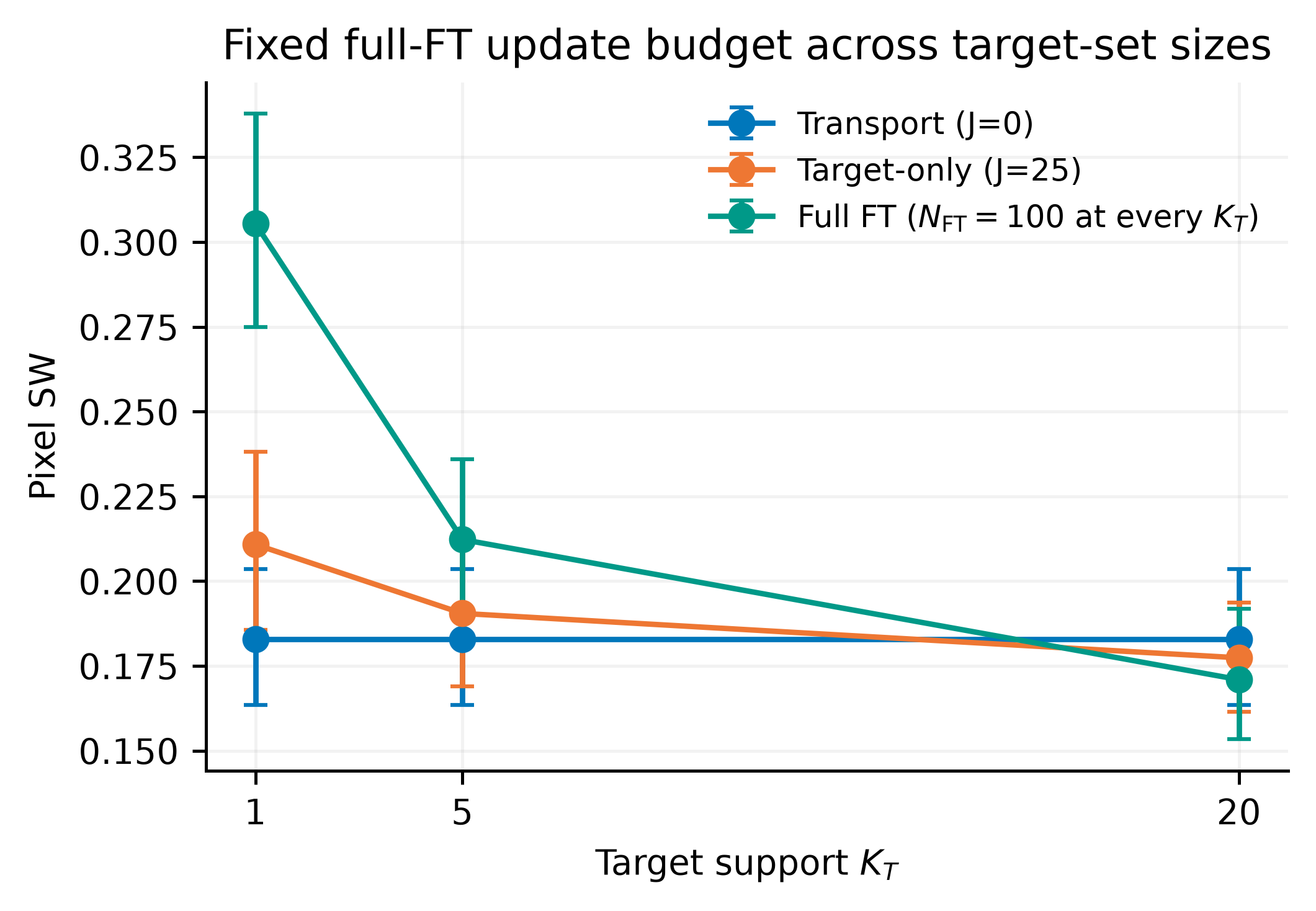}
\caption{Controlled full-FT update schedule across target budgets. Error bars are marginal 95\% class-cluster intervals; route and support-size comparisons use the paired intervals in Table~\ref{tab:new-fixed-effects}.}
\label{fig:new-fixed-budget}
\end{figure}

Under this fixed schedule, the $K_T=20$ versus 1 SW reduction is resolved. Transport has lower SW at $K_T=1,5$; at $K_T=20$, the full-FT-minus-transport effect is $-0.0119\,[-0.0429,0.0213]$, which remains unresolved (Figure~\ref{fig:new-fixed-budget}). These intervals do not cover training-seed uncertainty.

\FloatBarrier

\subsection{Matched CIFAR semantic evaluation}
All routes use a frozen DINOv2-small feature extractor \citep{oquab2023dinov2} with normalised CLS features and cosine nearest-prototype classification over 100 fine classes. Each prototype averages 20 fixed clean source-support images of its class and is normalised again; no target-query image enters the prototypes. Generated and real images undergo the same conversion from $[-1,1]$ to uint8 and the standard DINOv2 image processing. The resulting DINO-100 requested-class fraction is distinct from both the headline class consistency and the independent 80-way direct-$z_T$ diagnostic. It uses labelled source prototypes, including prototypes for the held-out semantic classes.

The three routes are evaluated on the same 60 held-out class--corruption episodes per target budget, with identical query images and generation noise. The original budget-specific full-FT schedule is used here, separately from the controlled-budget panel above. Real-query calibration is scored on exactly the same 128 images used to evaluate each route.

\begin{table}[ht]
\centering\small
\caption{Matched DINO-100 requested-class fractions (\%) and real-query calibration, with 95\% class-cluster intervals. Paired effects are percentage points; positive favours transport.}
\label{tab:new-cifar-semantics}
\begin{tabular}{lccc}
\toprule
Route / paired effect & $K_T=1$ & $K_T=5$ & $K_T=20$\\
\midrule
Transport & \shortstack{$2.53$\\$[0.54,5.73]$} & \shortstack{$2.85$\\$[0.54,6.46]$} & \shortstack{$2.34$\\$[0.35,5.49]$}\\
Target-only & \shortstack{$2.38$\\$[0.36,5.62]$} & \shortstack{$2.71$\\$[0.45,6.25]$} & \shortstack{$2.50$\\$[0.49,5.54]$}\\
Full FT & \shortstack{$3.68$\\$[1.02,7.20]$} & \shortstack{$9.58$\\$[3.85,16.82]$} & \shortstack{$10.50$\\$[5.35,16.70]$}\\
Real-query calibration & \shortstack{$50.49$\\$[42.20,58.79]$} & \shortstack{$49.79$\\$[41.59,57.99]$} & \shortstack{$50.65$\\$[42.23,59.13]$}\\
Transport $-$ target-only & \shortstack{$0.16$\\$[-0.14,0.49]$} & \shortstack{$0.14$\\$[-0.07,0.33]$} & \shortstack{$-0.16$\\$[-0.45,0.09]$}\\
Transport $-$ full FT & \shortstack{$-1.15\sigstar$\\$[-2.67,-0.09]$} & \shortstack{$-6.74\sigstar$\\$[-11.79,-3.00]$} & \shortstack{$-8.16\sigstar$\\$[-12.00,-4.67]$}\\
\bottomrule
\end{tabular}
\end{table}

Full FT has higher DINO-100 requested-class fraction than transport at all three target budgets, whereas transport-versus-target-only differences remain unresolved. All generated-route means remain substantially below their matched real-query calibration.

\begin{table}[ht]
\centering\small
\caption{SW and energy MMD for the new seeded full-FT replication used for semantic evaluation. Table~\ref{tab:cifar-main}B retains the earlier budget-specific comparison.}
\label{tab:new-replication-audit}
\begin{tabular}{lccc}
\toprule
Metric / paired effect & $K_T=1$ & $K_T=5$ & $K_T=20$\\
\midrule
Full FT: SW & \shortstack{$0.2649$\\$[0.2382,0.2927]$} & \shortstack{$0.2177$\\$[0.2002,0.2355]$} & \shortstack{$0.1487$\\$[0.1395,0.1591]$}\\
Full FT $-$ transport: SW & \shortstack{$0.0868\sigstar$\\$[0.0625,0.1107]$} & \shortstack{$0.0343\sigstar$\\$[0.0122,0.0577]$} & \shortstack{$-0.0357\sigstar$\\$[-0.0586,-0.0146]$}\\
Full FT: energy MMD & \shortstack{$5.9003$\\$[4.7552,7.1604]$} & \shortstack{$3.9034$\\$[3.2829,4.5988]$} & \shortstack{$1.9151$\\$[1.7299,2.1178]$}\\
Full FT $-$ transport: MMD & \shortstack{$3.5698\sigstar$\\$[2.5478,4.6246]$} & \shortstack{$1.5648\sigstar$\\$[0.8988,2.2642]$} & \shortstack{$-0.4890$\\$[-1.0680,0.0466]$}\\
\bottomrule
\end{tabular}
\end{table}

\FloatBarrier

\subsection{Matched Office-Home semantic evaluation}
For Office-Home, we analyse matched semantic scores from the effective-batch-64, 50,000-update EMA model: 10 test classes, eight repeats, three target budgets ($K_T=1,5,10$), and three routes. Within each class/repeat/budget, target queries and generation seeds match across routes; support and queries are disjoint. Each repeat contains 96 generated images.

The same 20 fixed real queries per class are additionally evaluated with the identical frozen CLIP ViT-B/32 classifier \citep{radford2021clip}: 65 candidate classes, a three-prompt ensemble, and the same 64-pixel image preparation followed by the CLIP image processor. The existing generated-image scores are compared with this new same-query real-image calibration. This 200-image, class-balanced calibration is separate from the 564-image preflight audit reporting 65.96\%, which applies the CLIP processor to the original-resolution images. Both the evaluation images and their preprocessing differ. Repeats are averaged within class before 10,000 class-bootstrap resamples.

\begin{table}[ht]
\centering\small
\caption{Matched Office-Home CLIP-65 identity (\%) and same-query real calibration. Effects are percentage points with 95\% class intervals; positive favours transport. The identical real-query set is used at every target budget.}
\label{tab:new-office-semantics}
\begin{tabular}{lccc}
\toprule
Route / paired effect & $K_T=1$ & $K_T=5$ & $K_T=10$\\
\midrule
Transport & \shortstack{$1.67$\\$[0.60,3.19]$} & \shortstack{$1.67$\\$[0.60,3.19]$} & \shortstack{$1.67$\\$[0.60,3.19]$}\\
Target-only & \shortstack{$2.23$\\$[0.52,5.01]$} & \shortstack{$1.81$\\$[0.65,3.35]$} & \shortstack{$1.61$\\$[0.59,2.94]$}\\
Full FT & \shortstack{$5.33$\\$[0.63,11.59]$} & \shortstack{$12.55$\\$[5.13,21.63]$} & \shortstack{$12.01$\\$[4.84,20.16]$}\\
Real-query calibration & \shortstack{$52.00$\\$[40.00,66.00]$} & \shortstack{$52.00$\\$[40.00,66.00]$} & \shortstack{$52.00$\\$[40.00,66.00]$}\\
Transport $-$ target-only & \shortstack{$-0.56$\\$[-1.98,0.42]$} & \shortstack{$-0.14$\\$[-0.70,0.30]$} & \shortstack{$0.05$\\$[-0.64,0.57]$}\\
Transport $-$ full FT & \shortstack{$-3.66$\\$[-8.57,0.18]$} & \shortstack{$-10.89\sigstar$\\$[-19.82,-4.14]$} & \shortstack{$-10.34\sigstar$\\$[-17.88,-3.84]$}\\
\bottomrule
\end{tabular}
\end{table}

The transport-versus-target-only semantic intervals contain zero at all three budgets. Full FT improves identity relative to transport at $K_T=5,10$, while the $K_T=1$ difference remains unresolved. Even the higher full-FT means remain well below the real-query calibration. These matched measurements reinforce the need to distinguish distributional distances from reliable requested-class generation; they do not isolate the cause of the semantic limitation.
\FloatBarrier

\section{Implementation and Reproducibility}
The CIFAR score model uses a 128-base-channel U-Net and $k=16$ coordinates; its encoder has approximately 0.33M parameters. The basis contributes 55,344 conditioning parameters. The actual three-relation linear transport is $Az_S+b_c$: $16^2+3\times16=304$ active parameters. Its saved module has 352 registered parameters because a 48-value compatibility embedding is not used by its forward map. A single-relation affine map has $16^2+16=272$ parameters. The headline CIFAR correct source $+$ transport route directly decodes the predicted coordinate at all $K_T$, without target images, coordinate refinement or backbone update. Office-Home retains its separate 64-base-channel, $64\times64$ FiLM model, DDIM-50 and no-refinement headline protocol.

Semantic splits are fixed and target support/query images are disjoint. Within each paired CIFAR generation or ablation package, routes share the task, support/query episode and generation noise; the class-cluster bootstrap retains each class's three corruption episodes and resamples classes rather than treating those episodes as independent. The independent frozen-coordinate retrieval diagnostic instead bootstraps fixed support resamples conditional on the class set, and its direct-target generation and timestep diagnostics resample classes. The direct-target diagnostic uses the explicitly stated post-hoc support/query repartition. Independent conditioning packages are not pooled. Validation-selected full-FT budgets are frozen before matched final-test evaluation, and no test score selects a new model, budget or subset. Other benchmarks retain their own frozen protocols and statistical units.

The paper distinguishes final-model interventions from historical evidence. The GMM calibration table and the GMM interpolation panel use separate 80k-step GMM checkpoints, while the CIFAR interpolation panel and the main-text own/mean/shuffled/zero generation grid use the headline 50k-step CIFAR checkpoint; these diagnose coordinates, not deployment superiority. Historical GMM capacity/relatedness aggregates remain accurately labelled in the appendix, and transport-structure ablations compare MLP and linear maps. The refinement table reports matched on/off measurements on the correct source $+$ transport checkpoint. Frozen-coordinate and direct-target semantic probes are reported as independent diagnostics.

\subsection{Evaluation protocols}
\label{app:protocol-index}
Table~\ref{tab:protocol-index} summarises the sampling and evaluation settings. Independent comparisons retain their own models, task splits, evaluators and statistical units.

\begin{table}[h]
\centering\footnotesize
\caption{Protocol matrix. Counts refer to actual metric evaluation unless marked as pool capacities. Independent comparisons retain their own statistical units. The new supplementary experiments are described in Appendix~\ref{app:new-experiments}.}
\label{tab:protocol-index}
\setlength{\tabcolsep}{4pt}
\begin{tabular}{p{.20\linewidth}p{.31\linewidth}p{.39\linewidth}}
\toprule
Comparison & Data and sampling & Evaluation and uncertainty\\
\midrule
GMM factorial & 42 $(\alpha,\eta)$ cells; five seeds; 64 held-out tasks per seed & SW/MMD paired by seed and task; hierarchical intervals; abundant-target encoding reference uses 512 target points.\\[3pt]
CIFAR main zero-target & 20 validation classes $\times$ three corruptions; 64 source images; 96 generated/128 query images & Pixel SW: 256 projections; energy MMD: 64/64 samples. Frozen semantic instrument from the original package. Class-cluster intervals.\\[3pt]
CIFAR original adaptation comparison & 20 test classes $\times$ three corruptions at each $K_T=1,5,20$; 96 generated/128 query images & Pixel SW and energy MMD as above. Budget-specific full-FT settings; class-cluster intervals.\\[3pt]
CIFAR conditioning, transport and source-budget ablations & Separately matched validation comparisons; nested source prefixes for $M_S=16,\ldots,300$ & Each comparison uses matched class--corruption episodes; the FiLM conditioning and FiLM-linear arms share one trained model. Paired class-cluster intervals; one training seed per arm.\\[3pt]
CIFAR independent direct-$z_T$ diagnostic & Its separate 200-support/50-query repartition & Its stated DINO identity instrument and real-query calibration; class intervals. The coordinate-retrieval probe instead resamples supports conditional on its class set.\\[3pt]
Office-Home main routes & 10 held-out classes; eight repeats; 96 generated images per repeat; 20 fixed real queries per class & DINO feature-space distances and 65-way CLIP identity are distinct metrics. Repeats are averaged within class; class intervals.\\[3pt]
New CIFAR supplementary routes & Same frozen model; matched source/query images and initial generation noise; nested supports in the controlled-budget panel & Pixel SW/MMD and a separate DINO-100 requested-class fraction; matched real-query calibration. Class-cluster intervals, conditional on one training seed.\\
\bottomrule
\end{tabular}
\end{table}

For the main CIFAR split, the pooled official train/test images are partitioned by semantic superclass, and the per-class source-support/source-query/target-support-reserve/target-query pool capacities are 300/50/20/230. The model uses a cosine 1,000-step diffusion schedule, simple denoising loss, loss weights $(\lambda_T,\lambda_{\rm tr},\lambda_z)=(1,1,10^{-4})$, training learning rate $2\times10^{-4}$, zero weight decay and EMA decay $0.999$. Evaluation uses the 50,000-update EMA model and deterministic DDIM-50 sampling.

\end{document}

%% file: plots/gmm_full_grid_diagnostics.tex
\begin{tikzpicture}[x=1cm,y=1cm,every node/.style={font=\scriptsize}]
\begin{scope}[shift={(0,0)}]
\fill[red!12] (0.00,0.00) rectangle ++(.7,.52);
\node[text=black] at (0.35,0.26) {-0.00};
\fill[red!12] (0.70,0.00) rectangle ++(.7,.52);
\node[text=black] at (1.05,0.26) {-0.00};
\fill[blue!12] (1.40,0.00) rectangle ++(.7,.52);
\node[text=black] at (1.75,0.26) {0.00};
\fill[blue!12] (2.10,0.00) rectangle ++(.7,.52);
\node[text=black] at (2.45,0.26) {0.00};
\fill[red!12] (2.80,0.00) rectangle ++(.7,.52);
\node[text=black] at (3.15,0.26) {-0.00};
\fill[red!12] (3.50,0.00) rectangle ++(.7,.52);
\node[text=black] at (3.85,0.26) {-0.00};
\fill[blue!12] (0.00,0.52) rectangle ++(.7,.52);
\node[text=black] at (0.35,0.78) {0.00*};
\fill[blue!12] (0.70,0.52) rectangle ++(.7,.52);
\node[text=black] at (1.05,0.78) {0.00*};
\fill[blue!12] (1.40,0.52) rectangle ++(.7,.52);
\node[text=black] at (1.75,0.78) {0.00*};
\fill[blue!12] (2.10,0.52) rectangle ++(.7,.52);
\node[text=black] at (2.45,0.78) {0.00*};
\fill[blue!12] (2.80,0.52) rectangle ++(.7,.52);
\node[text=black] at (3.15,0.78) {0.00};
\fill[blue!12] (3.50,0.52) rectangle ++(.7,.52);
\node[text=black] at (3.85,0.78) {0.00};
\fill[blue!20] (0.00,1.04) rectangle ++(.7,.52);
\node[text=black] at (0.35,1.30) {0.05*};
\fill[blue!20] (0.70,1.04) rectangle ++(.7,.52);
\node[text=black] at (1.05,1.30) {0.06*};
\fill[blue!20] (1.40,1.04) rectangle ++(.7,.52);
\node[text=black] at (1.75,1.30) {0.05*};
\fill[blue!17] (2.10,1.04) rectangle ++(.7,.52);
\node[text=black] at (2.45,1.30) {0.04*};
\fill[blue!15] (2.80,1.04) rectangle ++(.7,.52);
\node[text=black] at (3.15,1.30) {0.02*};
\fill[blue!13] (3.50,1.04) rectangle ++(.7,.52);
\node[text=black] at (3.85,1.30) {0.01};
\fill[blue!32] (0.00,1.56) rectangle ++(.7,.52);
\node[text=black] at (0.35,1.82) {0.14*};
\fill[blue!31] (0.70,1.56) rectangle ++(.7,.52);
\node[text=black] at (1.05,1.82) {0.13*};
\fill[blue!29] (1.40,1.56) rectangle ++(.7,.52);
\node[text=black] at (1.75,1.82) {0.11*};
\fill[blue!28] (2.10,1.56) rectangle ++(.7,.52);
\node[text=black] at (2.45,1.82) {0.11*};
\fill[blue!23] (2.80,1.56) rectangle ++(.7,.52);
\node[text=black] at (3.15,1.82) {0.07*};
\fill[blue!19] (3.50,1.56) rectangle ++(.7,.52);
\node[text=black] at (3.85,1.82) {0.05*};
\fill[blue!46] (0.00,2.08) rectangle ++(.7,.52);
\node[text=black] at (0.35,2.34) {0.22*};
\fill[blue!47] (0.70,2.08) rectangle ++(.7,.52);
\node[text=black] at (1.05,2.34) {0.23*};
\fill[blue!45] (1.40,2.08) rectangle ++(.7,.52);
\node[text=black] at (1.75,2.34) {0.22*};
\fill[blue!38] (2.10,2.08) rectangle ++(.7,.52);
\node[text=black] at (2.45,2.34) {0.18*};
\fill[blue!34] (2.80,2.08) rectangle ++(.7,.52);
\node[text=black] at (3.15,2.34) {0.15*};
\fill[blue!25] (3.50,2.08) rectangle ++(.7,.52);
\node[text=black] at (3.85,2.34) {0.09*};
\fill[blue!58] (0.00,2.60) rectangle ++(.7,.52);
\node[text=white] at (0.35,2.86) {0.31*};
\fill[blue!62] (0.70,2.60) rectangle ++(.7,.52);
\node[text=white] at (1.05,2.86) {0.33*};
\fill[blue!59] (1.40,2.60) rectangle ++(.7,.52);
\node[text=white] at (1.75,2.86) {0.31*};
\fill[blue!52] (2.10,2.60) rectangle ++(.7,.52);
\node[text=white] at (2.45,2.86) {0.26*};
\fill[blue!44] (2.80,2.60) rectangle ++(.7,.52);
\node[text=black] at (3.15,2.86) {0.21*};
\fill[blue!30] (3.50,2.60) rectangle ++(.7,.52);
\node[text=black] at (3.85,2.86) {0.12*};
\fill[blue!75] (0.00,3.12) rectangle ++(.7,.52);
\node[text=white] at (0.35,3.38) {0.42*};
\fill[blue!76] (0.70,3.12) rectangle ++(.7,.52);
\node[text=white] at (1.05,3.38) {0.42*};
\fill[blue!77] (1.40,3.12) rectangle ++(.7,.52);
\node[text=white] at (1.75,3.38) {0.43*};
\fill[blue!71] (2.10,3.12) rectangle ++(.7,.52);
\node[text=white] at (2.45,3.38) {0.39*};
\fill[blue!66] (2.80,3.12) rectangle ++(.7,.52);
\node[text=white] at (3.15,3.38) {0.36*};
\fill[blue!46] (3.50,3.12) rectangle ++(.7,.52);
\node[text=black] at (3.85,3.38) {0.22*};
\node[left] at (0,0.26) {0};
\node[left] at (0,0.78) {0.1};
\node[left] at (0,1.30) {0.25};
\node[left] at (0,1.82) {0.5};
\node[left] at (0,2.34) {0.75};
\node[left] at (0,2.86) {1};
\node[left] at (0,3.38) {1.5};
\node[below] at (0.35,0) {0};
\node[below] at (1.05,0) {0.1};
\node[below] at (1.75,0) {0.25};
\node[below] at (2.45,0) {0.5};
\node[below] at (3.15,0) {1};
\node[below] at (3.85,0) {2};
\node at (2.1,4) {Relation-only SW minus correct};
\node at (2.1,-.6) {$\eta$};
\node[rotate=90] at (-.65,1.8) {$\alpha$};
\end{scope}
\begin{scope}[shift={(6,0)}]
\fill[blue!12] (0.00,0.00) rectangle ++(.7,.52);
\node[text=black] at (0.35,0.26) {0.03};
\fill[blue!12] (0.70,0.00) rectangle ++(.7,.52);
\node[text=black] at (1.05,0.26) {0.03};
\fill[blue!12] (1.40,0.00) rectangle ++(.7,.52);
\node[text=black] at (1.75,0.26) {0.03};
\fill[blue!12] (2.10,0.00) rectangle ++(.7,.52);
\node[text=black] at (2.45,0.26) {0.03};
\fill[blue!12] (2.80,0.00) rectangle ++(.7,.52);
\node[text=black] at (3.15,0.26) {0.03};
\fill[blue!12] (3.50,0.00) rectangle ++(.7,.52);
\node[text=black] at (3.85,0.26) {0.03};
\fill[blue!15] (0.00,0.52) rectangle ++(.7,.52);
\node[text=black] at (0.35,0.78) {0.30};
\fill[blue!15] (0.70,0.52) rectangle ++(.7,.52);
\node[text=black] at (1.05,0.78) {0.30};
\fill[blue!15] (1.40,0.52) rectangle ++(.7,.52);
\node[text=black] at (1.75,0.78) {0.30};
\fill[blue!15] (2.10,0.52) rectangle ++(.7,.52);
\node[text=black] at (2.45,0.78) {0.30};
\fill[blue!15] (2.80,0.52) rectangle ++(.7,.52);
\node[text=black] at (3.15,0.78) {0.30};
\fill[blue!15] (3.50,0.52) rectangle ++(.7,.52);
\node[text=black] at (3.85,0.78) {0.30};
\fill[blue!21] (0.00,1.04) rectangle ++(.7,.52);
\node[text=black] at (0.35,1.30) {0.75};
\fill[blue!21] (0.70,1.04) rectangle ++(.7,.52);
\node[text=black] at (1.05,1.30) {0.75};
\fill[blue!21] (1.40,1.04) rectangle ++(.7,.52);
\node[text=black] at (1.75,1.30) {0.75};
\fill[blue!21] (2.10,1.04) rectangle ++(.7,.52);
\node[text=black] at (2.45,1.30) {0.75};
\fill[blue!21] (2.80,1.04) rectangle ++(.7,.52);
\node[text=black] at (3.15,1.30) {0.75};
\fill[blue!21] (3.50,1.04) rectangle ++(.7,.52);
\node[text=black] at (3.85,1.30) {0.75};
\fill[blue!32] (0.00,1.56) rectangle ++(.7,.52);
\node[text=black] at (0.35,1.82) {1.57};
\fill[blue!32] (0.70,1.56) rectangle ++(.7,.52);
\node[text=black] at (1.05,1.82) {1.57};
\fill[blue!32] (1.40,1.56) rectangle ++(.7,.52);
\node[text=black] at (1.75,1.82) {1.57};
\fill[blue!32] (2.10,1.56) rectangle ++(.7,.52);
\node[text=black] at (2.45,1.82) {1.57};
\fill[blue!32] (2.80,1.56) rectangle ++(.7,.52);
\node[text=black] at (3.15,1.82) {1.57};
\fill[blue!32] (3.50,1.56) rectangle ++(.7,.52);
\node[text=black] at (3.85,1.82) {1.57};
\fill[blue!43] (0.00,2.08) rectangle ++(.7,.52);
\node[text=black] at (0.35,2.34) {2.44};
\fill[blue!43] (0.70,2.08) rectangle ++(.7,.52);
\node[text=black] at (1.05,2.34) {2.44};
\fill[blue!43] (1.40,2.08) rectangle ++(.7,.52);
\node[text=black] at (1.75,2.34) {2.44};
\fill[blue!43] (2.10,2.08) rectangle ++(.7,.52);
\node[text=black] at (2.45,2.34) {2.44};
\fill[blue!43] (2.80,2.08) rectangle ++(.7,.52);
\node[text=black] at (3.15,2.34) {2.44};
\fill[blue!43] (3.50,2.08) rectangle ++(.7,.52);
\node[text=black] at (3.85,2.34) {2.44};
\fill[blue!54] (0.00,2.60) rectangle ++(.7,.52);
\node[text=white] at (0.35,2.86) {3.31};
\fill[blue!54] (0.70,2.60) rectangle ++(.7,.52);
\node[text=white] at (1.05,2.86) {3.31};
\fill[blue!54] (1.40,2.60) rectangle ++(.7,.52);
\node[text=white] at (1.75,2.86) {3.31};
\fill[blue!54] (2.10,2.60) rectangle ++(.7,.52);
\node[text=white] at (2.45,2.86) {3.31};
\fill[blue!54] (2.80,2.60) rectangle ++(.7,.52);
\node[text=white] at (3.15,2.86) {3.31};
\fill[blue!54] (3.50,2.60) rectangle ++(.7,.52);
\node[text=white] at (3.85,2.86) {3.31};
\fill[blue!77] (0.00,3.12) rectangle ++(.7,.52);
\node[text=white] at (0.35,3.38) {5.08};
\fill[blue!77] (0.70,3.12) rectangle ++(.7,.52);
\node[text=white] at (1.05,3.38) {5.08};
\fill[blue!77] (1.40,3.12) rectangle ++(.7,.52);
\node[text=white] at (1.75,3.38) {5.08};
\fill[blue!77] (2.10,3.12) rectangle ++(.7,.52);
\node[text=white] at (2.45,3.38) {5.08};
\fill[blue!77] (2.80,3.12) rectangle ++(.7,.52);
\node[text=white] at (3.15,3.38) {5.08};
\fill[blue!77] (3.50,3.12) rectangle ++(.7,.52);
\node[text=white] at (3.85,3.38) {5.08};
\node[left] at (0,0.26) {0};
\node[left] at (0,0.78) {0.1};
\node[left] at (0,1.30) {0.25};
\node[left] at (0,1.82) {0.5};
\node[left] at (0,2.34) {0.75};
\node[left] at (0,2.86) {1};
\node[left] at (0,3.38) {1.5};
\node[below] at (0.35,0) {0};
\node[below] at (1.05,0) {0.1};
\node[below] at (1.75,0) {0.25};
\node[below] at (2.45,0) {0.5};
\node[below] at (3.15,0) {1};
\node[below] at (3.85,0) {2};
\node at (2.1,4) {Source pairwise SW ($\times10^{-3}$)};
\node at (2.1,-.6) {$\eta$};
\node[rotate=90] at (-.65,1.8) {$\alpha$};
\end{scope}
\end{tikzpicture}